\documentclass[onefignum,onetabnum]{siamart220329}

\usepackage{graphicx}
\usepackage{epstopdf}
\usepackage{todonotes}
\usepackage{caption}
\usepackage{amsmath}
\usepackage{amsfonts}
\usepackage{enumerate}
\usepackage{enumitem}

\newsiamthm{assumption}{Assumption}
\newsiamremark{remark}{Remark}

\ifpdf
\DeclareGraphicsExtensions{.eps,.pdf,.png,.jpg}
\else
\DeclareGraphicsExtensions{.eps}
\fi

\renewcommand{\paragraph}[1]{\textbf{#1.}}

\newcommand{\R}{\mathbb{R}}

\newcommand{\bx}{x}

\newcommand{\by}{y}

\newcommand{\mG}{\mathsf G}

\newcommand{\mM}{\mathsf M}

\newcommand{\chol}{\mathsf L}

\renewcommand{\ldots}{\makebox[1em][c]{.\hfil.\hfil.}}

\newcommand{\indi}{\mathtt{1}}
\renewcommand{\d}{\mathrm{d}}

\newcommand{\rmse}{\mathrm{rmse}}
\newcommand{\supp}{\mathrm{supp}}
\newcommand{\E}{E}
\newcommand{\V}{\mathrm{var}}

\newcommand{\cov}{\mathrm{cov}}
\newcommand{\bigO}{\mathcal{O}}
\newcommand{\uni}{\mu}

\newcommand{\textqed}{\hfill\ensuremath{\Box}}

\newcommand{\lowsup}[1]{{\raisebox{-3pt}{\scriptsize$#1$}}}

\usepackage{tikz}
\usetikzlibrary{positioning,shapes,calc,3d}
\usepackage{pgfplots}
\usepgfplotslibrary{fillbetween}

\pgfplotsset{%
every axis/.append style={width=.45\textwidth,
axis x line=bottom, axis y line=left,
x axis line style={very thick,->}, y axis line style={very thick,->},
tick align=inside, tick style={thick},
every x tick label/.style={font=\footnotesize},
every y tick label/.style={font=\footnotesize},
},
every axis legend/.append style={
legend columns=1,
font=\footnotesize,
draw=none,
fill=white,
},
every axis x label/.style={at={(0.5,-0.1)},below,fill=none,fill opacity=1,text opacity=1},
every axis y label/.style={at={(-0.12,0.5)},fill=none,fill opacity=1,text opacity=1,rotate=90},
compat=newest,
}

\title{Deep importance sampling using tensor trains with application to {\it\bfseries a priori} and {\it\bfseries a posteriori} rare events\thanks{Submitted to the editors DATE.\funding{TC acknowledges support from the Australian Research Council under the grant DP210103092. SD acknowledges support from the Engineering and Physical Sciences Research Council New Investigator Award EP/T031255/1. RS is supported by
the Deutsche Forschungsgemeinschaft %(DFG, German Research Foundation)
under Germany's Excellence Strategy EXC 2181/1 - 390900948.
(STRUCTURES Excellence Cluster).
TC and RS also gratefully
acknowledge support from the Erwin Schr\"odinger Institute.
}}}

\author{Tiangang Cui\thanks{School of Mathematics, Monash University, Victoria 3800, Australia \email{tiangang.cui@monash.edu}}
\and Sergey Dolgov\thanks{Department of Mathematical Sciences, University of Bath, Bath, UK \email{s.dolgov@bath.ac.uk}}
\and Robert Scheichl\thanks{Institute for Mathematics and Interdisciplinary Center for Scientific Computing, Heidelberg University, Im Neuenheimer Feld 205, 69120 Heidelberg, Germany \email{r.scheichl@uni-heidelberg.de}}
}

\usepackage[centering]{geometry}
\allowdisplaybreaks

\begin{document}

\maketitle

\begin{abstract}
We propose a deep importance sampling method that is suitable for estimating rare event probabilities in high-dimensional problems.
We approximate the optimal importance distribution in a general importance sampling problem as the pushforward of a reference distribution under a composition of order-preserving transformations, in which each transformation is formed by a squared tensor-train decomposition.
The squared tensor-train decomposition provides a scalable ansatz for building order-preserving high-dimensional transformations via density approximations.
The use of composition of maps moving along a
sequence of intermediate densities alleviates the difficulty of directly approximating concentrated density functions.
To compute expectations over unnormalized probability
distributions, we design a ratio estimator that estimates the
normalizing constant using a separate importance distribution, again
constructed via a composition of transformations in tensor-train format.
This offers better theoretical variance reduction compared with
self-normalized importance sampling, and thus opens the door to
efficient computation of rare event probabilities in Bayesian
inference problems. Numerical experiments on problems constrained by
differential equations show little to no increase in
the computational complexity with the event probability going to zero,
and allow to compute hitherto unattainable estimates of rare event
probabilities for complex, high-dimensional posterior densities.
\end{abstract}



\section{Introduction}

In the analysis of many scientific and engineering systems, practitioners often assess the performance and the inherent uncertainty using expectations of functions of random variables or random processes. As a starting point, the potential sources of input uncertainty in the system are parametrized by some random variable and equipped with a prior distribution. Then, given some model that maps the uncertain parameters to observables, the {\it a priori} uncertainty can be reduced to the {\it a posteriori} uncertainty by conditioning on observed data to obtain the posterior distribution under the Bayesian framework. Depending on the availability of data, accurate estimates of {\it a priori} and {\it a posteriori} expectations of some output functionals are both of interest.

Analytical or asymptotic characterizations of the abovementioned expectations are often unavailable, because of non-analytically tractable posterior distributions,  nonlinear functions of interests, or a combination of both.  Thus, numerical techniques such as Monte Carlo methods must be employed.
Importance sampling provides a general tool to efficiently compute expectations of this sort by allocating computational resources to the ``important'' regions of the expectation problem. 
In the literature, adaptive importance sampling strategies have been developed to iteratively identify the important region and also to adaptively estimate importance distributions in some parametric family, e.g., mixture distributions \cite{cappe2008adaptive,douc2007convergence}.
In general, the construction of importance distributions in high dimensions is challenging, especially when the important region localizes to the tail of the input distribution, as we may not be able to accurately approximate the optimal importance distribution using parametric families. 
As a result, the mean square error of an importance sampling estimator may deteriorate quickly, sometimes exponentially, as the parameter dimension increases. 
This becomes more critical for rare event problems, where the rather small event probability, often on a scale of $10^{-6}$ or less, requires an accurate approximation to the optimal importance distribution, so that the relative mean square error can be controlled for a fixed computational budget.

We present a deep importance sampling method suitable for high-dimensional rare event problems. It employs the deep inverse Rosenblatt transport (IRT) developed in \cite{dafs-tt-bayes-2019} and \cite{cui2021deep} to adaptively approximate the optimal importance density using a composition of order-preserving maps. When the optimal importance density is multi-modal and concentrated in the tails of the input distribution, the composite structure is able to adapt to those complicated features.
Each of the maps in the composition is constructed using functional tensor-train (TT) decomposition and the cross algorithm \cite{bigoni2016spectral,gorodetsky2019continuous,hackbusch2012tensor,oseledets2010tt,oseledets2011tensor}. It provides a non-parametric ansatz for approximating the optimal importance density. Thus, it can be significantly more accurate than alternative importance sampling densities based on mixture distributions. 
In addition, for problems with sufficient regularity, the accuracy of TT approximations can be independent of the parameter dimension; see \cite{griebel2021analysis} for details. The computational complexity of building TT decompositions and the resulting transport maps scales linearly in the dimension. 
The proposed importance sampling scheme is further extended to handle input probability distributions with unknown normalizing constants, so it can be applied to estimate {\it a posteriori} expectations. Crucially, it is possible to construct a significantly more effective estimator than the familiar self-normalized importance sampling scheme, by constructing an additional importance density, again based on the deep IRT framework, but now targeting the optimized importance density for the normalizing constant.

To demonstrate the power of the proposed deep importance sampling, we present non-trivial applications in risk assessment of spatial, susceptible-infectious-removed models and contaminant transport in groundwater systems in the challenging regime of rare events.
Our numerical results suggest that the proposed method can accurately estimate both {\it a priori} and {\it a posteriori} expectations using several orders of magnitude smaller sample sizes compared to importance densities based on mixtures distributions. More importantly, the use of composition of maps and TT decomposition allows us to estimate rare event probabilities in high dimensions so far intractable by standard importance sampling methods. 

This paper is organized as follows. Section \ref{sec:background} provides background of the problem of interest. Section \ref{sec:deepis} presents and analyses the deep importance sampling scheme for computing {\it a priori} and {\it a posteriori} expectations. Section \ref{sec:rare} discusses the application to rare event estimation problems. Section \ref{sec:sir} and \ref{sec:elliptic} apply the proposed method to a spatial, susceptible-infectious-removed model and to contaminant transport in groundwater systems, respectively. Additional numerical examples and derivations are provided in Appendix. %Supplementary Material. 

\section{Background}\label{sec:background}

We consider a random variable $X$ taking values in $\mathcal{X} \subseteq \R^d$ and assign a prior probability density $\pi_0$ to it. Given an integrable function $f : \mathcal{X} \rightarrow \R$, our  goal is to estimate the expectation
$F = \E_{\pi_0}\{f(X)\}$.
Importance sampling methods approach this goal by choosing a suitable importance density $p$, satisfying the sufficient condition $\supp(f\pi_0) \subseteq \supp(p)$, 
and then estimating $\E_{p}\{f(X)\pi_0(X)/p(X)\}$ instead. 
Drawing $N$ independent and identically distributed (i.i.d.) samples from $p$, one can construct the {\it unbiased importance sampling estimator} of $F$:
\begin{equation}\label{eq:is_prior}
\hat{F}_{p,N} = \frac{1}{N} \sum_{i = 1}^N  \frac{f(X^i)\pi_0(X^i)}{p(X^i)}, \quad X^i \sim p.
\end{equation}
The performance of $\hat{F}_{p,N}$ is measured using the relative mean square error, 
\begin{equation}\label{eq:rmse}
\rmse(\hat{F}_{p,N},F) = \frac{\E\{(\hat{F}_{p,N} - F)^2\} }{F^2} =  \frac{\V_p(\hat{F}_{p,N})}{F^2}  + 
\frac{\vert \E(\hat{F}_{p,N}) - F \vert^2}{F^2},
\end{equation}
where $\V_{p}(g) = \E_{p}\{g(X)^2\} - \E_{p}\{g(X)\}^2$ gives the variance of a function $g: \mathcal{X} \rightarrow \R$ with respect to the density $p$. The relative mean square error~\eqref{eq:rmse} is minimized for any sample size $N$ by choosing the \emph{optimal} importance density $p^\ast \propto | f | \pi_0$ that minimizes $\V_p(f\pi_0/p)$ over all densities $p$ with $\supp(f\pi_0) \subseteq \supp(p)$. If the function of interest $f(x)$ is non-negative on $\mathcal{X}$, then we have $\V_{p^*}(f\pi_0/p^*) = 0$, which leads to a \emph{zero-variance} estimator. 

\begin{remark}
The estimator of the {\it a priori} expectation in \eqref{eq:is_prior} implicitly assumes that the normalizing constants of the prior $\pi_0$ and of the importance distribution~$p$ are known, or at least the ratio of those two constants. This is also one of the necessary conditions to ensure unbiasedness of the resulting estimator. In situations where the normalizing constants are unknown---such as in the estimation of {\it a posteriori} expectations discussed below---the normalizing constants or their ratio need to be estimated. The expectation is then estimated either as the ratio or as the product of two (potentially unbiased) estimators, leading in general to a biased estimator for finite sample sizes. 
\end{remark}

Given observed data $y \in \mathcal{Y} \subseteq \R^m$, under the Bayesian paradigm, the likelihood function $x \mapsto \mathcal{L}^y(x)$ updates the prior distribution $\pi_0$ on $X$ to the posterior distribution with density
\begin{equation}\label{eq:post}
\pi^y(x) = \frac{1}{Z} \mathcal{L}^y(x) \pi_0(x),\quad Z = \E_{\pi_0}\{\mathcal{L}^y(X)\},
\end{equation}
where $Z$ is the normalizing constant. Conditioned on observed data, the central goal of the paper is to estimate the {\it a posteriori} expectation
\begin{equation}\label{eq:Epost}
R = \E_{\pi^y}\{f(X)\} = \frac1Z \int_\mathcal{X} f(x)\,\mathcal{L}^y(x) \pi_0(x) \,\d x. %
\end{equation}

The {\it a posteriori} setting adds additional challenges. In particular, simulating i.i.d. random variables from the posterior is often impossible and the normalizing constant $Z$ is typically unknown. Since the posterior expectation can be written as the ratio 
\begin{equation}\label{eq:EpostR}
R = \frac{\E_{\pi_0}\{f(X) \mathcal{L}^y(X)\} }{ \E_{\pi_0}\{\mathcal{L}^y(X)\}},
\end{equation}
an alternative importance sampling estimator can be constructed by carefully selecting two
importance densities $p$ and $q$ such that $\supp(f\pi)\subseteq\supp(p)$ and $\supp(\pi)\subseteq\supp(q)$ to estimate the numerator and the denominator of \eqref{eq:EpostR}, which now can be equivalently written as
\begin{equation}\label{eq:ratio2}
Q = \E_p\bigg\{ \frac{f(X) \mathcal{L}^y(X) \pi_0(X)}{p(X)} \bigg\}, \quad  Z = \E_q\bigg\{ \frac{\mathcal{L}^y(X) \pi_0(X)}{q(X)}\bigg\},
\end{equation}
respectively. Drawing i.i.d. samples $X_p^i\sim p$ and $X_q^i\sim q$, we can construct unbiased importance sampling estimators
\begin{equation}\label{eq:ratio1}
\hat{Q}_{p,N} = \frac1N\sum_{i=1}^N \frac{f(X_p^i) \mathcal{L}^y(X_p^i) \pi_0(X_p^i)}{p(X_p^i)},  \quad \hat{Z}_{q,N} = \frac1N\sum_{i=1}^N \frac{\mathcal{L}^y(X_q^i) \pi_0(X_q^i)}{q(X_q^i)},
\end{equation}
to estimate $Q$ and $Z$, respectively. This leads to the {\it ratio estimator} 
\begin{equation}\label{eq:ratio}
\hat{R}_{p,q,N} := \frac{\hat{Q}_{p,N}}{ \hat{Z}_{q,N}},
\end{equation}
for the {\it a posteriori} expectation. Although  $\hat{Q}_{p,N}$ and $\hat{Z}_{q,N}$ are unbiased, the ratio estimator $\hat{R}_{p,q,N}$ is biased. We will discuss the impact of this bias in later sections. 

A computationally convenient choice is the so-called {\it self-normalized importance sampling estimator} with $p = q$. However, the respective optimal importance densities $p^* \propto | f | \mathcal{L}^y\pi_0$ and $q^* = \pi^y$ for $Q$ and $Z$ may differ significantly, e.g., when the function of interest $f$ only takes significant values in the tail of the posterior density $\pi^y$. We propose to construct separate, near-optimal importance densities $p$ and $q$ to reduce the overall relative mean square error of the ratio estimator \eqref{eq:ratio}. 

A particular application is the estimation of failure probabilities of physical or engineering systems to assess their reliability or to inform policy makers. Given a response function $h: \mathcal{X} \mapsto \R$, system failure is characterized by determining whether the output of $h$ falls inside of a set $\mathcal{A} \subset \R$. Thus, the function of interest representing a system failure becomes
\begin{equation}\label{eq:qoi}
f(x) = \indi_\mathcal{A} \{ h(x) \} 
\end{equation}
where $\indi_\mathcal{A}(\cdot)$ denotes the indicator function of the set $\mathcal{A}$. Depending on the availability of data, both the {\it a priori} and the {\it a posteriori} failure probabilities, 
\begin{equation}\label{eq:prior_failure}
\mathrm{pr}_{\pi_0}\{ h(X) \in \mathcal{A}\} = \E_{\pi_0}\{f(X)\}, \quad 
\mathrm{pr}_{\pi^y}\{ h(X) \in \mathcal{A}\} = \E_{\pi^y}\{ f(X) \},
\end{equation} 
provide risk assessment criteria associated with the response function $h$. Estimating those probabilities is particularly challenging when the failure set 
\(
\mathcal{X}_F := \{ x\in\mathcal{X} : f(x) = 1 \}
\)
has a very small probability mass, also referred to as a \emph{rare event}. 

Most of the existing literature for complex high-dimensional applications focuses on estimating \emph{a priori} failure probabilities, e.g., \cite{dodwell2021multilevel,elfverson2016multilevel,peherstorfer2016multifidelity,peherstorfer2018multifidelity,uribe2020cross,wagner2020multilevel,wagner2021error}, while our approach applies equally to {\it a posteriori} failure probabilities and clearly outperforms the classical cross entropy method \cite{botev2008efficient}; see Sections \ref{sec:rare}--\ref{sec:elliptic} for numerical examples.

\section{Deep importance sampling using TT}\label{sec:deepis}
\subsection{Problem setup}
\label{sec:problem}

To encompass both {\it a priori} and {\it a posteriori} expectations the optimal importance density is presented in the general form of
\begin{equation}\label{eq:is_tt}
p^\ast(x) = \frac{1}{\zeta^\ast} \, \rho^\ast(x),\quad \zeta^\ast = \int_\mathcal{X} \rho^\ast(x) \d x,
\end{equation}
where $\rho^\ast(x)$ is the unnormalized optimal importance density and $\zeta^\ast$ is the normalizing constant. This includes {\it a priori} expectations, where $\rho^\ast=|f|\pi_0$, as well as the numerator and the denominator of the ratio estimator \eqref{eq:ratio} for {\it a posteriori} expectations, where $\rho^\ast = |f|\mathcal{L}^y\pi_0$ and  $\rho^\ast = \mathcal{L}^y\pi_0$, respectively. For the remainder we assume that $\zeta^\ast$ is unknown and that we can only evaluate the unnormalized density $\rho^*$.

Our ultimate goal is to build a normalized approximation to the optimal $p^*$ as the pushforward of  an analytically tractable and product-form reference density $\lambda(x) = \prod_{k = 1}^d \!\lambda_k(x_k)$ under an order-preserving map $\mathcal{T}:\R^d \rightarrow \R^d$. Then, the resulting transformation can be used to generate i.i.d. random variables for  importance sampling. We make the following assumptions about the importance sampling problem: 

\begin{assumption}\label{ass:is1}
The function of interest $f$ is non-negative. 
\end{assumption}

\begin{assumption}\label{ass:is2}
The ratio $\rho^*/\pi_0$ has finite mean and finite second moment with respect to $\pi_0$. 
\end{assumption}

\begin{assumption}\label{ass:is3}
The reference density $\lambda$ satisfies $\sup_{x\in\mathcal{X}} \pi_0(x)/\lambda(x) < \infty$.
\end{assumption}

Assumption \ref{ass:is1} holds for the failure probability  problem, which is our main application. By focusing on non-negative $f$, the optimal importance density leads to a zero-variance estimator. Thus, our goal is to design importance densities that closely approximate the optimal density to provide \emph{near zero-variance} estimators.
However, our discussion can easily be extended to general functions. One can decompose any function $f$ as the difference of two non-negative functions
\(
f(x) = f_+(x) - f_-(x),
\)
where $f_+(x) = f(x)\indi_{\{f(x)>0\}}(x)$ and $f_-(x) = -f(x)\indi_{\{f(x)\leq0\}}(x)$.
The original expectation $\E_{\pi_0}\{f(X)\}$ can then be computed from $\E_{\pi_0}\{f_+(X)\} - \E_{\pi_0}\{f_-(X)\}$, if both $f_+$ and $f_-$ are integrable. 

Assumption \ref{ass:is2} guarantees that the nominal estimator, which uses the prior density $\pi_0$ as the importance density, satisfies the assumptions of the central limit theorem. We adopt this assumption to analyse the relative mean square error of our proposed estimators. Assumption \ref{ass:is3} is introduced to ensure $\supp(\rho^*) \subseteq \supp(\lambda)$ for all the cases of interest specified at the start of Section~\ref{sec:problem}. Then $\lambda$ can be used as reference density to avoid any potential singularities in approximating the optimal importance density. In most cases, $\lambda$ will be the prior density. 

\subsection{From TT to squared IRT}\label{sec:sirt}

The central tool in our new approach is an approximation of the square root of the unnormalized optimal importance density $\rho^*$ in a functional TT decomposition 
\begin{equation}\label{eq:tt_sqrt}
\surd \rho^*(x) \approx \tilde{g}(x) = \mG_{1}(x_1) \cdots \mG_{k}(x_k) \cdots \mG_{d}(x_d),
\end{equation}
where each of the $\mG_{k}(x_k)$ is a matrix-valued function of size $r_{k-1} \times r_k$, with $r_0 = r_d = 1$. Using a representation of $\surd \rho^*$ in tensor product form with $n_k$ basis functions in the $k$th coordinate, such a TT decomposition can be computed very efficiently without incurring the curse of dimensionality for a wide range of densities via alternating linear schemes together with cross approximation \cite{bigoni2016spectral,gorodetsky2019continuous,oseledets2010tt}. We employ the functional extension of the alternating minimal energy method with residual-based rank adaptation of \cite{dolgov2014alternating}. It requires only $\mathcal O(dnr^2)$ evaluations of the density $\rho^*$ and $\mathcal O(dnr^3)$ floating point operations, where $n = \max_k n_k$ and  $r = \max_k r_k$. For more details see \cite{cui2021deep,dafs-tt-bayes-2019}. In general, the maximal rank $r$ depends on the dimension $d$ and can be large when the density $\rho^*$ concentrates in some part of its domain, but some theoretical results exist that provide rank bounds. While \cite{rdgs-tt-gauss-2020} establish specific bounds for certain multivariate Gaussian densities that depend poly-logarithmically on $d$, \cite{griebel2021analysis} prove dimension-independent bounds for general functions in weighted spaces with dominating mixed smoothness.

Starting with a TT decomposition of $\surd \rho^*$, we construct the following approximation to the normalized optimal importance density,
\begin{equation}\label{eq:pdf_sirt}
p(x)  = \frac{1}{ \zeta} \, \rho(x), \quad \rho(x) = \tilde g(x)^2 +\tau \lambda(x),\quad \zeta = \int_{\mathcal{X}} \{ \tilde g(x)^2 + \tau \lambda(x)\} \d x,
\end{equation}
for some $\tau > 0$. The additional term $\tau \lambda(x)$ guarantees that $\supp(\rho^*) \subseteq \supp(p)$, and thus the importance sampling estimator defined by the approximate density $p$ is unbiased. The following lemma, whose original proof is given in \cite{cui2021deep}, shows how to choose $\tau$ as a function of the error in $\tilde g$ in the $L^2$-norm, to be able to control the overall error of the approximate density $p$ in Hellinger distance.

\begin{lemma}
\label{lemma:dhell_sirt}
Suppose $\|\surd \rho^* - \tilde g \|_2 \leq \epsilon$ and $\tau \leq \epsilon^2$. Then, the exact normalizing constant $\zeta^*$ in \eqref{eq:is_tt} and its approximation $\zeta$  in \eqref{eq:pdf_sirt} satisfy $| \zeta^* - \zeta| \leq \surd 2 \epsilon$ and the Hellinger distance between $p^*$ and its normalized approximation $p$ defined in \eqref{eq:pdf_sirt} can be bounded by $D_{\rm H}(p^*, p)\leq 2 \epsilon / \surd \zeta^*$.
\end{lemma}

\begin{definition}
For any vector $x \in \R^d$ and any index $k\in \{1, \ldots,d\}$, the first $k-1$ coordinates and the last $d-k$ coordinates of $x$ are expressed as
\(
x_{<k} = [x_1, \ldots, x_{k-1}]^\top
\) 
and
\(
x_{>k} = [x_{k+1}, \ldots, x_{d}]^\top,
\)
respectively. Similarly, we write $x_{\leq k} = (x_{<k}, x_k)$, $x_{\geq k} = (x_{k}, x_{>k})$, $x_{\leq 1} = x_1$, $x_{\geq d} = x_{d}$, and $x_{\leq d} = x$.
\end{definition}

Following \cite{cui2021deep}, to build an efficient sampling method based on this density approximation we now build an order-preserving map $\mathcal{Q}: \R^d \rightarrow \mathcal{X} $, the \emph{generalized IRT}, such that the pushforward of the reference density $\lambda$ under the map $\mathcal{Q}$ is the normalized approximate density $p$, i.e., $\mathcal{Q}_\sharp \, \lambda = p$.
Exploiting the separable structure of the TT approximation $\tilde g$, the unnormalized marginal densities
\begin{align}
\rho_{\leq k}(x_{\leq k}) & = \int_{\mathcal{X}_{>k}}  \rho(x_{\leq k}, x_{>k}) \, \d x_{>k} =  \int_{\mathcal{X}_{>k}}  \tilde g(x_{\leq k}, x_{>k})^2 \, \d x_{>k} + \tau \lambda_{\leq k}(x_{\leq k}), \label{eq:marginal_sirt}
\end{align}
with $\lambda_{\leq k}(x_{\leq k}) {=} \prod_{j = 1}^{k} \lambda_j(x_j)$ for $1 {\leq} k {<} d$, can be computed analytically via a sequence of one-dimensional integrations. Finally, by integrating the univariate unnormalized marginal density $\rho_{\leq 1}(x_1)$, we obtain the normalizing constant $\zeta$. We provide the implementation detail of the marginalization procedure in Appendix. %Supplementary Material.

Thus, the normalized densities for the marginal random variables $X_{\leq k}$ are 
\[
p_{\leq k}(x_{\leq k}) = \frac1\zeta \rho_{\leq k}(x_{\leq k}).
\]
Now, the joint random variable $X$ can be equivalently expressed as a one-dimensional marginal and a sequence of $d-1$ one-dimensional conditional random variables,
\(
X_1, X_2 | X_{<2}, \cdots, X_{d} | X_{<d},
\)
with distribution functions 
\begin{equation}\label{eq:cond_cdf}
\mathcal{F}_{\leq 1}(x_1) = \int_{-\infty}^{x_1} p_{\leq 1}(x_1')\,\d x_1', \quad \mathcal{F}_{k | < k}(x_k | x_{<k}) = \int_{-\infty}^{x_k} \frac{p_{\leq k}(x_{< k}, x_k')}{p_{< k}(x_{< k})} \, \d x_k',
\end{equation}
respectively. This defines the {\it Rosenblatt transport} according to \cite{rosenblatt1952remarks},
\begin{equation}\label{eq:rosenblatt}
\xi = \left[\begin{array}{l} \xi_1 \\ \vdots \\ \xi_d\end{array}\right] =  \left[\begin{array}{l} \mathcal{F}_{\leq 1}\;\;(x_1)\\ \vdots \\ \mathcal{F}_{d | < d}(x_d | x_{<d})  \end{array}\right] = \mathcal{F}(x).
\end{equation}
Given $X \sim p$, the random variable $\Xi = \mathcal{F}(X)$ is distributed uniformly in the unit hypercube $[0,1]^d$. Since the $k$-th component of $\mathcal{F}$ is a scalar valued function $\mathcal{F}_{k | < k}: \R^{k} \mapsto \R$, depending on the first $k$ variables only, the map $\mathcal{F}$ is {\it lower-triangular}.

The reason for decomposing the square root $\surd \rho^*$ of the unnormalized importance density instead of $\rho^*$ becomes apparent here. Directly decomposing the density $\rho^*$ using TTs, the non-negativity of the approximated density function can not be guaranteed due to rank truncation. Approximating $\surd \rho^*$ preserves non-negativity without any loss of smoothness in the resulting approximate density $\rho$ and in all marginal densities $\rho_{\leq k}$, $1 \leq k < d$. Crucially, it also guarantees that all one-dimensional distribution functions in \eqref{eq:cond_cdf} are monotonically increasing and that the map $\mathcal{F}$, as well as its inverse are order-preserving and almost surely differentiable.
For a wide range of basis functions---including piecewise Lagrange polynomials, (weighted) spectral polynomials such as Chebyshev and Hermite polynomials, and Fourier series---closed-form, analytical expressions of the marginal densities in \eqref{eq:marginal_sirt}, of the conditional distribution functions in \eqref{eq:cond_cdf}, and of the resulting Rosenblatt transport in \eqref{eq:rosenblatt} are available. We refer the reader to the appendix of \cite{cui2023self} for details.

Denoting the uniform density on $[0,1]^d$ by $\uni$,  the pullback of $\uni$ under $\mathcal{F}$ satisfies
\[
\mathcal{F}^\sharp \, \uni(x) = \uni \big( \mathcal{F}(x)\big)\,\big| \nabla_{x} \mathcal{F}(x) \big|  = \big| \nabla_{x} \mathcal{F}(x) \big| = p(x).
\]
The product-form reference density $\lambda(u)$ is naturally equipped with the diagonal map 
\[
\xi = \mathcal{R}(u) = \big[ \mathcal{R}_1(u_1), \ldots, \mathcal{R}_k(u_k), \ldots, \mathcal{R}_d(u_d) \big]^\top, \quad \mathcal{R}_k(u_k) = \int_{-\infty}^{u_k} \lambda_k(u_k') \d u_k',
\]
such that $\mathcal{R}_\sharp \,\lambda = \mu$. Thus, the composite map $\mathcal{Q}=\mathcal{F}^{-1}\circ \mathcal{R}$ also has the lower-triangular structure and satisfies $\mathcal{Q}_\sharp\,\lambda = p$. Thus, one can first generate random variables $U \sim \lambda$, distributed according to the reference density $\lambda$, and then apply the {\it general IRT} 
\(
X = \mathcal{Q} (U)
\)
to obtain a random variable $X \sim p$. The map $\mathcal{Q}: \R^d \rightarrow \mathcal{X}$ is again {\it lower-triangular} and can be evaluated successively as
\begin{equation}\label{eq:irt}
x = \Big[ \mathcal{F}_{\leq 1}^{-1}\{R_1(u_1) \}, \ldots, \mathcal{F}_{d | < d}^{-1}\{R_d(u_d) | x_{<d}\} \Big]^\top.
\end{equation}
Thus defined \emph{squared IRT} can be also used as an efficient \emph{conditional distribution method} in the classical sense, see, e.g., \cite{Johnson-1987}.

We want to highlight some relevant work. In the Bayesian context, the work of \cite{eigel2020low,eigel2018sampling} employs TT to approximate elements of the posterior density, such as the log-likelihood function, to compute posterior statistics.  In comparison, our method approximates the optimal importance density and the expectation to be estimated for general problems using TT, while naturally devising an IRT to remove potential approximation bias via sampling. 

Practical implementations of the general Rosenblatt transport in high-dimensions were previously investigated within a variational framework. One such class of methods, cf.~\cite{baptista2020adaptive,parno2018transport,wan2020coupling}, adopts a \emph{map-from-samples} approach that estimates the map $\mathcal{Q}$ by minimizing the Kullback--Leibler divergence of the target density from the pushforward of the reference density under $\mathcal{Q}$. In particular, the work of \cite{wan2020coupling} learns the map $\mathcal{Q}$ using reduced order models to accelerate importance sampling estimators. The map-from-samples approach is flexible to implement, as it only requires a set of samples drawn from the target density. However, it comes with an $O(N^{-1/2})$ error rate, where $N$ is the sample size, due to the Monte Carlo estimate of the KL divergence. See \cite{wang2022minimax} and references therein for the analysis. Another related approach is the variational density estimation in the TT format \cite{novikov2021ttde}, in which it is possible to derive the Rosenblatt transport in TT format after the density estimation. 

When samples from the target density are hard to obtain---e.g., the computation of {\it a posteriori} expectations and rare event estimations considered in this work---one may employ an alternative class of methods that adopts a \emph{map-from-density} approach. The map-from-density approach builds the Rosenblatt transport $\mathcal{Q}$ by minimizing the Kullback--Leibler divergence of the pushforward of the reference density under $\mathcal{Q}$ from the target density, cf.~\cite{bigoni2019greedy,el2012bayesian,spantini2018inference}. The training of this class of methods is often quite involved in practice---the objective function presents many local minima and each optimization iteration requires many evaluations of the unnormalized target density at transformed reference variables under the candidate map. Our method also uses pointwise evaluation of the tagret density, and thus can be considered as a map-from-density approach. Instead of the computationally demanding iterative minimization of the Kullback--Leibler divergence, our method builds the TT-Cross approximation of the square root of an unnormalized density function, which naturally relates to the Hellinger distance (cf. Lemma \ref{lemma:dhell_sirt}). Under our construction, the resulting Rosenblatt transport maps exactly to the approximated target density built by TT-Cross.

\subsection{From IRT to deep importance sampling}\label{sec:dirt}

For problems such as rare event estimation, the optimal importance density can concentrate to a small region of the parameter space, or even to a sub-manifold, due to complex nonlinear interactions. In this situation, constructing in one step a TT approximation of $\surd \rho^*$ may result in rather high tensor ranks. It is also challenging to find an appropriate basis to efficiently discretize $\surd \rho^*$ that can adapt to high-probability regions of the optimal importance density. As a consequence, both $r$ and $n$ can become very large.

We overcome this difficulty by building a composition of maps
\(
\mathcal{T}^\lowsup{(L)} = \mathcal{Q}^\lowsup{(1)} \circ \mathcal{Q}^\lowsup{(2)} \circ \cdots \circ \mathcal{Q}^\lowsup{(L)},
\)
that can adapt to a concentrated optimal importance density layer-by-layer. The adaptive construction is guided by a sequence of unnormalized intermediate densities
\(
\phi^\lowsup{(1)}, \phi^\lowsup{(2)}, \ldots , \phi^\lowsup{(L)} \equiv \rho^*
\)
with increasing complexity. To specify the adaptation, we denote the $\ell$th normalized intermediate density as
\[
\varphi^\lowsup{(\ell)}(x) = \frac{1}{\omega^\lowsup{(\ell)}}\, \phi^\lowsup{(\ell)}(x),\quad \omega^\lowsup{(\ell)} = \int_\mathcal{X} \phi^\lowsup{(\ell)}(x) \d x.
\]

At any layer $\ell$, the pushforward of the reference density $\lambda$ under the partial composition $\mathcal{T}^\lowsup{(\ell)}$ is constructed such that it approximates the $\ell$th normalized intermediate density, i.e., $\{\mathcal{T}^\lowsup{(\ell)} \}_\sharp \,\lambda \approx \varphi^\lowsup{(\ell)}$, with a controlled error. This leads to a recursive construction procedure. Given $\mathcal{T}^\lowsup{(\ell)}$, we need to add a new layer $\mathcal{Q}^\lowsup{(\ell{+}1)}$ so that the new composition $\mathcal{T}^\lowsup{(\ell{+}1)} = \mathcal{T}^\lowsup{(\ell)}\circ \mathcal{Q}^\lowsup{(\ell{+}1)}$ yields 
\[
\{ \mathcal{T}^\lowsup{(\ell)} \circ \mathcal{Q}^\lowsup{(\ell{+}1)} \}_\sharp \,\lambda \approx \varphi^\lowsup{(\ell{+}1)}.
\]
This is equivalent to finding $\mathcal{Q}^\lowsup{(\ell{+}1)}$ such that
\(
\{\mathcal{Q}^\lowsup{(\ell{+}1)} \}_\sharp \,\lambda \approx \{\mathcal{T}^\lowsup{(\ell)}\}^\sharp \, \varphi^\lowsup{(\ell{+}1)}. 
\)
Thus, we can build $\mathcal{Q}^\lowsup{(\ell{+}1)}$ as a squared IRT that pushes forward the reference density $\lambda$ to the pullback density $\{\mathcal{T}^\lowsup{(\ell)}\}^\sharp \varphi^\lowsup{(\ell{+}1)}$. Since the pushforward of $\lambda$ under $\mathcal{T}^\lowsup{(\ell)}$ approximates $\varphi^\lowsup{(\ell)}$, the pullback of the normalized density $\varphi^\lowsup{(\ell)}$ under $\mathcal{T}^\lowsup{(\ell)}$ satisfies
\begin{equation}\label{eq:layerk1}
\{\mathcal{T}^\lowsup{(\ell)}\}^\sharp \, \varphi^\lowsup{(\ell)} (u) = \varphi^\lowsup{(\ell)} \{ \mathcal{T}^\lowsup{(\ell)}(u)\}  \big| \nabla \mathcal{T}^\lowsup{(\ell)}(u) \big| \approx \lambda(u).
\end{equation}
Similarly, we can see that
\begin{align*}%
\{\mathcal{T}^\lowsup{(\ell)}\}^\sharp \, \varphi^\lowsup{(\ell{+}1)} (u) & = \varphi^\lowsup{(\ell{+}1)} \{ \mathcal{T}^\lowsup{(\ell)}(u) \}  \big| \nabla \mathcal{T}^\lowsup{(\ell)}(u) \big| 
\approx \frac{\varphi^\lowsup{(\ell{+}1)} \{ \mathcal{T}^\lowsup{(\ell)}(u) \} }{\varphi^\lowsup{(\ell)} \{ \mathcal{T}^\lowsup{(\ell)}(u) \} } \lambda(u) . 
\end{align*}
With suitable intermediate densities, the ratio $\varphi^\lowsup{(\ell{+}1)}/\varphi^\lowsup{(\ell)}$ is significantly less concentrated than the optimal importance density $\rho^*$. As a result, it will be much easier to approximate the map $\mathcal{Q}^\lowsup{(\ell{+}1)}$ rather than directly attempting to approximate the pullback of $\rho^*$.

Although the normalizing constant of $\varphi^\lowsup{(\ell{+}1)}$ is unknown, it is possible to recursively decompose the square root of the unnormalized pullback density $\{\mathcal{T}^\lowsup{(\ell)}\}^\sharp \phi^\lowsup{(\ell{+}1)}$ in TT format using the construction outlined in Section \ref{sec:sirt}. This procedure is summarized in Alg.~\ref{alg:dirt}.

\begin{algorithm}[h]
\caption{Construction of deep importance density. \label{alg:dirt}}
\begin{tabbing}
\enspace Input: reference density $\lambda$ and unnormalized intermediate densities $\phi^\lowsup{(1)}, \ldots, \phi^\lowsup{(L)}$ \\
\enspace Initialize the map as $\mathcal{T}^\lowsup{(0)}\gets I$ to have $\mathcal{T}^\lowsup{(0)}(x) = x$. \\
\enspace For $\ell = 1, \ldots, L$, apply all steps as outlined in Section \ref{sec:sirt}: \\
\qquad Factorize the square root of $\{\mathcal{T}^\lowsup{(\ell-1)}\}^\sharp \phi^\lowsup{(\ell)}(x)$ in a TT format $\tilde g^\lowsup{(\ell)}(x)$. \\
\qquad Choose appropriate $\tau^\lowsup{(\ell)}$. \\ 
\qquad Construct the approximation $\{\mathcal{T}^\lowsup{(\ell-1)}\}^\sharp \phi^\lowsup{(\ell)} (x) \approx \rho^\lowsup{(\ell)}(x) = \tilde g^\lowsup{(\ell)}(x)^2 + \tau^\lowsup{(\ell)} \lambda(x)$. \\
\qquad Compute the normalizing constant $\zeta^\lowsup{(\ell)}$.\\
\qquad Compute the IRT $\mathcal{Q}^\lowsup{(\ell)}$ associated with $\rho^\lowsup{(\ell)}$ as in \eqref{eq:irt}.\\
\qquad Update the composition as $\mathcal{T}^\lowsup{(\ell)} \gets \mathcal{T}^\lowsup{(\ell{-}1)} \circ \mathcal{Q}^\lowsup{(\ell)}$.\\
\enspace Return $\{\tilde g^\lowsup{(\ell)}, \tau^\lowsup{(\ell)}, \zeta^\lowsup{(\ell)}\}_{\ell=1}^{L}$ and the composite map $\mathcal{T}^\lowsup{(L)}$. 
\end{tabbing}
\end{algorithm}

Given the output of Alg.~\ref{alg:dirt}, the 
pushforward of the reference density $\lambda$ under the  composite map $\mathcal{T}^\lowsup{(L)}$ has the normalized density $\bar{p} = \{\mathcal{T}^\lowsup{(L)}\}_\sharp\,\lambda$ with
\begin{equation}\label{eq:dirt_density}
\bar{p}(x) = \bigg\{\prod_{\ell=1}^L \zeta^\lowsup{(\ell)}\bigg\}^{-1} \{ \tilde{g}^\lowsup{(1)}(x)^2 + \tau^\lowsup{(1)} \lambda(x) \} \prod_{\ell=2}^L \bigg(\frac{ \tilde{g}^\lowsup{(\ell)}[ \{\mathcal{T}^\lowsup{(\ell{-}1)}\}^{-1}(x)]^2}{ \lambda [ \{\mathcal{T}^\lowsup{(\ell{-}1)}\}^{-1}(x)] } + \tau^\lowsup{(\ell)} \bigg).
\end{equation}
Since the Hellinger distance is invariant to change of measure, the composition map satisfies
\[
D_\text{H}\left[\{\mathcal{T}^\lowsup{(\ell{-}1)}\circ\mathcal{Q}^\lowsup{(\ell)}\}_\sharp\lambda, \varphi^\lowsup{(\ell)}\right] = D_\text{H}\left[ \{\mathcal{Q}^\lowsup{(\ell)}\}_\sharp\lambda, \{\mathcal{T}^\lowsup{(\ell{-}1)}\}^\sharp\varphi^\lowsup{(\ell)}\right], \quad \text{for} \quad 1 \leq \ell \leq L.
\]
As a consequence, the total Hellinger error of the approximate optimal importance density $\bar{p} = \{\mathcal{T}^\lowsup{(L)}\}_\sharp\lambda$ is equivalent to the Hellinger error in the final iteration, 
\(
D_\text{H}[\{\mathcal{Q}^\lowsup{(L)}\}_\sharp\lambda, \{\mathcal{T}^{(L{-}1)}\}^\sharp\varphi^\lowsup{(L)} ],
\)
which can be controlled by the $L^2$-error of the TT approximation, as shown in Lemma \ref{lemma:dhell_sirt}.

Assuming that the function of interest $f$ is non-negative, the goal of deep importance sampling is to estimate the normalizing constant $\zeta^* = E_{\bar p}\{\rho^*(X)/\bar{p}(X)\}$. 
Using the change of variable $X = \mathcal{T}^\lowsup{(L)}(U)$, where $X\sim\bar{p}$ and $U \sim \lambda$, the normalizing constant can be expressed equivalently as an expectation with respect to the reference density $\lambda$, such that 
\[
\zeta^* = E_{\lambda}\left[\frac{\rho^*\{T(U)\}}{\bar{p}\{T(U)\}}\right]. 
\]
This leads to the \emph{deep importance sampling estimator}
\begin{equation}\label{eq:dirt_est2}
\hat{\zeta}_{\bar{p}, N} = \frac{1}{N} \sum_{i = 1}^N \frac{\rho^*\{\mathcal{T}^\lowsup{(L)}(U^i)\}}{\bar{p}\{\mathcal{T}^\lowsup{(L)}(U^i)\}}, \quad  U^i \sim \lambda.
\end{equation}
Its properties are established in the following lemma.

\begin{lemma}
\label{lemma:rel_var}
Suppose Assumptions \ref{ass:is1}--\ref{ass:is3} holds, and let $p^*= \rho^*/\zeta^\ast$ and $\bar{p}$  be the exact optimal importance density \eqref{eq:is_tt} and its approximation in \eqref{eq:dirt_density}, respectively.
\begin{enumerate}[leftmargin=*]
\item Then %
$\supp(p^*)\subseteq \supp(\bar{p})$, 
\(
\E_{\bar{p}}(\rho^*/\bar{p}) = \zeta^*
\)
and
\(
\V_{\bar{p}}(\rho^*/\bar{p}) < \infty.
\)
\item Assuming furthermore $\int \{\rho^*(x)/\pi_0(x)\}^3 \pi_0(x) \d x < \infty$, then
\[
\V_{\bar{p}}(p^*/\bar{p}) \leq C_p  D_{\rm H}(p^\ast, \bar p),\quad \text{where} \quad C_p = 2 \left[ \E_{p^*}\{(p^*/\bar{p})^2\} - \E_{\bar{p}}\{(p^*/\bar{p})^2\} \right]^{1/2}. 
\]
\item Assuming instead that $\sup_{x\in\mathcal{X}}p^*(x)/\bar{p}(x) =  M_{p^*,\bar{p}} < \infty $, then
\[
\V_{\bar{p}}(p^*/\bar{p}) \leq C_m  D_{\rm H}(p^\ast, \bar p)^2, \quad \text{where} \ \ C_m = 4 + 4 M_{p^*,\bar{p}} . 
\]
\end{enumerate}
\end{lemma}
\begin{proof}
Because $\tilde{g}^\lowsup{(\ell)}(x)^2 \geq 0$ and $\lambda(x)\geq 0$ for all $x\in\mathcal{X}$, the density $\bar{p}(x)$ satisfies
\begin{equation}\label{eq:dirt_pbound}
\bar{p}(x) \geq \lambda(x)  \bigg\{\prod_{\ell=1}^L \frac{\tau^\lowsup{(\ell)} }{\zeta^\lowsup{(\ell)}}\bigg\}
\end{equation}
for all $x\in\mathcal{X}$, which leads to $\supp(\lambda) \subseteq \supp(\bar{p})$. Under Assumption \ref{ass:is3}, we have $\supp(\rho^*) \subseteq \supp(\lambda)\subseteq \supp(\bar{p})$, and thus we can express $\zeta^*$ as
\[
\zeta^* = \int_\mathcal{X} \frac{\rho^*(x)}{\pi_0(x)} \pi_0(x) \, \d x = \int_\mathcal{X} \frac{\rho^*(x)}{\bar{p}(x)} \bar{p}(x) \, \d x.
\]
Furthermore, the identity in \eqref{eq:dirt_pbound} also leads to 
\[
\frac{\pi_0(x)}{\bar{p}(x)} \leq \frac{\pi_0(x)}{\lambda(x)} \bigg\{\prod_{\ell=1}^L \frac{\zeta^\lowsup{(\ell)}}{\tau^\lowsup{(\ell)} }\bigg\}.
\]
Together with Assumption \ref{ass:is3}, we also have $\sup_{x\in\mathcal{X}} \pi_0(x)/\bar{p}(x) < \infty$. This way, the second moment $\E_{\bar{p}}[(\rho^*/\bar{p})^2]$ satisfies
\begin{align*}
\E_{\bar{p}}\Big\{\Big(\frac{\rho^*}{\bar{p}}\Big)^2\Big\} & = \int_\mathcal{X} \Big\{\frac{\rho^*(x)}{\pi_0(x)}\Big\}^2 \frac{\pi_0(x)}{\bar{p}(x)} \pi_0(x)\, \d x \leq \E_{\pi_0}\Big\{\Big(\frac{\rho^*}{\pi_0}\Big)^2\Big\} \sup_{x\in\mathcal{X}} \frac{\pi_0(x)}{\bar{p}(x)}.
\end{align*}
Then, we have $\E_{\bar{p}}\{(\rho^*/\bar{p})^2\}<\infty$ by Assumption \ref{ass:is2} and thus the first result follows. 

Recall that the relative variance takes the form
\[
\V_{\bar{p}}(p^*/\bar{p}) = \E_{\bar{p}}\{(p^*/\bar{p})^2\} - \E_{\bar{p}}(p^*/\bar{p})^2, 
\]
where $\E_{\bar{p}}(p^*/\bar{p}) = 1$. Together with $\supp(p^*)\subseteq \supp(\bar{p})$ in the first result, the relative variance can be expressed as
\begin{align}
\V_{\bar{p}}(p^*/\bar{p}) & = \E_{\bar{p}}\{(p^*\big/\bar{p})^2\} - 1 \nonumber \\
& = \E_{p^*}(p^* / \bar{p}) - \E_{\bar{p}}(p^*\big/\bar{p}) \nonumber \\
& = \int_\mathcal{X} \frac{p^*(x)}{\bar{p}(x)} p^\ast (x) \, \d x - \int_\mathcal{X} \frac{p^*(x)}{\bar{p}(x)} \bar{p}(x) \, \d x - \int_\mathcal{X} p^*(x) \, \d x  + \int_\mathcal{X} \bar{p}(x) \, \d x  \nonumber\\
& = \int_\mathcal{X} \Big\{\frac{p^*(x)}{\bar{p}(x)} - 1\Big\} p^*(x) \, \d x - \int_\mathcal{X} \Big\{\frac{p^*(x)}{\bar{p}(x)} - 1\Big\} \bar{p}(x) \, \d x  \nonumber\\
& = \int_\mathcal{X} \Big\{\frac{p^*(x)}{\bar{p}(x)} -1 \Big\}\{ p^*(x) - \bar{p}(x) \} \, \d x  \nonumber\\
& = \int_\mathcal{X} \Big\{\frac{p^*(x)}{\bar{p}(x)} -1 \Big\} \{ \surd p^*(x) + \surd \bar{p}(x) \} \{ \surd p^*(x) - \surd \bar{p}(x) \} \, \d x . \label{eq:rel_var_tmp1}
\end{align}
Apply the Cauchy-Schwartz inequality to \eqref{eq:rel_var_tmp1}, the relative variance has the bound
\begin{align}
\V_{\bar{p}}(p^*/\bar{p}) & \leq \Big[\int_\mathcal{X} \!\! \Big\{\frac{p^*(x)}{\bar{p}(x)} -1 \Big\}^2\!\!  \{ \surd p^*(x) + \surd \bar{p}(x) \}^2  \d x \Big]^{\frac12} \! \left[ \int_\mathcal{X} \!\!\{ \surd p^*(x) - \!\surd \bar{p}(x) \}^2  \d x \right]^\frac12 \nonumber \\
& = \Big[ \int_\mathcal{X} \!\! \Big\{\frac{p^*(x)}{\bar{p}(x)} -1 \Big\}^2\!\! \{ \surd p^*(x) + \surd \bar{p}(x) \}^2 \d x \Big]^\frac12 \surd 2\, D_\text{H}(p^\ast, \bar p) . \label{eq:rel_var_tmp2}
\end{align}
Depending on the assumption imposed on $p^*/\bar{p}$, the upper bound of $\V_{\bar{p}}(p^*/\bar{p})$ depends differently on the Hellinger error. 
We note that
\begin{align}
& \hspace{-50pt} \Big[\int_\mathcal{X} \Big\{\frac{p^*(x)}{\bar{p}(x)} -1 \Big\}^2 \{\surd p^*(x) + \surd \bar{p}(x) \}^2 \, \d x \Big]^\frac12\nonumber \\
& \leq \surd 2\Big[ \int_\mathcal{X}  \Big\{\frac{p^*(x)}{\bar{p}(x)} -1 \Big\}^2 p^*(x) \, \d x + \int_\mathcal{X} \Big\{\frac{p^*(x)}{\bar{p}(x)} -1 \Big\}^2 \bar{p}(x) \, \d x  \Big]^\frac12 \nonumber  \\
& = \surd 2 \Big[ \int_\mathcal{X} \Big\{\frac{p^*(x)}{\bar{p}(x)}\Big\}^2  p^*(x) \, \d x  - \int_\mathcal{X} \Big\{\frac{p^*(x)}{\bar{p}(x)}\Big\}^2 \bar{p}(x) \, \d x \Big]^\frac12 \nonumber  \\
& = \surd 2 \left[ \E_{p^*}\{(p^*/\bar{p})^2\} - \E_{\bar{p}}\{(p^*/\bar{p})^2\} \right]^\frac12. \label{eq:rel_var_tmp3}
\end{align}
Note that
\(
\E_{p^*}\{(p^*/\bar{p})^2\} \geq \{\E_{p^*}(p^*/\bar{p})\}^2
\)
and 
\(
\E_{\bar{p}}\{(p^*/\bar{p})^2\} \geq \{\E_{\bar{p}}(p^*/\bar{p})\}^2 = 1
\)
by Jensen's inequality. Together with $\E_{p^*}(p^*/\bar{p}) = \E_{\bar{p}}\{(p^*/\bar{p})^2\}$, the difference on the right hand side of \eqref{eq:rel_var_tmp3} is non-negative. In addition, we have 
\[
\E_{p^*}\{(p^*/\bar{p})^2\} = \E_{\bar{p}}\{(p^*/\bar{p})^3\} = \frac{1}{(\zeta^*)^3}\E_{\bar{p}}\{(\rho^*/\bar{p})^3\} < \infty,
\] 
which can be obtained using  a similar derivation as in the proof of the first result and the assumption that the ratio $\rho^*/\pi_0$ has finite third moment with respect to $\pi_0$. Thus, we have the upper bound 
\[
\V_{\bar{p}}(p^*/\bar{p}) \leq 2 \left[ \E_{p^*}\{(p^*/\bar{p})^2\} - \E_{\bar{p}}\{(p^*/\bar{p})^2\} \right]^\frac12  D_\text{H}(p^\ast, \bar p),
\]
which concludes the second result of this Lemma. 

With a more restrictive assumption $\sup_{x\in\mathcal{X}}p^*(x)/\bar{p}(x) = M_{p^*,\bar{p}}< \infty$, we can also use the identity
\begin{align}
& \hspace{-50pt} \Big[\int_\mathcal{X} \Big\{\frac{p^*(x)}{\bar{p}(x)} -1 \Big\}^2 \{\surd p^*(x) + \surd \bar{p}(x) \}^2 \, \d x \Big]^\frac12 \nonumber \\
& = \Big[\int_\mathcal{X} \Big\{\frac{\surd p^*(x)}{\surd \bar{p}(x)} + 1 \Big\}^4 \{ \surd p^*(x) - \surd \bar{p}(x) \}^2 \, \d x\Big]^\frac12 \nonumber \\
& \leq \Big[ \sup_{x\in\mathcal{X}} \Big\{ \frac{\surd p^*(x)}{\surd \bar{p}(x)} + 1 \Big\}^4  2\, D_\text{H}(p^\ast, \bar p)^2 \Big]^\frac12 \nonumber \\
& =  \surd 2 ( 1 + \surd M_{p^*,\bar{p}} )^2  D_\text{H}(p^\ast, \bar p) \nonumber \\
& \leq  2 \surd 2 \left( 1 + M_{p^*,\bar{p}} \right)  D_\text{H}(p^\ast, \bar p)
\end{align}
Plugging the above identity into \eqref{eq:rel_var_tmp2}, we obtain the upper bound 
\[
\V_{\bar{p}}(p^*/\bar{p}) \leq ( 4 + 4 M_{p^*,\bar{p}} )  D_\text{H}(p^\ast, \bar p)^2.
\]
This concludes the third result of this Lemma. %
\end{proof}

The first condition of Lemma \ref{lemma:rel_var} establishes that the estimator  $\hat{\zeta}_{\bar{p}, N}$ is unbiased and satisfies the central limited theorem, i.e.,  $\surd N\hat{\zeta}_{\bar{p}, N} \overset{i.d.}{\rightarrow} \mathcal{N}\{ \zeta^*, \V_{\bar{p}}(\rho^*/\bar{p}) \}$, where $\overset{i.d.}{\rightarrow}$ denotes convergence in distribution. Since $p^\ast = \rho^\ast/\zeta^\ast$, and thus $\E_{\bar{p}}(p^*/\bar{p}) = 1$, the variance $\V_{\bar{p}}(p^*/\bar{p})$ can be interpreted as the relative variance of the importance ratio $\rho^*/\bar{p}$, i.e., 
\(
(\zeta^\ast)^{-2} \V_{\bar{p}}( \rho^*/\bar{p} )  = \V_{\bar{p}}(p^*/\bar{p}).
\)
In this way, the relative mean square error of the estimator $\hat{\zeta}_{\bar{p}, N}$ is given by
\[
\rmse\big( \hat{\zeta}_{\bar{p}, N}, \zeta^* \big) = N^{-1} \{\zeta^{*}\}^{-2} \V_{\bar{p}}(\rho^*/\bar{p}) = N^{-1} \V_{\bar{p}}(p^*/\bar{p}).
\]
Thus, to guarantee a 
$\rmse\big( \hat{\zeta}_{\bar{p}, N}, \zeta^* \big) \leq \varepsilon$ for some error threshold $\varepsilon > 0$, it is sufficient to choose either
\(
N \geq  C_p\, \varepsilon^{-1}  D_\text{H}(p^\ast, \bar p)
\)
or 
\(
N \geq C_m\, \varepsilon^{-1}  D_\text{H}(p^\ast, \bar p)^2,
\)
depending on whether the assumption in Part 2 or Part 3 of Lemma \ref{lemma:rel_var} holds, respectively.

\subsection{The ratio estimator: from {\it\bfseries a priori} to {\it\bfseries a posteriori} expectations}\label{sec:ratio}

Finally, we want to extend the concept of deep importance sampling just introduced to the case of {\it a posteriori} expectations using the ratio estimator in \eqref{eq:ratio}.
The optimal importance densities for estimating the numerator and the denominator in \eqref{eq:ratio} are $p^* \propto f\,\mathcal{L}^y\pi_0$ and $q^*\propto \mathcal{L}^y \pi_0$, respectively.  We can apply Alg.~\ref{alg:dirt} to construct two composite maps $\mathcal{T}_p^{\lowsup{(L)}}$ and $\mathcal{T}_q^{\lowsup{(L)}}$ to approximately push forward the reference density $\lambda$ to $p^*$ and $q^*$, that is,
\(
\{\mathcal{T}_p^\lowsup{(L)}\}_\sharp \lambda = \bar{p} \approx p^*,
\) 
and 
\(
\{\mathcal{T}_q^\lowsup{(L)}\}_\sharp \lambda  = \bar{q} \approx q^*.
\)
In fact, the optimal importance density for estimating the denominator $Z$ is the normalized posterior, $q^* = \pi^y$. Thus, estimating the denominator here simply reduces to building a normalized posterior approximation. In general, we can choose different numbers of layers for $\mathcal{T}_p^\lowsup{(L)}$ and $\mathcal{T}_q^\lowsup{(L)}$ to adapt to the structures of two optimal densities. 

We are now ready to define the \emph{ratio estimator based on deep importance sampling}
\begin{equation}\label{eq:ratio_deep}
\hat{R}_{\bar{p},\bar{q},N} = \frac{\hat{Q}_{\bar{p}, N} }{\hat{Z}_{\bar{q}, N}}, \;
\hat{Q}_{\bar{p}, N}  = \frac{1}{N} \sum_{i = 1}^{N} w_Q(U_p^i), \;
\hat{Z}_{\bar{q},N}  = \frac{1}{N} \sum_{i = 1}^{N} w_Z(U_q^i), \; U^i_p, U^i_q \sim \lambda,
\end{equation}
where
\begin{align*}
w_Q(U) \!=\! \frac{f\{\mathcal{T}_p^\lowsup{(L)}(U)\!\}\mathcal{L}^y\{\mathcal{T}_p^\lowsup{(L)}(U)\!\}\pi_0\{\mathcal{T}_p^\lowsup{(L)}(U)\!\}}{\bar{p}\{\mathcal{T}_p^\lowsup{(L)}(U)\}}, \; w_Z(U) \!=\! \frac{\mathcal{L}^y\{\mathcal{T}_q^\lowsup{(L)}(U)\!\}\pi_0\{\mathcal{T}_q^\lowsup{(L)}(U)\!\}}{\bar{q}\{\mathcal{T}_q^\lowsup{(L)}(U)\}}.%
\end{align*}
For variance reduction, we consider that each pair of random variables $(U_p^i, U_q^i)$ follows some joint distribution but their marginal laws have the reference density $\lambda$.

To simplify notation, we define random variables $W_Q = w_Q(U_p)$ and $W_Z = w_Z(U_q)$. Under Assumptions \ref{ass:is2} and \ref{ass:is3}, we have $\E(W_Q) = Q$ and $\E(W_Z) = Z$, and thus $\hat{Q}_{\bar{p}, N}$ and $\hat{Z}_{\bar{q}, N}$ are unbiased estimators of $Q$ and $Z$, respectively. However, in general the resulting ratio estimator $\hat{R}_{\bar{p},\bar{q},N}$ is only asymptotically unbiased. In Lemmas \ref{lemma:delta} and \ref{coro:ratio}, we want to characterize the asymptotic behaviour of the relative mean square error of $\hat{R}_{\bar{p},\bar{q},N}$ using its relative deviation from the {\it a posteriori} expectation $R = Q/Z$. We define the relative mean square error of $\hat{R}_{\bar{p},\bar{q},N}$ as
\begin{align}\label{eq:rel_dev}
\Delta_{R,N} = \frac{\hat{R}_{\bar{p},\bar{q},N} - R}{R}  = \frac{\sum_{i = 1}^{N}W_Q^i/Q}{\sum_{i = 1}^{N} W_Z^i/Z } -1,
\end{align}
which is controlled by the laws of $W_Q^1, \ldots, W_Q^N$ and $W_Z^1, \ldots, W_Z^N$.

\begin{remark}\label{remark:taylor}
The following definitions and results are used for showing properties of $\Delta_{R,N}$. 
We introduce the relative derivations of  $\hat{Q}_{\bar{p}, N}$ and $\hat{Z}_{\bar{q}, N}$, which are given by 
\begin{align*}
\Delta_{Q,N} = \frac{\hat{Q}_{\bar{p}, N} - Q}{ Q} \quad \text{and} \quad \Delta_{Z,N} =  \frac{\hat{Z}_{\bar{q}, N} - Z}{Z},
\end{align*}
respectively. Defining random variables $\Theta_Q := W_Q/Q - 1$ and $\Theta_Z := W_Z/Z - 1$, the relative derivations $\Delta_{Q,N}$ and $\Delta_{Z,N}$ can be expressed as 
\begin{equation}\label{eq:rel_dev_sa}
\Delta_{Q,N} = \frac1N\sum_{i = 1}^{N} \Theta_Q^i, \quad \Delta_{Z,N} = \frac1N\sum_{i = 1}^{N} \Theta_Z^i. 
\end{equation}
Note that $\E(\Theta_Q) = \E(\Theta_Z) = 0$ as $\E(W_Q) = Q$ and $\E(W_Z) = Z$. Thus, \(\E(\Delta_{Q,N}) = \E(\Delta_{Z,N} ) = 0 \)
for any sample size $N$. 
The variances and the covariance of $\Theta_Q$ and $\Theta_Z$ can be given as
\begin{equation}\label{eq:rel_dev_2}
\V(\Theta_Q) = \frac{\V(W_Q)}{Q^2}, \;\; \V(\Theta_Z) = \frac{\V(W_Z)}{Z^2}, \;\; \cov(\Theta_Q, \Theta_Z) = \frac{\cov(W_Q, W_Z)}{QZ}.
\end{equation}

The relative deviation $\Delta_{R,N}$ can be expressed as
\[    
\Delta_{R,N} = \frac{\hat{R}_{\bar{p},\bar{q},N} - R}{R} = \frac{Z}{Q} \bigg( \frac{\hat{Q}_{\bar{p}, N} }{\hat{Z}_{\bar{q}, N} } - \frac{Q}{Z} \bigg) = \frac{1 +  \Delta_{Q,N}}{1 +  \Delta_{Z,N}} -1 = \frac{\Delta_{Q,N} - \Delta_{Z,N}}{1 + \Delta_{Z,N}}.
\]
Applying Taylor's theorem, there exist some $s,t \in[0,1]$ such that 
\begin{align}
(1 + \Delta_{Z,N})^{-1} & = 1 - (1 + s\Delta_{Z,N})^{-2} \Delta_{Z,N}, \label{eq:delta_taylor1} \\
(1 + \Delta_{Z,N})^{-1}  & = 1 - \Delta_{Z,N} + (1 + t\Delta_{Z,N})^{-3} \Delta_{Z,N}^2, \label{eq:delta_taylor2}
\end{align}
where $s$ and $t$ depend on $\Delta_{Z,N}$. The term $1 + s \Delta_{Z,N}$ (and similarly $1 + t \Delta_{Z,N}$) satisfies
\[
1 + s \Delta_{Z,N} = (1 - s) +  s (1 + \Delta_{Z,N}) = (1 - s) +  s \hat{Z}_{\bar{q}, N}/ Z > 0 
\]
almost surely, because the estimator $\hat{Z}_{\bar{q}, N}$ is almost surely positive by construction. Thus, the expansions in \eqref{eq:delta_taylor1} and \eqref{eq:delta_taylor2} are not subject to division-by-zero. \textqed
\end{remark}

\begin{lemma}
\label{lemma:delta}
Suppose Assumptions \ref{ass:is1}--\ref{ass:is3} hold and the sequence $\{(W_Q^i, W_Z^i)\}_{i = 1}^N$ is i.i.d., but allowing each pair $(W_Q^i, W_Z^i)$ to be correlated. Then, we have \vspace{-1ex}
\[
\surd N\Delta_{R,N}  \overset{i.d.}{\rightarrow} \mathcal{N}\left\{0, \frac{\V(W_Q)}{Q^2} + \frac{\V(W_Z)}{Z^2} - \frac{2\cov(W_Q. W_Z)}{QZ} \right\}  .  
\]
\end{lemma}  

\begin{proof}
Using the expansion \eqref{eq:delta_taylor1}, we have
\[
\surd N\Delta_{R,N} = \surd N (\Delta_{Q,N} - \Delta_{Z,N}) - \frac{\Delta_{Z,N}}{(1 + s\Delta_{Z,N})^2} \surd N (\Delta_{Q,N} - \Delta_{Z,N}).
\]
Since \(\Delta_{Q,N} {-} \Delta_{Z,N} {=} N^{-1} \sum_{i = 1}^{N}\Theta_Q^i {-} \Theta_Z^i\), we have $\surd N (\Delta_{Q,N} {-} \Delta_{Z,N})$ converges in distribution to $\mathcal{N}\{0, \V( \Theta_Q{-}\Theta_Z )\}$ by the central limit theorem, where $$\V( \Theta_Q{-}\Theta_Z ) = \V( \Theta_Q) {+} \V( \Theta_Z ) {-} 2\cov( \Theta_Q,\Theta_Z ).$$ Since the sequence $\Delta_{Z,N}$ converge to zero in probability as $N$ tends to infinity, i.e., $\Delta_{Z,N} = o_p(1)$, and the sequence $\surd N (\Delta_{Q,N} {-} \Delta_{Z,N})$ is tight, the result follows from Slutsky's theorem and the identities in \eqref{eq:rel_dev_2}.  
\end{proof}

Thus, the ratio estimator in \eqref{eq:ratio_deep} is asymptotically unbiased and converges at the correct rate with respect to the sample size $N$. We also see that by correlating each pair of random variables $(U_p^i, U_q^i)$ in \eqref{eq:ratio_deep}  we can maximize the correlation between $W_Q = w_Q(U_p)$ and $W_Z = w_Z(U_q)$ to minimize the relative variance of the ratio estimator.  For example, if $\lambda$ is a zero mean Gaussian distribution, one can use the antithetic formula
\(
U_p = a U_q + (1-a^2)^{1/2} \epsilon, 
\)
with $\epsilon \sim \lambda$ and some constant $a$ to correlate or anti-correlate the random variables. This way, the marginal distributions of $(U_p, U_q)$ still have the same density $\lambda$, but $U_p$ and $U_q$ are correlated and it is possible to maximize $\cov(W_Q, W_Z)$ as a function of $a$.

To get a more explicit, quantitative result regarding the benefits of the deep importance sampling strategy, in the following lemma we focus only on the case of independent samples $U_p^i,\ U_q^i$, for each $i=1,\ldots,N$.

\begin{lemma}
\label{coro:ratio}
Under the assumptions of Lemma \ref{lemma:delta}, but now assuming furthermore independence of $\{W_Q^i\}_{i = 1}^N$ and $\{W_Z^i\}_{i = 1}^N$, the relative bias of $\hat{R}_{\bar{p},\bar{q},N}$ satisfies 
\[
\frac{N \vert \E(\hat{R}_{\bar{p},\bar{q},N}) - R \vert}{ R} \rightarrow \frac{ \V(W_Z ) }{ Z^2} \quad \text{as} \quad N\rightarrow \infty,
\]
and the relative mean square error of $\hat{R}_{\bar{p},\bar{q},N}$ satisfies
\begin{equation}\label{eq:rmse_hell}
\rmse(\hat{R}_{\bar{p},\bar{q},N}, R) = \bigO \left\{ \frac{C_p  D_{\rm H}(p^\ast, \bar p) + C_q  D_{\rm H}(q^\ast, \bar q)}{N} + \frac{1}{N^2} \right\},
\end{equation}
where 
\[
C_p = 2[\E_{p^*}\{(p^*/\bar{p})^2\} {-} \E_{\bar{p}}\{(p^*/\bar{p})^2\} ]^{1/2}, \quad
C_q = 2[ \E_{q^*}\{(q^*/\bar{q})^2\} {-} \E_{\bar{q}}\{(q^*/\bar{q})^2\} ]^{1/2}.
\]
\end{lemma}

\begin{proof}
Using \eqref{eq:delta_taylor2}, the expected relative deviation $\Delta_{R,N}$ can be expressed as
\begin{align*}
\E(\Delta_{R,N}) 
& = \E(\Delta_{Q,N}) \!-\! \E(\Delta_{Z,N}) \!-\! \E(\Delta_{Q,N}\Delta_{Z,N}) \!+\! \E\Big[\Delta_{Z,N}^2 \Big\{ 1 + \frac{\Delta_{Q,N} - \Delta_{Z,N}}{(1 + t\Delta_{Z,N})^3}\Big\} \Big],
\end{align*}
where $t\in[0,1]$ depending on $\Delta_{Z,N}$. Recall Remark \ref{remark:taylor}, we have $\E(\Delta_{Q,N}) = 0$ and $\E(\Delta_{Z,N}) = 0$ for any given sample size $N$. 
With the additional assumption that the sequences $\{W_Q^i\}_{i = 1}^N$ and $\{W_Z^i\}_{i = 1}^N$ are independent, we have that $\{\Theta_Q^i\}_{i = 1}^N$ and $\{\Theta_Z^i\}_{i = 1}^N$ are also independent. Therefore, we have mutually independent $\Delta_{Q,N}$ and $\Delta_{Z,N}$ for all $N$, and hence $\E(\Delta_{Q,N}\Delta_{Z,N}) = 0$. This leads to
\begin{align*}
\E(\Delta_{R,N}) & = \E\Big[\Delta_{Z,N}^2 \Big\{ 1 + \frac{\Delta_{Q,N} - \Delta_{Z,N}}{(1 + t\Delta_{Z,N})^3}\Big\}\Big].
\end{align*}
Thus, we can introduce a random variable 
\[
B_N = \Delta_{Z,N}^2 \Big\{ 1 + \frac{\Delta_{Q,N} - \Delta_{Z,N}}{(1 + t\Delta_{Z,N})^3}\Big\}
\]
such that the relative bias of $\hat{R}_{\bar{p},\bar{q},N}$ satisfies
\[
\frac{\vert \E(\hat{R}_{\bar{p},\bar{q},N}) - R \vert}{R} = | \E(\Delta_{R,N}) | = | \E(B_N) |.
\]
We want to use Slutsky's theorem to examine the property of the sequence  
\[
\frac{N B_N}{\V(\Theta_Z )} = \Big\{\frac{\surd N  \Delta_{Z,N}}{\surd \V(\Theta_Z )}\Big\}^2 \Big\{ 1 + \frac{\Delta_{Q,N} - \Delta_{Z,N}}{(1 + t\Delta_{Z,N})^3}\Big\}.
\]
Since $\surd N  \Delta_{Z,N}\overset{i.d.}{\rightarrow}\mathcal{N}\{0, \V(\Theta_Z )\}$, the ratio 
\(
\surd N  \Delta_{Z,N} / \surd \V(\Theta_Z ) \overset{i.d.}{\rightarrow}\mathcal{N}(0, 1)
\).
Then, the continuous mapping theorem implies that the term
\(
\{\surd N  \Delta_{Z,N} / \surd \V(\Theta_Z )\}^2 
\)
converges in distribution to the random variable $\xi^2$, where $\xi \sim \mathcal{N}(0, 1)$.

Note that $\xi^2$ follows the chi-squared distribution with one degree of freedom, i.e., $\xi^2 \sim \chi^2_1$, and hence we equivalently have
\(
\{\surd N  \Delta_{Z,N} / \surd \V(\Theta_Z )\}^2 \overset{i.d.}{\rightarrow}\chi^2_1.
\)
Since $\Delta_{Z,N} = o_p(1)$ and $\Delta_{Q,N} = o_p(1)$, we have $N B_N / \V(\Theta_Z)\overset{i.d.}{\rightarrow}\chi^2_1$ by Slutsky's theorem. Thus, by the Portmanteau lemma, we have
\(
\E\{N B_N / \V(\Theta_Z )\} \rightarrow 1 % 
\)
as $N\rightarrow \infty$. Therefore, applying the identities in \eqref{eq:rel_dev_2}, as $N\rightarrow \infty$ the asymptotic behaviour of the relative bias satisfies 
\[
\frac{N \vert \E(\hat{R}_{\bar{p},\bar{q},N}) - R \vert}{R} \rightarrow \frac{\V(W_Z )}{Z^2}.
\]
Thus, the relative bias is asymptomatically $\mathcal{O}(N^{-1})$. 

With the additional assumption that the sequences $\{W_Q^i\}_{i = 1}^N$ and $\{W_Z^i\}_{i = 1}^N$ are also independent, we have  $\cov(W_Q, W_Z) = 0$. Applying the result of Lemma \ref{lemma:delta}, the relative mean square error of $\hat{R}_{\bar{p},\bar{q},N}$ asymptotically follows 
\[
\rmse(\hat{R}_{\bar{p},\bar{q},N}, R) = \bigO \left\{ \frac{\V(W_Q)}{N Q^2} + \frac{\V(W_Z)}{N Z^2} + \frac1{N^2} \right\}.
\]
Since $ \V(W_Q)/Q^2 = \V_{\bar{p}}(p^*/\bar{p})$ and $\V(W_Z)/Z^2 = \V_{\bar{q}}(q^*/\bar{q})$, the rest of the proof directly follows from the second result of Lemma \ref{lemma:rel_var}.
\end{proof}

Lemma \ref{coro:ratio} suggests that the bias is negligible with a large, finite sample size.  
More importantly, the relative mean square error can be greatly reduced by constructing two importance densities $\bar{p}$ and $\bar{q}$ that can accurately approximate the corresponding optimal densities $p^*$ and $q^*$. In theory, the Hellinger errors on the right hand side of \eqref{eq:rmse_hell} can be made to go to zero by increasing the tensor ranks and the number of discretization basis functions, leading to a zero-variance estimator. In comparison, the self-normalized importance sampling method uses identical importance densities for estimating the numerator and the denominator, i.e., $\bar{p} = \bar{q}$, which is always suboptimal at least for one of the terms. This leads to a theoretical lower bound on the estimation variance for finite sample size that cannot be further reduced. 

\section{Application to rare event estimation}\label{sec:rare}

We now use deep importance sampling to devise efficient estimators for {\it a priori} and {\it a posteriori} failure probabilities. The failure function $f(x) = \indi_\mathcal{A} \{ h(x) \}$ defined in \eqref{eq:qoi} will in general have discontinuities at the boundary of the failure set 
\(
\mathcal{X}_F := \{ x\in\mathcal{X} : f(x) = 1 \}. 
\)
When the boundary of $\mathcal{X}_F$ is not aligned with the coordinate axes in the parameter domain, the resulting TT approximation of the optimal importance density may have high ranks. The discontinuities also make it challenging to choose  appropriate bases to efficiently discretize the optimal importance density. 
To alleviate those difficulties and to provide a natural family of intermediate densities $\phi^{(1)}, \ldots, \phi^{(L)}$ for Alg.~\ref{alg:dirt}, we construct a smooth surrogate $g_\gamma(z; \mathcal{A})$ that converges to the indicator function $\indi_\mathcal{A} ( z )$ as $\gamma \to \infty$, that is, $g_\gamma(z; \mathcal{A})$ is continuous for $\gamma < \infty$ and $\lim_{\gamma \rightarrow \infty} g_\gamma(z; \mathcal{A}) = \indi_\mathcal{A}( z )$.

For simplicity, we assume that $\mathcal{A} = [a,b]$ for some $a < b$. In fact, since the indicator function satisfies
\(
\indi_{[a,b]} ( z ) = \indi_{[a,\infty)} ( z )  - \indi_{(b,\infty) } ( z )
\)
and 
\(
\indi_{(-\infty, a]} ( z ) = 1 - \indi_{(a, \infty)} ( z )
\) 
for any finite  $a$ and $b$, without loss of generality, it suffices to consider the case $\mathcal{A} = [a,\infty)$ with $a < \infty$.
Since the weak derivative of $\indi_{[a,\infty)} (z)$ is the Dirac delta $\delta(z-a)$, one can employ a probability density function $p_\gamma(z - a)$ such that $\lim_{\gamma \rightarrow \infty} p_\gamma(z - a)$ has the same distributional properties as $\delta(z-a)$, and then constructs the surrogate function via the corresponding distribution function
\(
g_\gamma(z; [a,\infty)) = \int_{-\infty}^z p_\gamma(z' - a) dz'. %
\) 
In this work, we consider to use the density 
\(
p_\gamma(z - a) = [ 1- \tanh\{(z-a)\gamma/2\}^2 ] \gamma / 4 
\),
which leads to the \emph{sigmoid function}
\begin{equation}\label{eq:sigmoid}
g_\gamma(z; [a,\infty)) = [1+ \exp\{\gamma\,(a - z)\}]^{-1} . 
\end{equation}
This defines a smoothed failure function
\begin{equation*}%
f_\gamma(x) = g_\gamma\{h(x);  [a,\infty)\}. 
\end{equation*}
Instead of directly approximating the optimal importance density $\rho^* = f \pi_0$ for estimating $\E_{\pi_0}\{f(X)\}$, we choose a sufficiently large $\gamma^*$ and approximate the smoothed version $f_{\gamma^*} \pi_0$ to avoid potential discontinuities. This smoothing strategy is also used in  \cite{papaioannou2016sequential,uribe2020cross} for applying gradient-based dimension reduction methods in estimating \textit{a priori} failure probability. 

For the \textit{a priori} rare event,  we can now directly apply Alg.~\ref{alg:dirt} to build a TT approximation of $f_{\gamma^*} \pi_0$. The smoothed failure function $f_{\gamma^*}(x)$ may still have a large  gradient near the boundary of the failure set $\mathcal{X}_F$ and it can concentrate in the tail of $\pi_0$. Thus, we use an increasing sequence of smoothing variables $\gamma_1 < \cdots < \gamma_L = \gamma^*$ to define the unnormalized intermediate densities 
\[
\phi^{(\ell)}(x) = f_{\gamma_\ell}(x) \pi_0(x),  \quad \ell = 1, \ldots, L,
\]
for Alg.~\ref{alg:dirt}. The computed composite map $\mathcal{T}^\lowsup{(L)}$ then provides an importance density $\bar{p} = \{\mathcal{T}^\lowsup{(L)}\}_\sharp \lambda$ that is close to the smoothed optimal importance density $f_{\gamma^*} \pi_0$, and for $\gamma^*$ sufficiently large, also close to the optimal importance density $p^*$. Finally, to estimate the \textit{a priori} rare event probability, we can use the deep importance sampling estimator \eqref{eq:dirt_est2} with $\rho^* = f\pi_0$.

To estimate the \textit{a posteriori} rare event probability, the ratio estimator based on deep importance sampling defined in \eqref{eq:ratio_deep} can be used. Using a tempering approach as in \cite{del2006sequential,gelman1998simulating}, the intermediate densities for the denominator $\E_{\pi_0}\{\mathcal{L}^y\}$ of the ratio estimator in Alg.~\ref{alg:dirt} are chosen to be
\[
\phi^{(\ell)}_d(x) =  \{\mathcal{L}^y(x) \}^{\alpha_\ell} \pi_0(x), \quad 1 \le \ell \le L,
\]
where $\alpha_1 {<} \cdots {<} \alpha_L {=} 1$. For $\alpha_\ell {\ll} 1$, the unnormalized density $\{\mathcal{L}^y(x)\}^{\alpha_\ell} \pi_0(x)$ is significantly less concentrated compared to the unnormalized posterior $\mathcal{L}^y(x) \pi_0(x)$ and can be approximated more easily using TTs.  The resulting composite map $\mathcal{T}_q^\lowsup{(L)}$ defines a density $\bar{q} = \{\mathcal{T}_q^\lowsup{(L)}\}_\sharp\lambda$ that approximates the optimal importance density
\(
q^* \equiv \pi^y.
\)
For the numerator $\E_{\pi_0}\{ f \mathcal{L}^y\}$ of the ratio estimator in \eqref{eq:ratio_deep}, we smooth the failure function, as in the {\it a priori} case, and temper the likelihood to define intermediate densities 
\[
\phi^{(\ell)}_n(x) =  f_{\gamma_\ell}(x) \{ \mathcal{L}^y(x) \}^{\beta_\ell} \pi_0(x), \quad 1 \le \ell \le L,
\]
for Alg.~\ref{alg:dirt}, where $\gamma_1 {<} \cdots {<} \gamma_L {\equiv} \gamma^*$ and $\beta_1 {<} \cdots {<} \beta_L {=} 1$. This leads to the second composite map $\mathcal{T}_p^\lowsup{(L)}$, which defines a density $\bar{p} = \{\mathcal{T}_p^\lowsup{(L)}\}_\sharp\lambda$ approximating the optimal importance density
\(
p^* \propto f \mathcal{L}^y \pi_0 . 
\)
Finally, the two importance densities $\bar{p}$ and $\bar{q}$ can be used in  \eqref{eq:ratio_deep} to evaluate the ratio estimator for the \textit{a posteriori} rare event probability. 

\section{Example 1: susceptible-infectious-removed model}\label{sec:sir}

\subsection{Problem setup}
\begin{figure}[t!]
\centering
\begin{tikzpicture}
\node [rotate=-15] (auMap) {\includegraphics[width=0.3\linewidth]{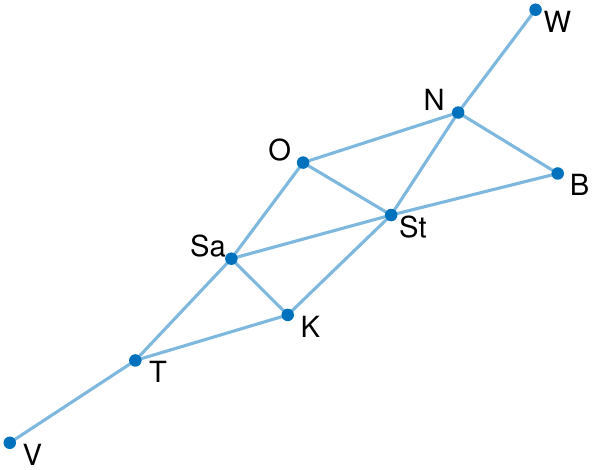}};
\footnotesize
\node[anchor=north west] (auV) at (auMap.north east) {V - };  \node[anchor=north west] (auVt) at ($(auV.north west)+(0.05\linewidth,0)$) {Vorarlberg};
\node[anchor=north west] (auT) at (auV.south west) {T - }; \node[anchor=north west] (auTt) at ($(auT.north west)+(0.05\linewidth,0)$) {Tyrol};
\node[anchor=north west] (auSa) at (auT.south west) {Sa - }; \node[anchor=north west] (auSat) at ($(auSa.north west)+(0.05\linewidth,0)$) {Salzburg};
\node[anchor=north west] (auK) at (auSa.south west) {K - }; \node[anchor=north west] (auKt) at ($(auK.north west)+(0.05\linewidth,0)$) {Carinthia};
\node[anchor=north west] (auSt) at (auK.south west) {St - }; \node[anchor=north west] (auStt) at ($(auSt.north west)+(0.05\linewidth,0)$) {Styria};

\node[anchor=north west] (auO) at ($(auVt.north east)+(0.08\linewidth,0)$) {O - }; \node[anchor=north west] (auOt) at ($(auO.north west)+(0.05\linewidth,0)$) {Upper Austria};
\node[anchor=north west] (auN) at (auO.south west) {N - }; \node[anchor=north west] (auNt) at ($(auN.north west)+(0.05\linewidth,0)$) {Lower Austria};
\node[anchor=north west] (auW) at (auN.south west) {W - }; \node[anchor=north west] (auWt) at ($(auW.north west)+(0.05\linewidth,0)$) {Vienna};
\node[anchor=north west] (auB) at (auW.south west) {B - }; \node[anchor=north west] (auBt) at ($(auB.north west)+(0.05\linewidth,0)$) {Burgenland};
\end{tikzpicture}\vspace{-3em}
\caption{Compartment connectivity graph of the Austrian states.}
\label{fig:austria}
\end{figure}
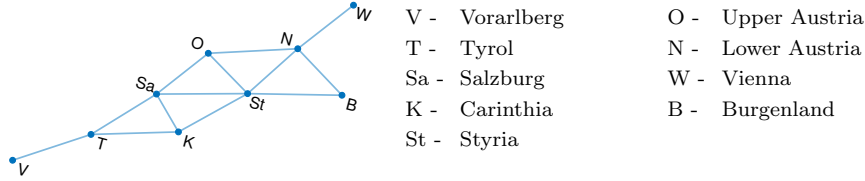

We consider a Bayesian parameter estimation problem for a compartmental
susceptible-infectious-removed model, a simplified version of
the model considered in \cite{DGKP-SEIR-2021}. Given a spatially
dependent demographic model consisting of $K\in\mathbb{N}$ compartments,
we denote the numbers of susceptible, infectious and removed
individuals in the $k$th compartment at a given time $t$ by $S_k(t)$,
$I_k(t)$ and $R_k(t)$, respectively. The interaction among the
individuals within and across the different compartments is modelled by the following
system of differential equations 
\begin{align}
\left\{
\begin{array}{ll}
\displaystyle\frac{dS_k}{dt} &  =  -\theta_k S_k I_k  + {\displaystyle\frac{1}{2}}\sum_{j \in \mathcal{J}_k} (S_{j} - S_k),\\
\displaystyle\frac{dI_k}{dt} & =  \theta_k S_k I_k - \nu_k I_k  + {\displaystyle\frac{1}{2}}\sum_{j \in \mathcal{J}_k} (I_{j}  - I_k) \vphantom{\displaystyle\sum^R_R}, \\
\displaystyle\frac{dR_k}{dt} & =  \nu_k I_k +{\displaystyle\frac{1}{2}}\sum_{j \in \mathcal{J}_k} (R_{j} - R_k), 
\end{array}
\right. \label{eq:sir}
\end{align}
where $\mathcal{J}_k$ is the index set containing all neighbours of
the $k$th compartment. See Fig.~\ref{fig:austria} for an example of
the demographic connectivity graph of the states in Austria. The
system of differential equations is parameterized by $\theta_k\in
\mathbb{R}$ and $\nu_k \in \mathbb{R}$, representing the infection 
and recovery rate in the $k$th compartment, respectively.
We aim to estimate the unknown parameters $x =
(\theta_1,\nu_1, \ldots , \theta_K,\nu_K) \in \R^{2K}$ from noisy
observations of $I_k(t)$ at discrete times. We also aim to estimate the {\it a posteriori} risk, which is the
probability of the number of infected individuals exceeding a chosen
threshold.

\subsection{Experiments on a one-dimensional lattice}

We fist consider a compartment model defined on a one-dimensional
lattice, in which the $k$th compartment is only connected to
compartments with adjacent indices $k-1$ and $k+1$. By changing the
number of compartments, $K$, we can vary the parameter dimension to
test the scalability of deep importance sampling. We impose
periodic boundary conditions, such that $Z_{K+1} = Z_{1}$ and $Z_{0} = Z_{K}$
for $Z \in \{S, I, R\}$. The differential equations in
\eqref{eq:sir} are solved for the time interval $t \in [0,5]$ with
fixed inhomogeneous initial states \( S_k(0) = 99 - K + k, \) \(
I_k(0) = K+1 - k,\) and \( R_k(0) = 0 \) for \(k=1,\ldots,K\). 

For parameter estimation, synthetic observations are generated from noisy measurements of infected population in each of compartments at $6$ equidistant time points,
\begin{equation*}
\by_{k,j} = I_k\Big(\frac{5j}{6} ; \bx_\text{true}\Big) + \eta_{k,j}, \quad \eta_{k,j} \sim \mathcal{N}(0,1), \quad k=1,\ldots,K, \quad j=1,\ldots,6,
\end{equation*}
where the  ``true'' parameter
\(
\bx_\text{true} = [0.1, 1, \ldots, 0.1, 1],
\)
is used for simulating the synthetic observations. This leads to the likelihood function
\begin{equation}
\mathcal{L}^{\by}(\bx) \propto \exp\Big[-\frac12 \sum_{k=1}^{K} \sum_{j=1}^{6} \Big\{I_k\Big(\frac{5j}{6}; \bx\Big) - \by_{k,j}\Big\}^2 \Big].
\end{equation}
The differential equations are solved by the explicit Runge--Kutta method with adaptive time steps that control both absolute and relative errors to be within $10^{-6}$. 
We specify a uniform prior on the domain $[0,2]$ for each of $\theta_k$ and $\nu_k$, which leads to $\pi_0(x) = \prod_{k=1}^{2K}\indi_{[0,2]}(x_k)$.
The {\it a posteriori} risk is defined as the posterior probability of the number of infected individuals in the last compartment at any time $t\in[0,5]$ exceeding a threshold $I_{\max}>0$,
\begin{equation*}
\mathrm{pr}_{\pi^y} \big\{\textstyle \max_{t\in [0,5]} I_K(t; X) > I_{\max}\big\}.
\end{equation*}

To apply deep importance sampling within the ratio estimator
\eqref{eq:ratio_deep},  we use a
sequence of intermediate densities $\phi^{(\ell)}_d(\bx) =
\{\mathcal{L}^{\by}(\bx)\}^{\alpha_{\ell}} \pi_0(\bx)$, $\ell = 1,
\ldots, L$, with tempered likelihood functions to guide
Alg.~\ref{alg:dirt} for the denominator. The tempering parameters start from 
$\alpha_1=10^{-4}$ and are incremented such that $\alpha_{\ell+1} = 10^{1/3} \alpha_{\ell} $ until $\alpha_L=1$.  Thus, $L=13$.
For the numerator of the ratio estimator \eqref{eq:ratio_deep}, we use
another sequence of intermediate densities with the sigmoid smoothing 
\begin{equation}\label{eq:sir-tempering}
\phi^{(\ell)}_n(\bx) {=} \{\mathcal{L}^{\by}(\bx)\}^{\beta_{\ell}} \pi_0(\bx) \big( 1 + \exp\big[\gamma_{\ell} \{I_{\max} - \textstyle\max_{t\in [0,5]} I_K(t; \bx)\}\big] \big)^{-1}, \; \ell {=} 1, \ldots, L.
\end{equation}
Here, we let $\beta_\ell = \alpha_\ell$. The smoothing widths are
chosen such that $\gamma_{\ell} = \beta_{\ell} \gamma^*$, where
$\gamma^*$ will be varied in different experiments.
In the construction of the tensor-train approximations, $\lambda(\bx)$
is a truncated normal reference distribution on $[-3,3]$, and we use piecewise linear basis functions on a uniform grid with
$n_k = n = 17$ points to discretize the densities in each coordinate
direction.

\paragraph{Scalability and accuracy} We vary the compartment number $K = \{3, 5, \ldots, 15\}$ and take
the threshold $I_{\max}=88$. The threshold yields challenging values
of the \emph{a posteriori} risk below $10^{-6}$ for all numbers of compartments in
this set of experiments. We use a sample size of $N=2^{14}$ in the ratio estimator.

We first fix the TT rank to $r_k {=} r {=} 7$ and the smoothing width to $\gamma^*{=}10^4/I_{\max}$.
The Hellinger errors of the deep importance densities, the estimated
{\it a posteriori} risks, and the number of density evaluations needed
are shown in Fig.~\ref{fig:SIR-K}. We observe that the
computational complexity, measured in the number of density
evaluations,  depends linearly on the dimension, while the Hellinger
error increases only moderately for fixed TT ranks, roughly
logarithmically in the probability value itself.

\begin{figure}[t]
\centering
\noindent\hspace{-4pt}\begin{tikzpicture}
\begin{axis}[%
width=0.36\linewidth,
height=0.3\linewidth,
xmode=normal,
xlabel={$d=2K$},
xtick={8,16,24,32},
xmax = 33,
ymin =.05,
title={\footnotesize Hellinger error},
legend style={at={(0.5,.6)},anchor=north west,font=\scriptsize, fill=none},
legend cell align={left},
xlabel shift = -3 pt,
]
\addplot+[error bars/.cd,y dir=both,y explicit] coordinates{
(6 , 0.068532) +- (0, 0.00282957)
(10 , 0.105533) +- (0, 0.00270046)
(12 , 0.126309) +- (0, 0.00675502)
(16 , 0.162541) +- (0, 0.00741734)
(24, 0.228121) +- (0, 0.0174061 )
(32, 0.273218) +- (0, 0.0140547 )
}; \addlegendentry{$D_\text{H}({\bar q},\pi^{\by})$};
\addplot+[error bars/.cd,y dir=both,y explicit] coordinates{
(6 , 0.166238) +- (0, 0.0160873)
(10 , 0.209704) +- (0, 0.0142827)
(12 , 0.212639) +- (0, 0.0091534)
(16 , 0.251123) +- (0, 0.0299916)
(24 , 0.297766) +- (0, 0.0239314)
(32 , 0.337396) +- (0, 0.0297823)
}; \addlegendentry{$D_\text{H}({\bar p},\phi^{(L)}_n)$};
\addplot+[mark=diamond*,error bars/.cd,y dir=both,y explicit,x dir=both,x explicit] coordinates{
(6 , 0.185489) +- (0, 0.0134669)
(10 , 0.218396) +- (0, 0.0148001)
(12 , 0.225906) +- (0, 0.0085992)
(16 , 0.262419) +- (0, 0.0294716)
(24 , 0.305968) +- (0, 0.0221383)
(32 , 0.344568) +- (0, 0.0267337)
}; \addlegendentry{$D_\text{H}({\bar p},p^*)$};
\end{axis}
\end{tikzpicture}
\hspace{-10pt}
\begin{tikzpicture}
\begin{axis}[%
width=0.36\linewidth,
height=0.3\linewidth,
ymode=log,
xlabel={$d=2K$},
xtick={8,16,24,32},
xmax = 33,
title={\hspace{-12pt} \small $\mathrm{pr}_{\pi^y} \big\{ \max_{t\in [0,5]} I_K(t; X) > I_{\max}\big\}$ \hspace{-12pt}},
xlabel shift = -3 pt,
]
\addplot+[error bars/.cd,y dir=both,y explicit] coordinates{
(6 , 1.646e-10) +- (0, 9.214e-13)
(10 , 6.973e-09) +- (0, 5.238e-11)
(12 , 1.459e-08) +- (0, 1.272e-10)
(16 , 4.578e-08) +- (0, 5.753e-10)
(24 , 2.085e-07) +- (0, 2.126e-09)
(32 , 5.315e-07) +- (0, 6.291e-09)
};
\end{axis}
\end{tikzpicture}
\hspace{-14pt}
\begin{tikzpicture}
\begin{axis}[%
width=0.36\linewidth,
height=0.3\linewidth,
xlabel={$d=2K$},
xtick={8,16,24,32},
xticklabels={8,16,24,32},
xmax = 33,
title={\footnotesize $N_{\mbox{total}} \times 10^{-3}$},
xlabel shift = -3 pt,
]
\addplot+[] coordinates{
(6 , 3570 * 13/1e3 + 16384/1e3)
(10 , 6902 * 13/1e3 + 16384/1e3)
(12 , 8568 * 13/1e3 + 16384/1e3)
(16 , 11900* 13/1e3 + 16384/1e3)
(24 , 18564* 13/1e3 + 16384/1e3)
(32 , 25228* 13/1e3 + 16384/1e3)
};
\end{axis}
\end{tikzpicture}\vspace{0pt}
\caption{Hellinger errors in the densities (left), estimated \emph{a
posteriori} risk
(middle) and total number of function evaluations in
Alg.~\ref{alg:dirt} (right) for different
numbers of compartments $K$ in Example 1. In all figures, points denote average
values, and error bars denote one standard deviation over $10$ runs.}
\label{fig:SIR-K}
\end{figure}
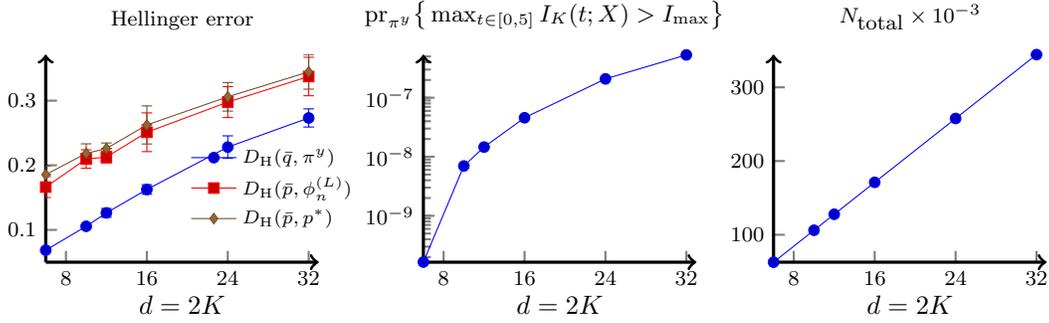

\begin{figure}[t!]
\centering
\noindent\hspace{-4pt}\begin{tikzpicture}
\begin{axis}[%
width=0.36\linewidth,
height=0.3\linewidth,
xlabel=$r$,
title={\footnotesize Hellinger error},
legend style={at={(0.45,1.15)},anchor=north west,font=\scriptsize, fill=none},
legend cell align={left},
xlabel shift = -3 pt,
]
\addplot+[error bars/.cd,y dir=both,y explicit] coordinates{
(3 , 0.245839) +- (0, 0.0150954)
(5 , 0.143838) +- (0, 0.0084124)
(7 , 0.105533) +- (0, 0.0027004)
(9 , 0.093929) +- (0, 0.0041339)
(11, 0.089929) +- (0, 0.0024738)
}; \addlegendentry{$D_\text{H}({\bar q}, \pi^y)$};
\addplot+[error bars/.cd,y dir=both,y explicit] coordinates{
(3 , 0.302884) +- (0, 0.0460232)
(5 , 0.236022) +- (0, 0.0216654)
(7 , 0.209704) +- (0, 0.0142827)
(9 , 0.201707) +- (0, 0.0099414)
(11, 0.199793) +- (0, 0.0220769)
}; \addlegendentry{$D_\text{H}({\bar p}, \phi^{(L)}_n)$};
\addplot+[mark=diamond*,error bars/.cd,y dir=both,y explicit] coordinates{
(3 , 0.311617) +- (0, 0.0431148)
(5 , 0.243763) +- (0, 0.0147339)
(7 , 0.218396) +- (0, 0.0148001)
(9 , 0.214556) +- (0, 0.0089023)
(11, 0.212959) +- (0, 0.0195311)
}; \addlegendentry{$D_\text{H}({\bar p},p^*)$}; 
\end{axis}
\end{tikzpicture}
\hspace{-3pt}
\begin{tikzpicture}
\begin{axis}[%
width=0.36\linewidth,
height=0.3\linewidth,
xlabel=$r$,
title={\footnotesize $N_{\mbox{total}} \times 10^{-3}$},
xlabel shift = -3 pt,
]
\addplot+[] coordinates{
(3 , 1326 * 13 / 1e3 + 16384/1e3)
(5 , 3570 * 13 / 1e3 + 16384/1e3)
(7 , 6902 * 13 / 1e3 + 16384/1e3)
(9 , 11322* 13 / 1e3 + 16384/1e3)
(11, 16830* 13 / 1e3 + 16384/1e3)
};
\end{axis}
\end{tikzpicture}
\hspace{-3pt}
\begin{tikzpicture}
\begin{axis}[%
width=0.36\linewidth,
height=0.3\linewidth,
xlabel={$\smash{\gamma^* \times I_{\max}}$\vphantom{$r$}},
title={\footnotesize Hellinger error},
xmode=log,
ymin=0,ymax=0.7,
legend style={at={(0.4,1.1)},anchor=north west,font=\scriptsize, fill=none},
legend cell align={left},
xlabel shift = -3 pt,
]
\pgfplotsset{cycle list shift=1}
\addplot+[error bars/.cd,y dir=both,y explicit] coordinates{
(1e2, 0.112321) +- (0, 0.00598099)
(2e2, 0.130502) +- (0, 0.00589768)
(3e2, 0.135972) +- (0, 0.00808893)
(5e2, 0.141737) +- (0, 0.00747825)
(1e3, 0.145806) +- (0, 0.007384  )
(3e3, 0.152224) +- (0, 0.0104433 )
(1e4, 0.209704) +- (0, 0.0142827 )
(3e4, 0.26124 ) +- (0, 0.0292032 )
(1e5, 0.307803) +- (0, 0.0377633 )
(3e5, 0.365372) +- (0, 0.0738628 )
}; \addlegendentry{$D_\text{H}({\bar p}, \phi^{(L)}_n)$};
\addplot+[mark=diamond*,error bars/.cd,y dir=both,y explicit] coordinates{
(1e2, 0.950962) +- (0, 0.0024106)
(2e2, 0.697789) +- (0, 0.0061091)
(3e2, 0.543592) +- (0, 0.0058371)
(5e2, 0.412327) +- (0, 0.0074049)
(1e3, 0.291526) +- (0, 0.0061930)
(3e3, 0.211598) +- (0, 0.0071605)
(1e4, 0.218396) +- (0, 0.0148001)
(3e4, 0.263672) +- (0, 0.0290178)
(1e5, 0.308484) +- (0, 0.0376666)
(3e5, 0.365573) +- (0, 0.0738692)
}; \addlegendentry{$D_\text{H}({\bar p}, p^*)$};
\end{axis}
\end{tikzpicture}\vspace{0pt}
\caption{Hellinger errors in the densities (left) and total number of
function evaluations in Alg.~\ref{alg:dirt} (middle) for different TT ranks
$r$, as well as Hellinger errors for different smoothing widths
$\gamma^*$ (right) in Example 1.}
\label{fig:SIR-rank-gamma}
\end{figure}
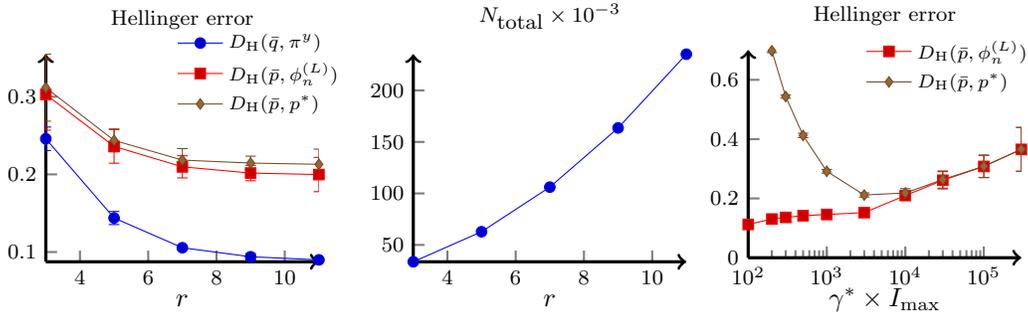

Then, we fix the number of compartments to $K = 5$, and investigate
the impact of the TT rank~$r$ and the smoothing width $\gamma^*$
on the accuracy of deep importance sampling. 
Firstly, we set  $\gamma^*=10^4/I_{\max}$ and vary $r$. As shown in
Fig.~\ref{fig:SIR-rank-gamma}, the
errors in all approximate densities decay with $r$ until the
discretization error is reached, whereas the number of function evaluations in
Alg.~\ref{alg:dirt} appears to depend quadratically on~$r$. 
Secondly, we fix the TT rank to $r=7$ and vary the smoothing
width $\gamma^*$. As shown in the right plot of
Fig.~\ref{fig:SIR-rank-gamma}, the error in approximating the smoothed
optimal importance density depends monotonically on $\gamma^*$. This
is expected, since a larger $\gamma^*$ leads to a less smooth final
biasing density
$\phi^{(L)}_n(\bx)$ that is more difficult to approximate for Alg.~\ref{alg:dirt}. In
contrast, the Hellinger distance of the approximation to the true
optimal biasing density $p^*(\bx)$ grows strongly as $\gamma^*$
decreases. The optimal value of
$\gamma^*$ is therefore an intermediate one, achieved for this example
between $10^3/I_{\max}$ and $10^4/I_{\max}$.

\paragraph{Variance reduction via sample correlation} 
To confirm the variance reduction suggested by Lemma~\ref{lemma:delta}
we let $K = 5$, $I_{\max} = 88$, $\gamma^* = 3000/I_{\max}$ and $r=7$. We
consider positively correlated seed samples $U_p=U_q \sim \lambda$
with $a=1$, uncorrelated samples $U_p \sim\lambda$ and
$U_q\sim\lambda$ with $a=0$, and negatively correlated samples with
$a=-2/3$ and produce $20$ batches of ratio estimators with $N =
2^{12}$ samples each.
The relative standard deviations of the estimated {\it a posteriori} risk are
1.2e-2,
1.4e-2 and
2.4e-2
for positively correlated, uncorrelated, and negatively correlated
samples, respectively. Thus, the error is indeed reduced by making the
correlation $\mathrm{corr}(\Theta_Q. \Theta_Z)>0$ positive, which
confirms the result of Lemma~\ref{lemma:delta}.   

\paragraph{Comparison with cross entropy}
To benchmark our deep importance sampling approach we compare it to the
cross entropy method of \cite{botev2008efficient}. 
We vary the number of compartments, $K$, and compare the estimation
accuracy of the cross entropy method and deep importance sampling.  
The cross entropy method has difficulties in estimating the rather
small {\it a posteriori} risk in the above experiments. Therefore we
reduce the threshold to $I_{\max}{=}80$ in this experiment. 
For cross entropy, we use an importance density with a
mixture of $4$ Gaussian distributions. For our deep importance
sampling method we use a TT rank of $r{=}7$ and a smoothing width of 
$\gamma^* {=} 3000/I_{\max}$. 
The estimated risks and their empirical standard deviations, which are computed over 10 replications, are summarized in  Table~\ref{tab:SIR-CE}, together with $\mbox{N}/\mbox{ESS}$ estimates, where ESS denotes the effective sample size (see \cite{evans1995methods,kong1992note} for details). 
We observe that the accuracy of the cross entropy method deteriorates drastically with the dimension, making $K{=}3$ compartments intractable even with a million samples per iteration. Increasing the number of mixture distributions gives similar results, while reducing it makes the results worse.
In comparison, Alg.~\ref{alg:dirt} is able to estimate the probability with less than $1\%$ relative error in a fraction of the number of samples needed for the cross entropy method.

\begin{table}[t]
\centering{\footnotesize
\caption{Average value of the \emph{a
posteriori} risk in Example 1 over $10$ runs, $\pm$ 1 standard deviation, using the cross entropy method and
deep importance sampling, as well as $\mbox{N}/\mbox{ESS}$ (in brackets).}
\label{tab:SIR-CE}
\noindent
\begin{tabular}{c|cc|c}
& \multicolumn{2}{c|}{Cross entropy} & Deep importance
            sampling%
\\
$K$ & $N=10^5$                & $N=10^6$                & $N \approx 2\cdot 10^4$ \\\hline
1   & 4.731e-5 $\pm$ 9.58e-8  & 4.724e-5 $\pm$ 3.92e-8  & 4.728e-5 $\pm$ 9.22e-8 \\
    & (1.753 $\pm$ 3e-3)      & (1.721 $\pm$ 5.4e-2)    & (1.096 $\pm$ 3e-3) \\\hline
2   & 5.914e-4 $\pm$ 9.11e-4  & 6.202e-5 $\pm$ 3.53e-5  & 8.270e-5 $\pm$ 2.03e-7 \\
    & (3689 $\pm$ 5197)       & (89259 $\pm$ 2e+5)      & (1.113 $\pm$ 6e-3) \\\hline
3   & ---                     & ---                     & 3.378e-4 $\pm$ 1.10e-6 \\
    &                         &                         & (1.150 $\pm$ 5.5e-2)
\end{tabular}}
\end{table}

\subsection{Experiments on the Austria model}
Finally, we consider a more realistic setting where the model has
$K=9$ compartments following the Austrian state adjacency map shown in Fig.~\ref{fig:austria}. The initial condition is given as
$S_1(0)=99$, $I_1(0)=1$, $R_1(0)=0$ (in Vorarlberg), and $S_k(0)=100$,
$I_k(0)=R_k(0)=0$ elsewhere. 
We estimate parameters $x\in \R^{18}$ from synthetic noisy observation of $\{I_k(5j / 12; x_\mathrm{true})\}$, $k=1,\ldots,9$, $j=1,\ldots,12$, with the same ``true'' parameter and likelihood model specified in the first experiment.
The risk is defined as the number of infected individuals in
Burgenland, indexed by $k = 9$, at any time $t\in[0,5]$ exceeding a
threshold $I_{\max} = 69$. This value of $I_{\max}$ corresponds to a
dimensionless ratio of the highest number of hospitalizations
($20000$) and the expected initial number of infected individuals
($290$) employed in the modeling of lockdown strategies in England by
\cite{DGKP-SEIR-2021}. 

To apply Alg.~\ref{alg:dirt}, we use the intermediate densities
defined above, with different starting tempering parameters
$\alpha_1{=}10^{-5}$ and $L {=}16$.
The final smoothing width is fixed to
$\gamma^* {=} 10^4/I_{\max}$. The TT ranks in each layer are adaptively chosen, with the maximum rank set
to $r{=}7$. To estimate the performance we use again $10$ replicated
experiments. 
The performance is as in the
previous experiments. Both importance densities used in the ratio
estimator can be accurately estimated using the layered transport
maps. For the denominator and the numerator, the estimated Hellinger
errors of the approximate importance densities are
$D_\mathrm{H}(\pi^y,\bar q){=} 0.135 \pm 0.005$ and
$D_\mathrm{H}(p^*,\bar p){=} 0.282 \pm 0.008$, respectively, using a
total of $314371 \pm 11727$ density evaluations. The estimated  {\it a
posteriori} risk is $ 4.370 \times 10^{-10}$ with estimated standard
derivation $1.05\times 10^{-12}$. 

\section{Example 2: contaminant transport in groundwater}\label{sec:elliptic}

\subsection{Problem setup}
\begin{figure}[t]
\centering
\noindent
\includegraphics[height=0.18\linewidth]{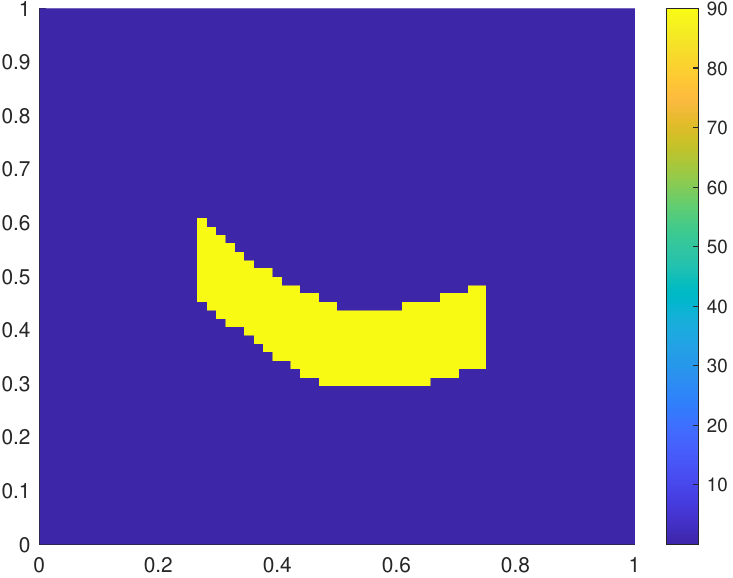}
\includegraphics[height=0.18\linewidth]{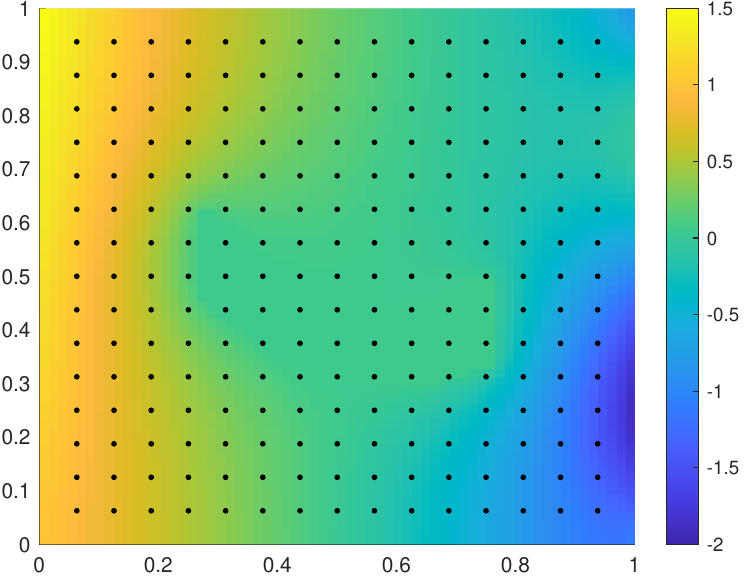}
\includegraphics[height=0.18\linewidth]{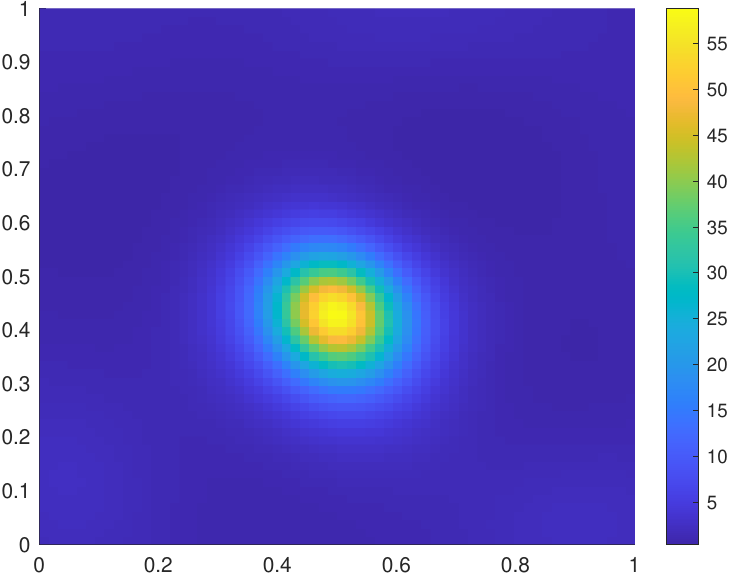}
\includegraphics[height=0.18\linewidth]{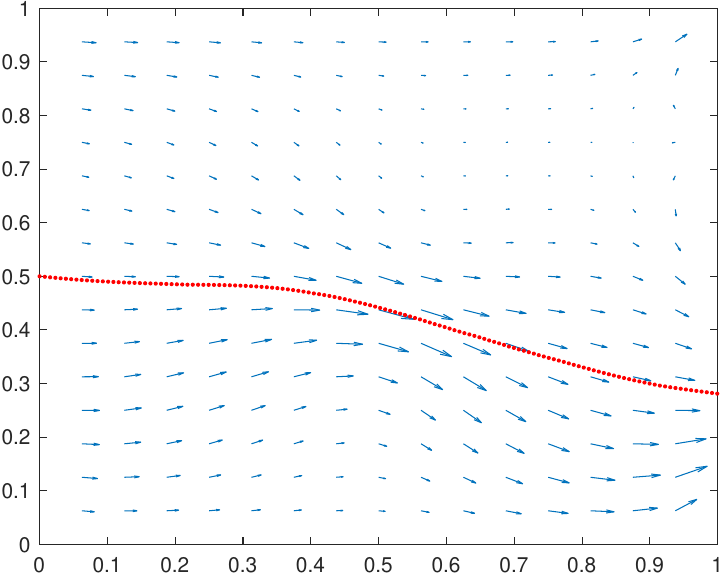}
\\
\noindent
\includegraphics[height=0.18\linewidth]{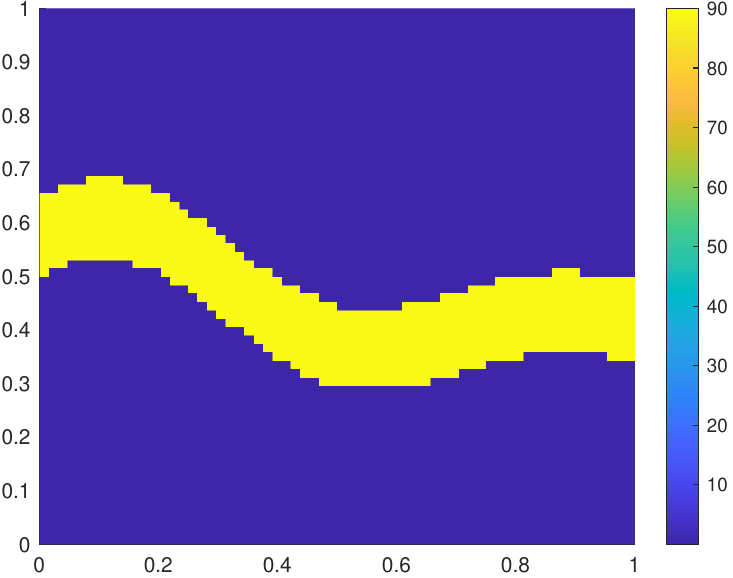}
\includegraphics[height=0.18\linewidth]{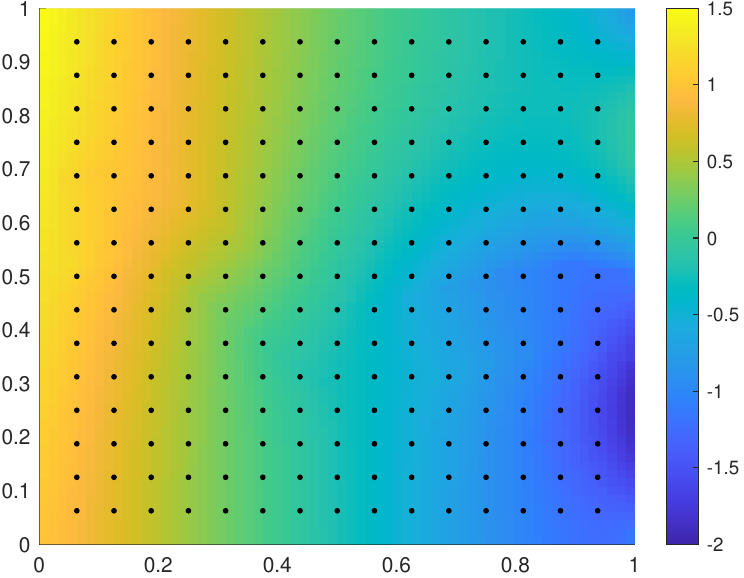}
\includegraphics[height=0.18\linewidth]{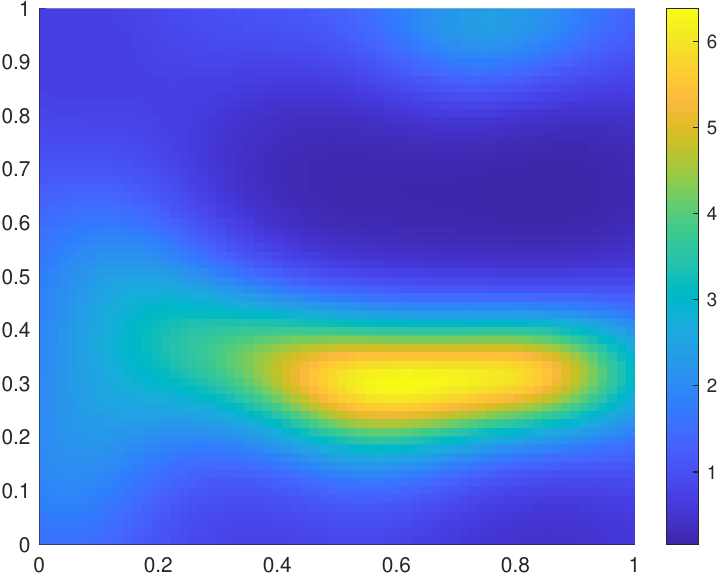}
\includegraphics[height=0.18\linewidth]{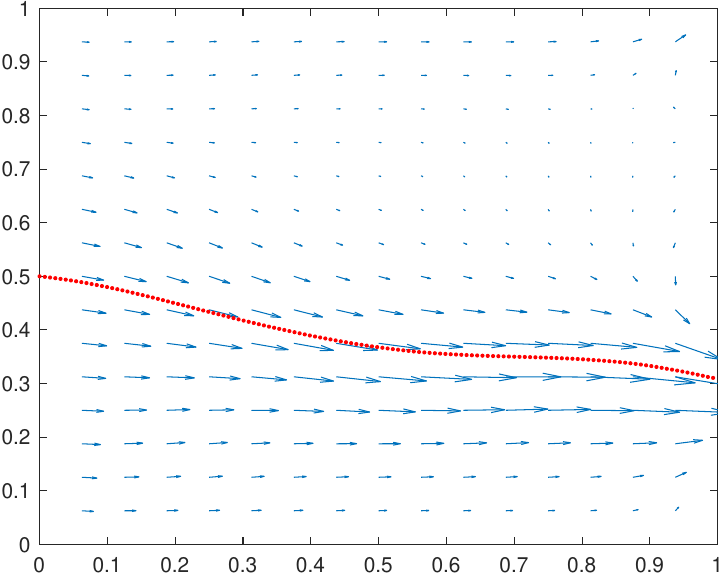}\vspace{6pt}

\caption{Two groundwater experiments with a low diffusivity barrier (top row) and a high diffusivity channel (bottom row). First column: true diffusivity fields $\kappa$. Second column: water tables $u$ generated by true $\kappa$ with observation locations (black dots). Third column: maximum {\it a posteriori} estimates of diffusivity fields $\kappa$. Fourth column: flow fields (blue arrows) and particle trajectories (red) computed using $\kappa$ in the third column.  The maximum {\it a posteriori} particle breakthrough times of the top row and the bottom row are $\tau = 0.1886$ and $\tau =  0.0929$, respectively.\label{fig:diff-channel}
}
\end{figure}

We aim to estimate the risk of contaminant transport in a steady-state
groundwater system; see \cite{cliffe2000} and the references therein.
Here, the physical system is driven by some
unknown random diffusivity field $\kappa(s,X)$ that cannot be directly
observed, where $s\in D=[0,1]^2$ is the spatial coordinate in the
physical domain $D$ and $X$, taking values in $\R^d$, is some parameter
describing the randomness of the diffusivity. The observable state of
the system is the water table $u(s,X)$, which is a function that
satisfies the partial differential equation 
\begin{align}
-\nabla \cdot \{\kappa(s,X) \nabla u(s,X)\} & = 0, \quad s\in(0,1)^2 ,  \label{eq:pde}
\end{align}
with Dirichlet boundary conditions
$u|_{s_1=0} = 1+s_2/2$ and $u|_{s_1=1} = -\sin(2\pi s_2)-1$
imposed horizontally and no-flux boundary conditions $\partial u / \partial s_2|_{s_2=0} = \partial u / \partial s_2|_{s_2=1} = 0$ imposed vertically.
The Dirichlet boundary conditions
generate an  inhomogeneous horizontal Darcy flow field
$\kappa(s,X) \nabla u(s,X)$. 
Figure \ref{fig:diff-channel} shows examples of flow fields and water
tables generated by two different synthetic  diffusivity fields.
Contaminant particles released at a fixed location $s^0 = (0, 0.5)$
on the left boundary are transported by the flow field
according to the advection equation 
\begin{equation}\label{eq:flux-ode}
\frac{ds(t,X)}{dt} = \kappa(s,X) \nabla u(s,X), \quad s(0,X) = s^0,
\end{equation}
to arrive at the right boundary after some time $\tau$. The particle
paths are shown in the right column of Fig.~\ref{fig:diff-channel}.
The risk in this scenario is defined as the
probability, subject to the random diffusivity $\kappa(s,X)$,
that the breakthrough time of contaminant particles, denoted by $\tau(X)$,
is below some threshold $\tau_*$. This way, the {\it a priori} risk and the {\it a posteriori} risk are given by
$\mathrm{pr}_{\pi_0} \{\tau(X) < \tau_*\} $ and
$\mathrm{pr}_{\pi^y} \{\tau(X) < \tau_*\} $, respectively.

For each realization of $X$, we first apply the Galerkin method with
continuous, bilinear finite elements to numerically solve \eqref{eq:pde}. The finite element solution $u_h$ is computed on a uniform
rectangular grid on $D$ with a mesh size $h = 1/64$ along each of the
coordinates of $D$.
The inhomogeneous horizontal Darcy flow field
$\kappa(s,X) \nabla u_h(s,X)$ is also calculated in the same finite
element space. Then, the advection equation \eqref{eq:flux-ode} with the discretized flow field is numerically solved by an explicit Runge-Kutta method with adaptive time stepping (\texttt{ode45} in MATLAB).

We assume that the logarithm of the diffusivity field follows a zero mean Gaussian process with the Mat\'ern covariance function
\[
C(s,t) = \frac{2^{1-\nu}}{\Gamma(\nu)}\left(\sqrt{2\nu} \frac{\|s-t\|_2}{\ell}\right)^{\nu} K_{\nu}\left(\sqrt{2\nu}\frac{\|s-t\|_2}{\ell}\right), \quad s,t \in D,
\]
where $\nu$ is the smoothness parameter, and $\ell$ is the correlation length.
This definition includes the Gaussian covariance function as the limit $\nu \rightarrow \infty$. 
Using the Karhunen-L\'oeve (KL)
expansion, $\log\kappa(s,X)$ can be approximated by the finite representation\vspace{-1ex}
\[
\log\kappa(s,X) \approx \sum_{k = 1}^{d} X_k \surd{\lambda_k} \psi_k(s) ,
\]
where $\{\psi_k(s), \lambda_k\}$ is the $k$th
eigenpair of the covariance operator in the descending
order of eigenvalues and each random coefficient $X_k$ follows a standard normal prior.

To setup the observation model, we measure the water table $u(s,X)$ at
$m=15 \times 15$ locations defined as the vertices of a uniform Cartesian grid
on $D=[0,1]^2$ with grid size $1/(\surd m+1)$, as shown in
Fig.~\ref{fig:diff-channel}. Measurements are 
corrupted by i.i.d. Gaussian noise.
For a realization of $X$, the observables are simulated numerically as the average
of $u_h(s,X)$ over subdomains $D_{i} \subset D$, $i=1,\ldots,m$,
around the measurement locations.
In our experiments, each  $D_{i}$ is a square with side length
$2/(\surd m +1)$ centred at the $i$th location.
This leads to the parameter-to-observable map
\begin{equation}\label{eq:pde-data}
y_i = Q_i(x) + \eta_i, \quad Q_{i}(x) = \frac{1}{|D_{i}|}\int_{D_i} u_h(s,x) ds, \quad \eta_i \sim \mathcal{N}(0,\sigma_n^2)
\end{equation}
for $i=1,\ldots,m,$ where $\sigma_n^2$ is the variance of the measurement noise.

\subsection{{\it\bfseries A posteriori} risk versus {\it\bfseries a priori} risk}

A common practice in the literature is to estimate the
{\it a priori} risk by only considering the randomness induced by the
prior of $\kappa(s,X)$; see
\cite{peherstorfer2016multifidelity,uribe2020cross} and references
therein for examples. As shown in Fig.~\ref{fig:diff-channel},
depending on the structure of the true diffusivity field, the contaminant
breakthrough time can change due to localized
changes that are difficult to detect. Thus, it is
critical to also assess the {\it a posteriori} risk, where the
uncertainty due to the unobserved diffusivity
field $\kappa(s,X)$ can be better characterized by conditioning on observations of the water table.  

\begin{figure}[ht]
\centering
\begin{tikzpicture}
\begin{axis}[%
name=AxisTauCDFZoom,
width=0.40\linewidth,
height=0.3\linewidth,
xlabel={$\log_{10}\tau$},
xmode=normal,
ymode=log,
legend style={at={(0.99,0.01)},anchor=south east,font=\scriptsize, fill=none},
legend cell align={left},
xmin=-1.1,xmax=-0.9,
]
\addplot+[no marks,line width=1pt] table[header=false,x index=0,y index=1] {prior-logtau-cdf.dat}; \addlegendentry{prior};
\addplot+[no marks,line width=1pt] table[header=false,x index=0,y index=1] {island-logtau-cdf.dat}; \addlegendentry{barrier};
\addplot+[no marks,line width=1pt] table[header=false,x index=0,y index=1] {channel-logtau-cdf.dat}; \addlegendentry{channel};
\end{axis}
\begin{axis}[%
name=AxisTauCDF,
at={($(AxisTauCDFZoom.north east)+(0.15\linewidth,0)$)},anchor=north west,
width=0.40\linewidth,
height=0.3\linewidth,
xlabel={$\log_{10}\tau$},
xmode=normal,
legend style={at={(0.99,0.01)},anchor=south east,font=\scriptsize, fill=none},
legend cell align={left},
xmin=-1.3,xmax=1.3,
]
\addplot+[no marks,line width=1pt] table[header=false,x index=0,y index=1] {prior-logtau-cdf.dat}; \addlegendentry{prior};
\addplot+[no marks,line width=1pt] table[header=false,x index=0,y index=1] {island-logtau-cdf.dat}; \addlegendentry{barrier};
\addplot+[no marks,line width=1pt] table[header=false,x index=0,y index=1] {channel-logtau-cdf.dat}; \addlegendentry{channel};

\draw[solid,line width=0.5pt] (axis cs:-1.1, 0) -- (axis cs: -1.1, 0.1) -- (axis cs: -0.9, 0.1) -- (axis cs: -0.9, 0);
\node[inner sep=0pt] (AxisTauCDFp1) at (axis cs: -1.1, 0.05) {};
\end{axis}

\draw[dashed,line width=0.5pt] (AxisTauCDFp1) -- ($(AxisTauCDFZoom.north east)+(0.005\linewidth,0)$);
\draw[dashed,line width=0.5pt] (AxisTauCDFp1) -- ($(AxisTauCDFZoom.south east)+(0.005\linewidth,0)$);
\end{tikzpicture}
\caption{Empirical cumulative density function of the breakthrough time, $\log_{10}\tau$, computed using $2^{17}$ samples from prior and posteriors conditional on two data sets shown in Figure~\ref{fig:diff-channel}. Left: zoom around the threshold $\tau_*=0.1$. \label{fig:diff-hist}}
\end{figure}
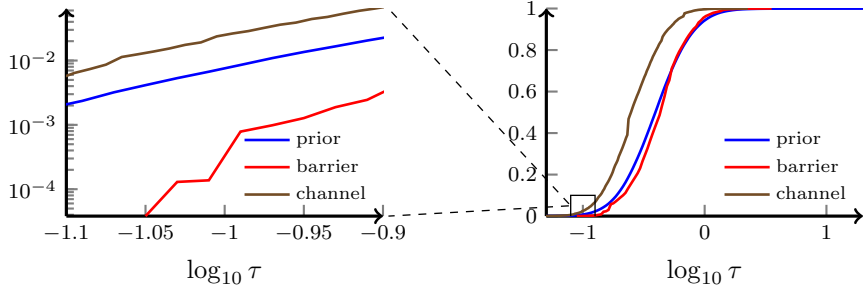

We first demonstrate the critical importance of computing the {\it a posteriori} risk rather than {\it a priori} risk in this example. We consider an experiment with the prior correlation length  $\ell = 1/\sqrt{50}$, prior smoothness $\nu=\infty$, $d=20$ in the KL expansion, and a breakthrough time threshold $\tau_* = 0.1$. Without any observed data, the {\it a priori} risk computes to $6.3 \times 10^{-3} \pm 6.4 \times 10^{-4}$. Next, we generate the solution $u$ from one of the ``truth'' coefficients depicted in Fig.~\ref{fig:diff-channel} (left), and observe the solution at $15 \times 15$ equispaced spatial points with a zero-mean normal noise with variance $3\times 10^{-2}$. Using the data generated from the diffusivity field with a low-diffusivity barrier in the top of Fig.~\ref{fig:diff-channel}, the {\it a posteriori} risk is $9.4 \times 10^{-4} \pm 1.3 \times 10^{-4}.$ In comparison, using the data generated from the diffusivity field with a high-diffusivity channel in the bottom of Fig.~\ref{fig:diff-channel}, the {\it a posteriori} risk is $2.8 \times 10^{-2} \pm 0.2 \times 10^{-2}$, which is an order of magnitude higher. 
In addition, Fig.~\ref{fig:diff-hist} shows cumulative density functions of the breakthrough time in the logarithmic scale. We observe that the law of breakthrough time significantly changes with observed data. 
In summary, the critical change of risk cannot be detected by computing the {\it a priori} risk in this example. Using observed data is essential to reliably estimate the risk of a groundwater system.

\subsection{Additional experiments of \textit{A priori} rare events and comparison with cross entropy}\label{sec:elliptic_prior}

Here, we provide additional experiments for changing the risk threshold $\tau_*$, the smoothing width $\gamma^*$, and the dimension of the truncated random field $d$. We also compare deep importance sampling with the cross entropy method. 
To enable computation using cross entropy and in a wide range of parameters, we change the smoothness parameter to $\nu=2$, noise variance to $\sigma_n^2=10^{-2}$ and the correlation length to $\ell=1$. With these parameters, the truncated representation of the dimension $d=25$ can capture 99.99\% of the variance of the KL expansion. We also change the Dirichlet boundary conditions to $u|_{s_1=0} = 1$ and $u|_{s_1=1}=0$.

To apply Alg.~\ref{alg:dirt}, we compute the approximation of the
optimal importance density with TT rank $r=9$,  intermediate
parameters $\beta_1=10^{-2}$, $\beta_{\ell+1}=\surd 10\,\beta_\ell$,
$\gamma_{\ell} = \beta_{\ell} \, \gamma^*$, and two options for the
smoothing width $\gamma^* = 30/\tau_*$ and $\gamma^* =100/\tau_*$. 
A total of $N_{total}=159885$ density evaluations is required to
construct the composite map. 
In the left plot of Fig.~\ref{fig:diff-tau-pi0}, we plot the Hellinger
errors of the deep importance densities versus the risk thresholds
$\tau_*$. We consider two Hellinger distances: the distance
$D_\text{H}(\bar p,p^*)$  between the computed deep
importance density $\bar p$ and the optimal importance density $p^*$, as well as the
distance $D_\text{H}\{\bar p,\phi^{(L)}\}$ between the deep
importance density $\bar p$ and the final layer of smoothed importance
densities $\phi^{(L)}$. As for \emph{a posteriori} risk estimation above, smaller
$\tau_*$ values lead to smaller probabilities of a particle
traversing the channel in a time below $\tau_*$, which
increases the difficulty to approximate the importance densities and is
reflected in higher Hellinger errors.

\begin{figure}[t]
\centering
\noindent
\begin{tikzpicture}
\begin{axis}[%
width=0.36\linewidth,
height=0.3\linewidth,
xlabel=$\tau_*$,
xmode=log,
xtick={0.01,0.02,0.04},
xticklabels={0.01,0.02,0.04},
title={\footnotesize Hellinger error},
ymin=0.1,
xmin=0.007,xmax=0.07,
legend style={at={(1.05,0.7)},anchor=north west,font=\scriptsize, fill=none},
legend cell align={left},
xlabel shift = -3 pt,
]
\pgfplotsset{cycle list shift=1};
\addplot+[error bars/.cd,y dir=both,y explicit] coordinates{
(0.0075, 0.212315) +- (0, 0.0189593 )
(0.01  , 0.169991) +- (0, 0.0166548 )
(0.02  , 0.151287) +- (0, 0.00804559)
(0.03  , 0.146631) +- (0, 0.00552464)
(0.06  , 0.146417) +- (0, 0.0094997 )
}; \addlegendentry{$D_\text{H}({\bar p}, \phi^{(L)})$};
\addplot+[mark=diamond*,error bars/.cd,y dir=both,y explicit] coordinates{
(0.0075, 0.728284) +- (0, 0.00758564)
(0.01  , 0.501731) +- (0, 0.00723943)
(0.02  , 0.320676) +- (0, 0.00720007)
(0.03  , 0.298811) +- (0, 0.00611428)
(0.06  , 0.277218) +- (0, 0.00753178)
};  \addlegendentry{$D_\text{H}({\bar p}, p^*)$};
\pgfplotsset{cycle list shift=-1};
\addplot+[dashed,line width=1pt,error bars/.cd,y dir=both,y explicit] coordinates{
(0.0075, 0.418353) +- (0, 0.101856 )
(0.01  , 0.272637) +- (0, 0.0403478)
(0.02  , 0.201659) +- (0, 0.0172452)
(0.03  , 0.199848) +- (0, 0.0163322)
(0.06  , 0.196489) +- (0, 0.0222851)
}; %
\addplot+[dashed,line width=1pt,mark=diamond*,error bars/.cd,y dir=both,y explicit] coordinates{
(0.0075, 0.562493) +- (0, 0.0790708)
(0.01  , 0.380715) +- (0, 0.0287436)
(0.02  , 0.261109) +- (0, 0.0140687)
(0.03  , 0.250737) +- (0, 0.0136895)
(0.06  , 0.238857) +- (0, 0.0168114)
}; % 
\node[anchor=east] at (axis cs: 0.075, 0.7) {\small solid: $\gamma^*{=}30/\tau_*$};
\node[anchor=east] at (axis cs: 0.085, 0.6) {\small dashed: $\gamma^*{=}100/\tau_*$};
\end{axis}
\end{tikzpicture}
\begin{tikzpicture}
\begin{axis}[%
width=0.36\linewidth,
height=0.3\linewidth,
ymode=log,
xlabel=$\tau_*$,
xmode=log,
xtick={0.01,0.02,0.04},
xticklabels={0.01,0.02,0.04},
title={\footnotesize $\mathrm{pr}_{\pi_0} \{\tau(X) < \tau_*\}$},
xmin=0.007,xmax=0.07,
xlabel shift = -3 pt,
]
\addplot+[error bars/.cd,y dir=both,y explicit] coordinates{
(0.0075, 6.930e-11) +- (0, 8.100e-13)
(0.01  , 4.658e-08) +- (0, 6.905e-10)
(0.02  , 1.076e-05) +- (0, 4.846e-08)
(0.03  , 7.500e-05) +- (0, 3.535e-07)
(0.06  , 1.173e-03) +- (0, 9.023e-06)
};
\end{axis}
\end{tikzpicture}
\caption{Hellinger errors in computed deep importance  densities for
\emph{a priori} risk estimation for different breakthrough time
thresholds $\tau_*$ and smoothing widths $\gamma_*$ (left), as well as
the associated values of the \emph{a priori} risk (right). Points
denote average, and error bars denote one standard deviation over $10$ runs.}
\label{fig:diff-tau-pi0}
\end{figure}
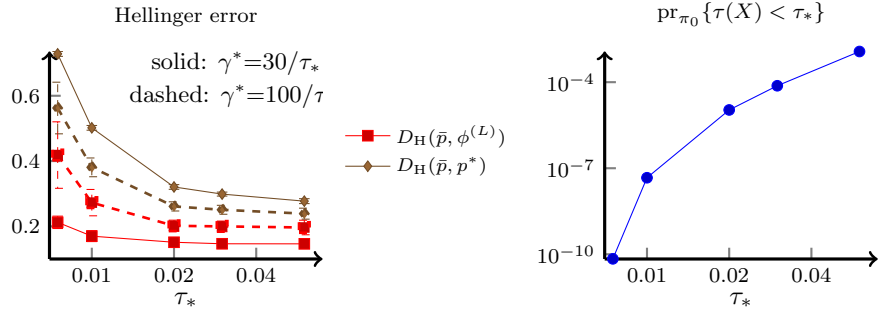

\begin{figure}[t]
\centering
\begin{tikzpicture}
\begin{axis}[%
width=0.36\linewidth,
height=0.3\linewidth,
xlabel=$d$,
xmode=log,
xtick={5,10,20},
xticklabels={5,10,20},
title={\footnotesize rel.std{$\big[\mathrm{pr}_{\pi_0} \{\tau(X) < \tau_*\}\big]$}},
ymode=log,
ymin=3e-3,
xmax=26,
legend style={at={(0.35,0.65)},anchor=north west,font=\small, fill=none},
legend cell align={left},
xlabel shift = -3 pt,
]
\addplot+[] coordinates{
(5,  9.6785e-03)
(8,  1.0462e-02)
(10, 3.9981e-03)
(12, 8.5926e-03)
(14, 4.0448e-03)
(16, 4.8613e-03)
(18, 4.6491e-03)
(20, 5.3415e-03)
(25, 4.6825e-03)
};\addlegendentry{deep importance sampling};
\addplot+[] coordinates{
(5,  4.8933e-01)
(8,  2.3386e-01)
(10, 4.2076e-01)
(12, 6.2513e-01)
(14, 4.1182e-01)
(16, 1.3821e-01)
(18, 2.3786e-01)
};\addlegendentry{cross entropy};
\end{axis}
\end{tikzpicture}
\quad
\begin{tikzpicture}
\begin{axis}[%
width=0.36\linewidth,
height=0.3\linewidth,
xlabel=$d$,
xmode=log,
xtick={5,10,20},
xticklabels={5,10,20},
ymode=log,
title={\footnotesize $N_{total} \times 10^{-3}$},
ymin=10,ymax=3e3,
xmax=26,
legend style={at={(0.05,1.1)},anchor=north west,font=\small, fill=none},
legend cell align={left},
xlabel shift = -3 pt,
]
\addplot+[] coordinates{
(5 ,       13685/1e3 )
(8 ,       26180/1e3 )
(10,       34510/1e3 )
(12,       42840/1e3 )
(14,       51170/1e3 )
(16,       59500/1e3 )
(18,       67830/1e3 )
(20,       76160/1e3 )
(25,       96985/1e3 )
}; %
\addplot+[error bars/.cd,y dir=both,y explicit] coordinates{
(5 , 20*8.888889e+00) +- (0, 20*2.420973e+00)
(8 , 20*7.600000e+00) +- (0, 20*1.646545e+00)
(10, 20*7.444444e+00) +- (0, 20*1.810463e+00)
(12, 20*17          ) +- (0, 20*1.009950e+01)
(14, 40*9           ) +- (0, 40*4.570436e+00)
(16, 60*8.333333e+00) +- (0, 60*2.121320e+00)
(18, 80*7.666667e+00) +- (0, 80*1.861899e+00)
}; %
\end{axis}
\end{tikzpicture}
\caption{Relative standard deviation of the \emph{a priori} risk,
estimated using $10$ runs, comparing deep importance sampling and
the cross entropy method (left), as well as total number of density
evaluations used in each case (right).}
\label{fig:diff-CE}
\end{figure}
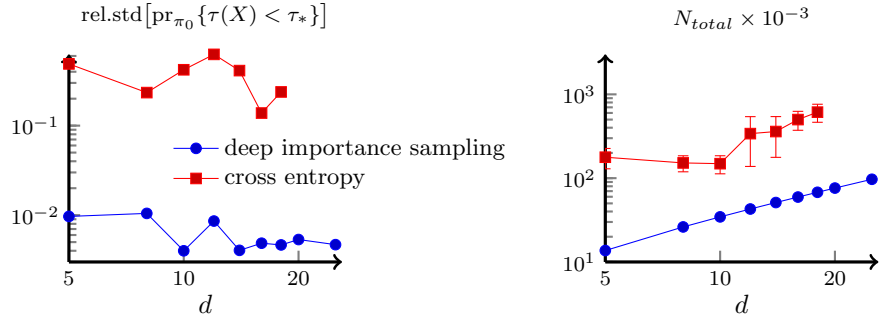

In Fig.~\ref{fig:diff-CE}, we compare deep importance sampling to
the cross entropy method of \cite{botev2008efficient}, for the risk
threshold fixed at $\tau_*=0.03$. Here, the cross entropy method uses
only one single Gaussian density, which is the best we were able to fit,
while the smoothing width $\gamma^*=100/\tau_*$ is used to build
intermediate densities for deep importance sampling in Alg.~\ref{alg:dirt}. 
We run $10$ replicas of each method to estimate relative standard
deviations of the risk probabilities, which are shown in the left plot
of Fig.~\ref{fig:diff-CE}. In the right plot of
Fig.~\ref{fig:diff-CE}, we also show the total number of density
evaluations used by each of the methods. 
In this example, the cross entropy method is able to compute
qualitatively correct risk estimates in higher dimensions, albeit
requiring a larger number of density evaluations (starting from
$2\times 10^5$ samples per iteration at $d=5$, growing to $6 \times 10^5$ for
$d=18$). However, for $d\geq20$, the cross entropy method is unable
to converge, even using $N = 10^6$ samples per iteration. In
comparison, the number of density evaluations in deep importance
sampling demonstrates a linear scaling in the dimension and
nearly constant errors that are about two orders of
magnitude below those of the cross entropy method. Moreover, this is achieved
using one order of magnitude fewer density evaluations compared
to the cross entropy method.

\subsection{Additional experiments of \textit{a posteriori} rare events}

Here, we provide additional experiments for changing the
risk threshold $\tau_*$, the smoothing width $\gamma^*$, and the dimension of the truncated random field $d$.
In this set of experiments, we use the model setup in Section \ref{sec:elliptic_prior}, a sample size of $N = 2^{15}$, a fixed TT rank $7$, and intermediate parameters
$\beta_1=10^{-3}$, $\beta_{\ell+1}=\surd 10 \, \beta_{\ell}$,
$\gamma_{\ell} = \beta_{\ell} \, \gamma^*$ and $\alpha_\ell =
\beta_\ell$.

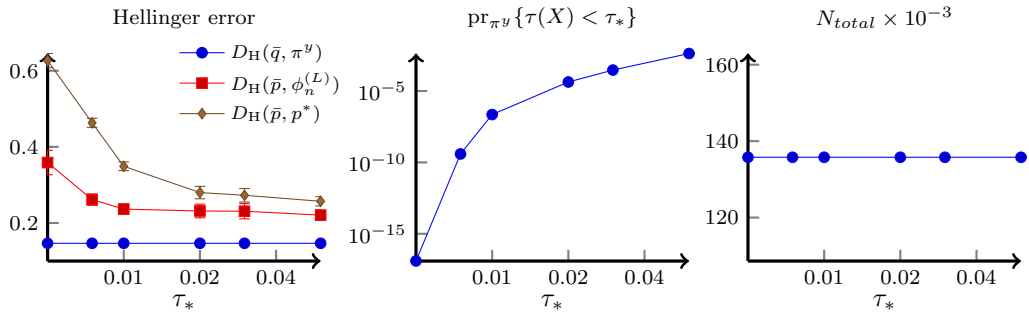
\begin{figure}[t]
\centering
\noindent

\hspace{-12pt}
\begin{tikzpicture}
\begin{axis}[%
width=0.36\linewidth,
height=0.3\linewidth,
xlabel=$\tau_*$,
xmode=log,
xtick={0.01,0.02,0.04},
xticklabels={0.01,0.02,0.04},
title={\footnotesize Hellinger error},
ymin=0.1,
legend style={at={(0.45,1.1)},anchor=north west,font=\scriptsize, fill=none},
legend cell align={left},
xlabel shift = -3 pt,
]
\addplot+[error bars/.cd,y dir=both,y explicit] coordinates{
(0.005, 0.146653) +- (0, 0.00249813)
(0.0075, 0.146653) +- (0, 0.00249813)
(0.01 , 0.146653) +- (0, 0.00249813)
(0.02 , 0.146653) +- (0, 0.00249813)
(0.03 , 0.146653) +- (0, 0.00249813)
(0.06 , 0.146653) +- (0, 0.00249813)
};\addlegendentry{$D_\text{H}({\bar q}, \pi^y)$};
\addplot+[error bars/.cd,y dir=both,y explicit] coordinates{
(0.005, 0.35875 ) +- (0, 0.0319949)
(0.0075, 0.261455) +- (0, 0.0148388)
(0.01 , 0.236518) +- (0, 0.0129287)
(0.02 , 0.231569) +- (0, 0.0180045)
(0.03 , 0.230856) +- (0, 0.0200671)
(0.06 , 0.220675) +- (0, 0.0137024)
}; \addlegendentry{$D_\text{H}({\bar p}, \phi^{(L)}_n)$};
\addplot+[mark=diamond*,error bars/.cd,y dir=both,y explicit] coordinates{
(0.005, 0.626685) +- (0, 0.0186588)
(0.0075, 0.463051) +- (0, 0.0118729)
(0.01 , 0.348964) +- (0, 0.0114133)
(0.02 , 0.279922) +- (0, 0.0162924)
(0.03 , 0.272785) +- (0, 0.0177496)
(0.06 , 0.256994) +- (0, 0.0122116)
}; \addlegendentry{$D_\text{H}({\bar p}, p^*)$};
\end{axis}
\end{tikzpicture}
\hspace{-12pt}
\begin{tikzpicture}
\begin{axis}[%
width=0.36\linewidth,
height=0.3\linewidth,
ymode=log,
xlabel=$\tau_*$,
xmode=log,
xtick={0.01,0.02,0.04},
xticklabels={0.01,0.02,0.04},
title={\footnotesize $\mathrm{pr}_{\pi^y} \{\tau(X) < \tau_*\}$},
xlabel shift = -3 pt,
]
\addplot+[error bars/.cd,y dir=both,y explicit] coordinates{
(0.005, 1.252e-17) +- (0, 2.266e-19)
(0.0075, 3.897e-10) +- (0, 2.739e-12)
(0.01 , 2.244e-07) +- (0, 1.232e-09)
(0.02 , 4.267e-05) +- (0, 1.835e-07)
(0.03 , 2.929e-04) +- (0, 1.698e-06)
(0.06 , 4.229e-03) +- (0, 2.754e-05)
};
\end{axis}
\end{tikzpicture}
\hspace{-6pt}
\begin{tikzpicture}
\begin{axis}[%
width=0.36\linewidth,
height=0.3\linewidth,
xlabel=$\tau_*$,
xmode=log,
xtick={0.01,0.02,0.04},
xticklabels={0.01,0.02,0.04},
title={\footnotesize $N_{total} \times 10^{-3}$},
xlabel shift = -3 pt,
]
\addplot+[] coordinates{
(0.005, 19397*7/1e3)
(0.0075, 19397*7/1e3)
(0.01 , 19397*7/1e3)
(0.02 , 19397*7/1e3)
(0.03 , 19397*7/1e3)
(0.06 , 19397*7/1e3)
};
\end{axis}
\end{tikzpicture}
\caption{Hellinger errors in the density approximations (left),
\emph{a posteriori} breakthrough probabilities (middle) and total
number of density evaluations in Alg.~\ref{alg:dirt} (right) for
different  breakthrough thresholds $\tau_*$. Points denote
average, and error bars denote one standard deviation over 10 runs.}
\label{fig:diff-tau-pi}
\end{figure}

We first vary $\tau_*$ and calculate the \emph{a posteriori} risks of
breakthrough using a default smoothing width $\gamma^* =
100/\tau_*$. The results are shown in  Fig.~\ref{fig:diff-tau-pi}
together with Hellinger errors of the importance density functions used
in the ratio estimator, as well as the total number of density
evaluations needed in Alg.~\ref{alg:dirt}.
As above, we consider three Hellinger errors: the distance
$D_\text{H}(\bar p,p^*)$ between the
computed deep importance density and the optimal importance density
for the numerator of the ratio estimator, the
distance $D_\text{H}\{\bar p,\phi^{(L)}_n\}$
between the deep importance density and the final layer of
smoothed importance densities for the numerator of the ratio
estimator, as well as the distance $D_\text{H}(\bar q,
\pi^y)$ between the computed deep importance density and the
optimal importance density for the
denominator of the ratio estimator. 
Clearly smaller $\tau_*$ lead to smaller probabilities of a particle
travelling through the channel in a time below $\tau_*$. Consequently,
the optimal importance density of the numerator becomes harder to
approximate when $\tau_*$ decreases. Correspondingly, we observe that the
Hellinger errors  $D_\text{H}(\bar p,p^*)$ and $D_\text{H}\{\bar
p,\phi^{(L)}_n\}$ increase as $\tau_*$ decreases.
Nevertheless, even extremely small probabilities (below $10^{-10}$) can be estimated accurately.
For this set of
experiments, the number of function evaluations stays constant, as the
same parameters are used in  Alg.~\ref{alg:dirt}.

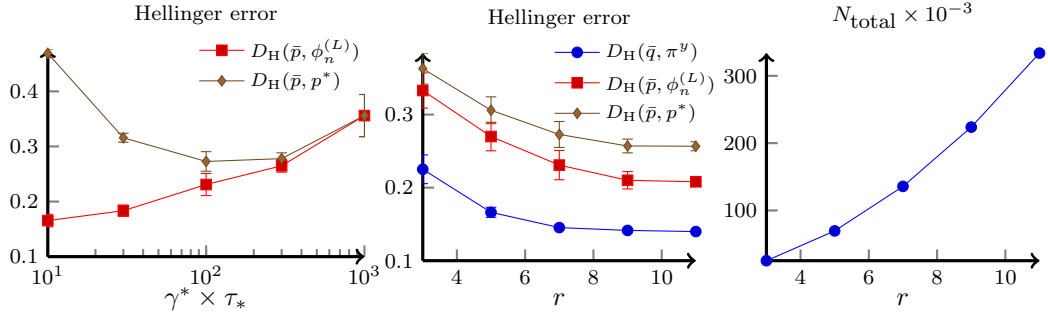
\begin{figure}[htb]
\centering
\noindent
\hspace{-12pt}
\begin{tikzpicture}
\begin{axis}[%
width=0.4\linewidth,
height=0.3\linewidth,
xlabel={$\smash{\gamma^* \times \tau_*}$\vphantom{$r$}},
title={\footnotesize Hellinger error},
xmode=log,
ymin=0.1,
legend style={at={(0.45,1.1)},anchor=north west,font=\scriptsize, fill=none},
legend cell align={left},
xlabel shift = -3 pt,
]
\pgfplotsset{cycle list shift=1}
\addplot+[error bars/.cd,y dir=both,y explicit] coordinates{
(10  , 0.165406) +- (0, 0.0113562)
(30  , 0.183397) +- (0, 0.0102603)
(100 , 0.230856) +- (0, 0.0200671)
(300 , 0.26488 ) +- (0, 0.0115195)
(1000, 0.356024) +- (0, 0.0383195)
};  \addlegendentry{$D_\text{H}({\bar p}, \phi^{(L)}_n)$};
\addplot+[mark=diamond*,error bars/.cd,y dir=both,y explicit] coordinates{
(10  , 0.468973) +- (0, 0.0078211)
(30  , 0.315821) +- (0, 0.0081621)
(100 , 0.272785) +- (0, 0.0177496)
(300 , 0.277892) +- (0, 0.0105321)
(1000, 0.356024) +- (0, 0.0383195)
}; \addlegendentry{$D_\text{H}({\bar p}, p^*)$};
\end{axis}
\end{tikzpicture}
\hspace{-12pt}
\begin{tikzpicture}
\begin{axis}[%
width=0.36\linewidth,
height=0.3\linewidth,
xlabel=$r$,
title={\footnotesize Hellinger error},
ymin=0.1,
legend style={at={(0.45,1.1)},anchor=north west,font=\scriptsize, fill=none},
legend cell align={left},
xlabel shift = -3 pt,
]
\addplot+[error bars/.cd,y dir=both,y explicit] coordinates{
(3 , 0.225128) +- (0, 0.0196729)
(5 , 0.166223) +- (0, 0.0068727)
(7 , 0.145324) +- (0, 0.0015884)
(9 , 0.141436) +- (0, 0.0012566)
(11, 0.139891) +- (0, 0.0005376)
};  \addlegendentry{$D_\text{H}({\bar q}, \pi^y)$};
\addplot+[error bars/.cd,y dir=both,y explicit] coordinates{
(3 , 0.332955) +- (0, 0.0240978)
(5 , 0.269948) +- (0, 0.0195202)
(7 , 0.230856) +- (0, 0.0200671)
(9 , 0.210078) +- (0, 0.0119062)
(11, 0.208033) +- (0, 0.0066880)
};  \addlegendentry{$D_\text{H}({\bar p}, \phi^{(L)}_n)$};
\addplot+[mark=diamond*,error bars/.cd,y dir=both,y explicit] coordinates{
(3 , 0.36292 ) +- (0, 0.0212058)
(5 , 0.305921) +- (0, 0.0183389)
(7 , 0.272785) +- (0, 0.0177496)
(9 , 0.257029) +- (0, 0.009425 )
(11, 0.256648) +- (0, 0.006154 )
};  \addlegendentry{$D_\text{H}({\bar p}, p^*)$};
\end{axis}
\end{tikzpicture}
\hspace{-12pt}
\begin{tikzpicture}
\begin{axis}[%
width=0.36\linewidth,
height=0.3\linewidth,
xlabel=$r$,
title={\footnotesize $N_{\mbox{total}} \times 10^{-3}$},
xlabel shift = -3 pt,
]
\addplot+[] coordinates{
(3 , 3621 * 7 / 1e3)
(5 , 9945 * 7 / 1e3)
(7 , 19397* 7 / 1e3)
(9 , 31977* 7 / 1e3)
(11, 47685* 7 / 1e3)
};
\end{axis}
\end{tikzpicture}
\hspace{-24pt}
\caption{Hellinger errors for \emph{a posteriori} risk
estimation for different smoothing widths $\gamma^*$ (left) and
TT ranks $r$ (middle) where $\tau^* = 0.03$. The right figure
shows the total number of density evaluations in Alg.~\ref{alg:dirt}
as a function of the rank $r$ for $\gamma^* = 100/\tau_*$.
Points denote averages, and error bars denote one standard deviation over 10 runs.}
\label{fig:diff-gamma}
\end{figure}

Then, with a fixed risk threshold $\tau_* = 0.03$, we study the
behaviour of Alg.~\ref{alg:dirt} when the smoothing width $\gamma^*$ and
the TT ranks are changed. The left plot of Fig.~\ref{fig:diff-gamma}
shows the resulting Hellinger errors for approximating the optimal
importance density of the numerator as a function of $\gamma^*$.
The tensor-train approximation error increases with increasing
$\gamma^*$ due to the loss of smoothness, while the bias error between
the exact optimal importance density $p^*$ and the smoothed density
$\phi^{(L)}_n$ decreases. Thus, there is an optimal $\gamma^*$ to
obtain the most accurate
approximation of the optimal importance function $p^*(x)$, where the
two error contributions balance. Regarding the dependency on the
maximum rank $r$, for a fixed $\gamma^* = 100/\tau_*$ we observe that
all Hellinger errors decay with $r$ until the discretisation error is
reached, whereas the number of function evaluations in
Alg.~\ref{alg:dirt} appears to depend quadratically on~$r$,
as expected from the number of degrees of freedom in the tensor-train decomposition.

\begin{figure}[t!]
\centering
\noindent\hspace{-10pt}\begin{tikzpicture}
\begin{axis}[%
width=0.36\linewidth,
height=0.3\linewidth,
xmode=normal,
xlabel=$d$,
xtick={5,10,15,20,25},
xmax=26,
ymax = 0.3,
title={\footnotesize Hellinger error},
legend style={at={(1.05,0.95)},anchor=north west,font=\scriptsize, fill=none},
legend cell align={left},
legend columns = 1,
xlabel shift = -3 pt,
]
\addplot+[error bars/.cd,y dir=both,y explicit] coordinates{
(5 , 0.072367) +- (0, 0.011863  )
(8 , 0.106417) +- (0, 0.0149113 )
(10, 0.113878) +- (0, 0.00973978)
(12, 0.127492) +- (0, 0.0278621 )
(14, 0.126792) +- (0, 0.0108118 )
(16, 0.136096) +- (0, 0.00870636)
(18, 0.140094) +- (0, 0.0103826 )
(20, 0.152496) +- (0, 0.00894435)
(25, 0.161171) +- (0, 0.019422  )
}; \addlegendentry{$D_\text{H}({\bar q}, \pi^y)$};
\addplot+[error bars/.cd,y dir=both,y explicit] coordinates{
(5 , 0.098456) +- (0, 0.0163434)
(8 , 0.155611) +- (0, 0.0268825)
(10, 0.149242) +- (0, 0.0230091)
(12, 0.174712) +- (0, 0.0254705)
(14, 0.19394 ) +- (0, 0.0540873)
(16, 0.179719) +- (0, 0.0116873)
(18, 0.189252) +- (0, 0.0139171)
(20, 0.187082) +- (0, 0.0114766)
(25, 0.219721) +- (0, 0.0265351)
}; \addlegendentry{$D_\text{H}({\bar p}, \phi^{(L)}_n)$};
\addplot+[mark=diamond*,error bars/.cd,y dir=both,y explicit] coordinates{
(5 , 0.144074) +- (0, 0.0107972)
(8 , 0.189964) +- (0, 0.0215804)
(10, 0.183588) +- (0, 0.0206449)
(12, 0.205526) +- (0, 0.0212741)
(14, 0.223339) +- (0, 0.0479875)
(16, 0.20945 ) +- (0, 0.0101347)
(18, 0.218594) +- (0, 0.0117049)
(20, 0.215491) +- (0, 0.0100126)
(25, 0.245414) +- (0, 0.0228814)
}; \addlegendentry{$D_\text{H}({\bar p}, p^*)$};
\end{axis}
\end{tikzpicture}
\quad
\begin{tikzpicture}
\begin{axis}[%
width=0.36\linewidth,
height=0.3\linewidth,
xlabel=$d$,
xtick={5,10,15,20,25},
xmode=normal,
xmax=26,
ymode=normal,
ytick={100,200,300,400},
title={\footnotesize $N_{total} \times 10^{-3}$},
legend style={at={(0.05,0.95)},anchor=north west,font=\scriptsize, fill=none},
legend cell align={left},
xlabel shift = -3 pt,
]
\addplot+[error bars/.cd,y dir=both,y explicit] coordinates{
(5 , 4242.44*7/1e3) +- (0, 455.766*7/1e3)
(8 , 9990.87*7/1e3) +- (0, 767.529*7/1e3)
(10, 13237.1*7/1e3) +- (0, 1329.68*7/1e3)
(12, 17089.9*7/1e3) +- (0, 1409.55*7/1e3)
(14, 20990.4*7/1e3) +- (0, 1663.77*7/1e3)
(16, 21712.8*7/1e3) +- (0, 1501.94*7/1e3)
(18, 23283.5*7/1e3) +- (0, 2753.51*7/1e3)
(20, 25663.6*7/1e3) +- (0, 1805.85*7/1e3)
(25, 29954  *7/1e3) +- (0, 2223.26*7/1e3)
};  \addlegendentry{$\bar q$};

\addplot+[error bars/.cd,y dir=both,y explicit] coordinates{
(5 , 10060.5*7/1e3) +- (0, 1134.85*7/1e3)
(8 , 22054.9*7/1e3) +- (0, 2869.9 *7/1e3)
(10, 30433  *7/1e3) +- (0, 4005.63*7/1e3)
(12, 38887.5*7/1e3) +- (0, 4862.6 *7/1e3)
(14, 45100.4*7/1e3) +- (0, 3524.27*7/1e3)
(16, 47182.6*7/1e3) +- (0, 3729.64*7/1e3)
(18, 55307.2*7/1e3) +- (0, 5790.08*7/1e3)
(20, 56576.6*7/1e3) +- (0, 4643.67*7/1e3)
(25, 69328.4*7/1e3) +- (0, 4968.81*7/1e3)
};  \addlegendentry{$\bar p$};
\end{axis}
\end{tikzpicture}
\hspace{-9pt}
\caption{Hellinger errors in the densities (left) and total number of
function evaluations in Alg.~\ref{alg:dirt} (right) for estimating the
\emph{a posteriori} risk  in Example 2 with $\tau_* = 0.15$ and
varying the parameter dimension $d$.
Points denote averages, and error bars denote one standard deviation over 10 runs.}
\label{fig:post_pde}
\end{figure}
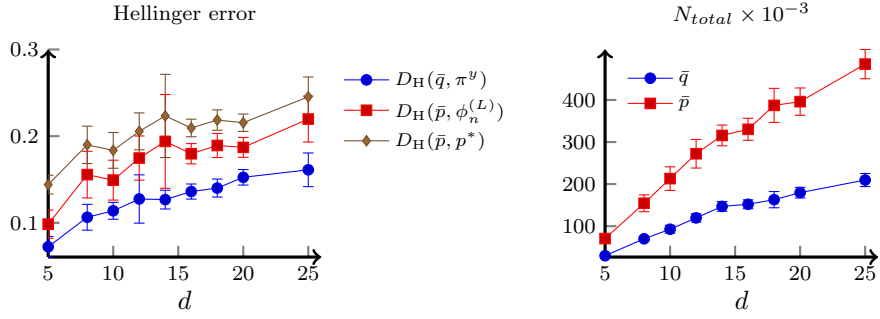

Finally, we vary the dimension of the random field from $d=5$ to
$25$ and take the threshold $\tau_*=0.15$ to test the dimension
scalability of deep importance sampling for estimating the {\it a
posteriori} risk.
The synthetic observations are generated using the diffusivity field
with high diffusivity channel, depicted in the bottom of
Fig.~\ref{fig:diff-channel}. The TT ranks are adaptively
chosen using $5$ iterations of the cross algorithm, starting from rank
$1$ and increasing the ranks by at most $2$ in each iteration to obtain a
relative Frobenius-norm error below $3\cdot 10^{-2}$.
We use piecewise linear basis functions on $17$ grid points to
discretize the density in each coordinate direction, truncating the
unbounded domain to $[-5,5]$. We choose a smoothing width of
$\gamma^* = 100/\tau_*$.
The results are shown in Fig.~\ref{fig:post_pde}.
We observe that the computational
complexity, measured in terms of density evaluations, depends
no worse than linearly on the dimension, while the Hellinger error
increases logarithmically with respect to the dimension.

\section{Future work}\label{sec:conc}

We demonstrated that on problems constrained by differential equations, our proposed deep importance sampling is able to compute hitherto unattainable estimates of rare event probabilities for complex, high-dimensional posterior densities with $d > 20$. 
For problems with very high-dimensional parameters, e.g., $d>10^3$, even though the computational complexity of TT may be independent of the apparent problem dimension if the underlying probability density lies in a Sobolev space with appropriately decaying dimension weights (see \cite{griebel2021analysis} and references therein), it can still be computationally demanding to build TT approximations if the decay in the weights is too slow. To alleviate this challenge, we can apply gradient-based dimension reduction methods \cite{DimRedu:CDW_2014,cui2014likelihood,cui2021unified,uribe2020cross,zahm2018certified} to identify subspaces that capture the most relevant variations of the optimal importance distribution with respect to the underlying weighted norm. The TT approximation in each layer of deep importance sampling can then be further improved using the variable reordering/reparametrization technique in \cite{cui2023scalable} after the gradient-based dimension reduction.

Although deep importance sampling demonstrates good statistical efficiency in terms of the effective sample size per function evaluation in our numerical experiments, the failure function can be computationally costly to evaluate due to the use of numerical solvers for the differential equations. This may prevent a reliable estimation of the failure probability with a limited computational budget. 
To address this bottleneck, one can use surrogate modeling techniques---for example, those based on polynomial chaos \cite{babuvska2007stochastic,cohen2010convergence,SuMo:XiuKar_2002,SuMo:MarNa_2009,schwab2012sparse,yan2019adaptive}, reduced order models \cite{ROM:BWG_2008_2,IP:ChenSch_2015,cohen2022nonlinear,ROM:CMW_2014,ROM:CMW_2016,ROM:GFWG_2008,ROM:LWG_2010,wan2020coupling}, and neural networks \cite{li2020fourier,lu2021learning,tripathy2018deep,zhou2020adaptive,zhu2019physics}---to replace the forward model, so that the training of the Rosenblatt transport can be accelerated. Furthermore, our method can also be combined with either the multilevel Monte Carlo estimator \cite{elfverson2016multilevel,scheichl-qmc-bayes-2017,wagner2020multilevel} or used in a multi-fidelity framework \cite{peherstorfer2016multifidelity,peherstorfer2018multifidelity} to achieve further variance reduction.

\appendix

\section{Proof of Lemma \ref{lemma:dhell_sirt}}\label{proof:dhell_sirt}
Recall that the unnormalized optimal importance density $\rho^*$ is approximated by $\rho = \tilde g^2 + \tau \lambda$, where $\lambda$ is a normalized probability density, $\tau > 0$, and $\tilde g$ satisfies $\|\surd \rho^* - \tilde g \|_2 \leq \epsilon$. 
Since $\rho^*$ and $\lambda$ are non-negative functions and $\tau>0$, we have the identity
\begin{align*}
( \surd \rho^* - \surd \rho)^2 & = \{ \surd \rho^* - (\tilde g^2 + \tau \lambda)^{1/2} \}^2 \\
& = \rho^* + \tilde g^2 + \tau \lambda -  2 \surd \rho^* (\tilde g^2 + \tau \lambda)^{1/2} \\
& \leq \rho^* + \tilde g^2 + \tau \lambda -  2 \surd \rho^* \tilde g \\
& = ( \surd \rho^* - \tilde g )^2 + \tau \lambda,
\end{align*}
which leads to $\|\surd \rho^* - \surd \rho\|^2_2 \leq \|\surd \rho^* - \tilde g  \|^2_2 + \tau \leq \epsilon^2 + \tau$. Choosing $\tau \leq \epsilon^2$, we have
\begin{equation}\label{eq:sirt_l2}
\|\surd \rho^* - \surd \rho\|_2 \leq  \surd 2 \epsilon.
\end{equation}

Since the square roots of the normalising constants can be expressed as $\surd \zeta^* = \|\surd \rho^*\|_2$ and $\surd \zeta = \|\surd \rho\|_2$, we have
\begin{align*}
|\surd \zeta^* - \surd \zeta | (\surd \zeta^* + \surd \zeta) & = | \zeta^* - \zeta | \\
& = \left| \int_\mathcal{X} \rho^*(x) - \rho(x)  \d x \right| \\
& = | \langle \surd \rho^* - \surd \rho, \surd \rho^* + \surd \rho \rangle | \\
& \leq \|\surd \rho^* - \surd \rho\|_2 \|\surd \rho^* + \surd \rho\|_2 \\
& \leq \|\surd \rho^* - \surd \rho\|_2 (\|\surd \rho^*\|_2 + \|\surd \rho\|_2 ) \\
& = \|\surd \rho^* - \surd \rho\|_2 (\surd \zeta^* + \surd \zeta ).
\end{align*}
This leads to 
\begin{equation}\label{eq:sirt_zeta}
|\surd \zeta^* - \surd \zeta | \leq  \|\surd \rho^* - \surd \rho\|_2.
\end{equation}
Thus, the result of the first property of Lemma \ref{lemma:dhell_sirt} follows. 

Recall that the Hellinger distance is proportional to the $L^2$ distance of the normalized densities, i.e., 
\begin{align*}
D_{\rm H}(p^*, p) & = \left[\frac12 \int \{\surd p^*(x) - \surd p(x)\}^2 \d x \right]^\frac12 = \frac1{\sqrt2} \|\surd p^* - \surd p\|_2.
\end{align*}
The $L^2$ distance of the normalized densities follows the identity
\begin{align*}
\|\surd p^* - \surd p\|_2 & = \left\| \frac{\surd \rho^*}{\surd \zeta^*} - \frac{\surd \rho}{\surd \zeta} \right\|_2 \\
& = \frac1{\surd \zeta^*} \left\| \surd \rho^* - \surd \rho + \surd \rho- \surd \rho\frac{\surd \zeta^*}{\surd \zeta} \right\|_2 \\
& \leq \frac1{\surd \zeta^*} \| \surd \rho^* - \surd \rho \|_2 + \frac1{\surd \zeta^*} \left\| \surd \rho \left(1 - \frac{\surd \zeta^*}{\surd \zeta} \right) \right\|_2 \\
& = \frac1{\surd \zeta^*} \| \surd \rho^* - \surd \rho \|_2 + \frac{\surd \zeta}{\surd \zeta^*}  \left(1 - \frac{\surd \zeta^*}{\surd \zeta}\right) \\
& = \frac1{\surd \zeta^*} \left( \| \surd \rho^* - \surd \rho \|_2 + \surd \zeta-\surd \zeta^* \right) \\
& \leq \frac2{\surd \zeta^*} \| \surd \rho^* - \surd \rho \|_2,
\end{align*}
where the last inequality follows from \eqref{eq:sirt_zeta}. Substituting \eqref{eq:sirt_l2} into the above inequality and the definition of the Hellinger distance, we obtain $D_{\rm H}(p^*, p) \leq 2 \epsilon / \surd \zeta^*$. This gives the second property. \textqed

\section{Sequential marginalisation}\label{proof:marginalization}
Here we provide implementation details of the sequence of one-dimensional integrations for building the Rosenblatt transport in Section \ref{sec:sirt}.
To realize the map $\mathcal{Q}$, our starting point is to construct a sequence of unnormalized marginal densities
\begin{align}
\rho_{\leq k}(x_{\leq k}) & = \int_{\mathcal{X}_{>k}}  \rho(x_{\leq k}, x_{>k}) \, \d x_{>k} =  \int_{\mathcal{X}_{>k}}  \tilde g(x_{\leq k}, x_{>k})^2 \, \d x_{>k} + \tau \lambda_{\leq k}(x_{\leq k}),
\end{align}
where $\lambda_{\leq k}(x_{\leq k}) = \prod_{j = 1}^{k} \lambda_j(x_j)$, for all $1 \leq k < d$. 
Recalling the tensor-train decomposition
\[
\tilde{g}(x) = \mG_{1}(x_1) \cdots \mG_{k}(x_k) \cdots \mG_{d}(x_d),
\]
we can define 
\[
\mG_{\leq k}(x_{\leq k}) = \mG_{1}(x_1) \cdots \mG_{k}(x_k), \quad 
\mG_{>k}(x_{>k}) = \mG_{k+1}(x_{k-1}) \cdots \mG_{d}(x_d),
\]
where $\mG_{\leq k}(x_{\leq k}) \in \R^{1 \times r_k}$ and $\mG_{>k}(x_{>k}) \in \R^{r_k \times 1}$ are row-vector-valued and column-vector-valued functions, respectively.
Then, $\tilde g$ can be written as 
\(
\tilde g(x_{\leq k}, x_{>k}) = \mG_{\leq k}(x_{\leq k}) \mG_{>k}(x_{>k}).
\)  
The integration of $\tilde g^2$ over $x_{>k}$ for any index $k$, and hence the unnormalized marginal densities, can be obtained dimension-by-dimension as follows.

\begin{enumerate}[leftmargin=*]
\item For $k = d-1$, we integrate $\tilde g^2$ over the last coordinate $x_d$ to obtain 
\begin{align}
\rho_{< d}(x_{<d}) % 
& =  \int_{\mathcal{X}_d} \bigg\{\sum_{\alpha_{d{-}1}=1}^{r_{d-1}} \mG_{<d}^{(\alpha_{d-1})}(x_{<d}) \, \mG_{d}^{(\alpha_{d-1})}(x_d) \bigg\}^2 \, \d x_d + \tau \lambda_{<d}(x_{<d}) \nonumber \\
& =  \sum_{\alpha_{d{-}1}=1}^{r_{d-1}} \sum_{\beta_{d{-}1}=1}^{r_{d-1}} \mG_{<d}^{(\alpha_{d-1})}(x_{<d}) \, \mG_{<d}^{(\beta_{d-1})}(x_{<d}) \, \mM^{(\alpha_{d-1}, \beta_{d-1})}_{d} + \tau \lambda_{<d}(x_{<d}), \nonumber
\end{align}
where $\mM_{d} \in \R^{ r_{d{-}1} \times r_{d{-}1}}$ is a symmetric positive definite mass matrix such that
\begin{equation}\label{eq:md}
\mM^{(\alpha_{d-1}, \beta_{d-1})}_{d} = \int_{\mathcal{X}_d} \mG_{d}^{(\alpha_{d-1})}(x_d)\,\mG_{d}^{(\beta_{d-1})}(x_d) \, \d x_d.
\end{equation}
Computing the Cholesky factorization $\chol_d^{} \chol_d^\top= \mM_d^{}$, we have the simplification
\begin{equation}\label{eq:marginal_d}
\rho_{< d}(x_{<d}) = \sum_{\alpha_{d{-}1}=1}^{r_{d-1}} \Big\{ \mG_{<d}(x_{<d}) \, \chol_d^{(:,\alpha_{d-1})} \Big\}^2+ \tau \lambda_{<d}(x_{<d}).
\end{equation}
\item For any index $1 < k < d$, suppose we have the symmetric positive definite mass matrix $\bar\mM_{>k} \in \R^{ r_k \times r_k}$ such that
\[
\bar\mM^{(\alpha_k, \beta_k)}_{>k} = \int_{\mathcal{X}_{>k}} \mG_{>k}^{(\alpha_k)}(x_d)\,\mG_{>k}^{(\beta_k)}(x_{>k}) \, \d x_{>k}
\]
and its Cholesky factorization $\bar\chol_{>k}^{} \bar\chol_{>k}^\top= \bar\mM_{>k}^{}$. Then, similar to the above case, we have the unnormalized marginal density
\[
\rho_{\leq k}(x_{\leq k}) = \sum_{\alpha_k=1}^{r_k} \Big\{ \mG_{\leq k}(x_{\leq k}) \, \bar\chol_{>k}^{(:,\alpha_k)} \Big\}^2+ \tau \lambda_{\leq k}(x_{\leq k}).
\]
This way, the next unnormalized marginal density $\rho_{< k}(x_{<k})$ can be constructed by a one-dimensional integration over $x_k$, which takes the form 
\begin{align}
\rho_{< k}(x_{<k}) & =   \sum_{\alpha_k=1}^{r_k} \int_{\mathcal{X}_k}  \!\!\Big\{\!\!  \sum_{\alpha_{k-1}=1}^{r_{k-1}} \mG_{<k}^{(\alpha_{k-1})}(x_{<k}) \, \mG_k^{(\alpha_{k-1},:)}(x_k) \, \bar\chol_{>k}^{(:,\alpha_k)} \Big\}^2 \d x_k + \tau \lambda_{<k}(x_{<k}) \nonumber \\
& = \sum_{\alpha_{k-1}=1}^{r_{k-1}} \sum_{\beta_{k-1}=1}^{r_{k-1}} \mG_{<k}^{(\alpha_{k-1})}(x_{<k})\, \mG_{<k}^{(\beta_{k-1})}(x_{<k}) \, \bar\mM^{(\alpha_{k-1}, \beta_{k-1})}_{\geq k} + \tau \lambda_{<k}(x_{<k}) , \nonumber
\end{align}
where $\bar\mM_{\geq k} \in \R^{ r_{k-1} \times r_{k-1}}$ is the next mass matrix such that
\begin{equation}\label{eq:Mk}
\bar\mM^{(\alpha_{k-1}, \beta_{k-1})}_{\geq k} = \sum_{\alpha_k=1}^{r_k} \int_{\mathcal{X}_k} \big\{\mG_k^{(\alpha_{k-1},:)}(x_k) \, \bar\chol_{>k}^{(:,\alpha_k)}\big\} \big\{\mG_k^{(\beta_{k-1},:)}(x_k) \, \bar\chol_{>k}^{(:,\alpha_k)}\big\}  \, \d x_k.
\end{equation}
Again, by computing the Cholesky factorization $\bar\chol_{\geq k}^{}\bar\chol_{\geq k}^\top= \bar\mM_{\geq k}^{}$, we have the simplified marginal density
\begin{equation}\label{eq:marginal_k}
\rho_{< k}(x_{< k}) = \sum_{\alpha_{k-1}=1}^{r_{k-1}} \Big\{ \mG_{< k}(x_{< k}) \, \bar\chol_{\geq k}^{(:,\alpha_{k-1})} \Big\}^2+ \tau \lambda_{< k}(x_{< k}).
\end{equation}
Following the above procedure, initializing $\bar\mM_{>k}$ with $\bar\mM_{>k} = \mM_d$ for $k = d-1$, we can recursively construct all unnormalized marginal densities. In each iteration, we only need to solve a one-dimensional integration problem in \eqref{eq:Mk}. Given $n_k$ number of discretization basis functions in $x_k$, the total computational complexity of solving the integration in \eqref{eq:Mk} and computing the Cholesky factorization $\bar\chol_{\geq k}$ is $\mathcal{O}(n_k r_k r_{k-1}^2 + r_{k-1}^3)$.

\item For $k = 1$, we have the unnormalized marginal density
\[
\rho_{\leq 1}(x_1) = \sum_{\alpha_1=1}^{r_1} \Big\{ \mG_{1}(x_{1}) \, \bar\chol_{>1}^{(:,\alpha_1)} \Big\}^2+ \tau \lambda_{\leq 1}(x_1).
\]
Carrying out one extra integration defined in \eqref{eq:Mk}, we obtain $\bar\mM_{\geq 1}\in\R$ as $r_0 = 1$. This gives the normalising constant $\zeta = \bar\mM_{\geq 1} + \tau$.
\end{enumerate}

\section{Pushforward density of the composite map}\label{proof:density_dirt}
Here we provide a detailed derivation of the normalized density $\bar{p} = \{\mathcal{T}^\lowsup{(L)}\}_\sharp\,\lambda$ in \eqref{eq:dirt_density}, which is the pushforward density of the reference $\lambda$ under the composition of maps 
\(
\mathcal{T}^\lowsup{(L)} = \mathcal{Q}^\lowsup{(1)} \circ \mathcal{Q}^\lowsup{(2)} \circ \cdots \circ \mathcal{Q}^\lowsup{(L)}.
\)
As a starting point, we derive the Jacobian of the incremental map $u' = \mathcal{Q}^\lowsup{(\ell)}(u)$, which has the form
\[
\mathcal{Q}^\lowsup{(\ell)} = \mathcal{F}^{-1} \circ \mathcal{R},
\]
with $\mathcal{F}_\sharp\,p^\lowsup{(\ell)} = \mu$ and $\mathcal{R}_\sharp\,\lambda = \mu$, where $\mu$ is the uniform density on $[0,1]^d$ and 
\begin{equation}\label{eq:density_tmp1}
p^\lowsup{(\ell)}(u') = \frac{1}{\zeta^\lowsup{(\ell)}} \Big\{ \tilde{g}^\lowsup{(\ell)}(u')^2 + \tau^\lowsup{(\ell)} \lambda(u') \Big\}
\end{equation}
is the $\ell$-th approximate density. Thus, we have the identity
\begin{equation}\label{eq:density_tmp2}
p^\lowsup{(\ell)}(u') = \{\mathcal{Q}^\lowsup{(\ell)}\}_\sharp\,\lambda(u') = \lambda\left[ \{\mathcal{Q}^\lowsup{(\ell)}\}^{-1}(u') \right] \left| \nabla \{\mathcal{Q}^\lowsup{(\ell)}\}^{-1}(u') \right|,
\end{equation}
which gives the Jacobian
\[
\left| \nabla \{\mathcal{Q}^\lowsup{(\ell)}\}^{-1}(u') \right| = \frac{p^\lowsup{(\ell)}(u')}{\lambda\left[ \{\mathcal{Q}^\lowsup{(\ell)}\}^{-1}(u') \right]}.
\]

Given a composite map $\mathcal{T}^\lowsup{(\ell)} = \mathcal{T}^\lowsup{(\ell{-}1)}\circ \mathcal{Q}^\lowsup{(\ell)}$, to avoid confusion, we define the associated change of variables as
\[
x = \mathcal{T}^\lowsup{(\ell)}(u) \quad \iff \quad u' = \mathcal{Q}^\lowsup{(\ell)}(u), \quad x = \mathcal{T}^\lowsup{(\ell{-}1)}(u'),
\]
and the reverse transform as
\[
u = \{\mathcal{T}^\lowsup{(\ell)}\}^{-1}(x) \quad \iff \quad  u' = \{\mathcal{T}^\lowsup{(\ell{-}1)}\}^{-1}(x), \quad  u = \big\{ \mathcal{Q}^\lowsup{(\ell)}\big\}^{-1} (u') .
\]
This way, the Jacobian of the inverse map satisfies
\[
\left| \nabla \{\mathcal{T}^\lowsup{(\ell)}\}^{-1}(x) \right| = \left| \nabla \{\mathcal{Q}^\lowsup{(\ell)}\}^{-1}(u') \right| \, \left| \nabla \{\mathcal{T}^\lowsup{(\ell{-}1)}\}^{-1}(x) \right| ,
\]
by the chain rule. Substituting \eqref{eq:density_tmp2} and $u' = \{\mathcal{T}^\lowsup{(\ell{-}1)}\}^{-1}(x)$ into the above identity, the Jacobian of the composite map satisfies the recurrence relationship
\begin{align}
\left| \nabla \{\mathcal{T}^\lowsup{(\ell)}\}^{-1}(x) \right| & = \left| \nabla \{\mathcal{T}^\lowsup{(\ell{-}1)}\}^{-1}(x) \right| \frac{p^\lowsup{(\ell)}\big[\{\mathcal{T}^\lowsup{(\ell{-}1)}\}^{-1}(x)\big]}{\lambda\big( \{\mathcal{Q}^\lowsup{(\ell)}\}^{-1} \big[ \{\mathcal{T}^\lowsup{(\ell{-}1)}\}^{-1}(x) \big] \big) } \nonumber \\
& = \left| \nabla \{\mathcal{T}^\lowsup{(\ell{-}1)}\}^{-1}(x) \right|  \frac{p^\lowsup{(\ell)}\big[\{\mathcal{T}^\lowsup{(\ell{-}1)}\}^{-1}(x)\big]}{\lambda\big[\{\mathcal{T}^\lowsup{(\ell)}\}^{-1}(x)\big]} \label{eq:density_tmp3}
\end{align}
Thus, by induction, the Jacobian of the composite of $L$ layers of maps, $\mathcal{T}^\lowsup{(L)}$, satisfies
\begin{align}
\left| \nabla \{\mathcal{T}^\lowsup{(L)}\}^{-1}(x) \right| & = \left| \nabla \{\mathcal{T}^\lowsup{(0)}\}^{-1}(x) \right| \Bigg( \frac{p^\lowsup{(1)}\big[\{\mathcal{T}^\lowsup{(0)}\}^{-1}(x)\big]}{\lambda\big[\{\mathcal{T}^\lowsup{(1)}\}^{-1}(x)\big]} \cdots \frac{p^\lowsup{(L)}\big[\big\{\mathcal{T}^\lowsup{(L{-}1)}\big\}^{-1}(x)\big]}{\lambda\big[\{\mathcal{T}^\lowsup{(L)}\}^{-1}(x)\big]} \Bigg)\nonumber \\
& = \frac{p^\lowsup{(1)}\big(x\big)}{\lambda\big[\{\mathcal{T}^\lowsup{(L)}\}^{-1}(x)\big]} \prod_{\ell=2}^L \frac{p^\lowsup{(\ell)}\big[\{\mathcal{T}^\lowsup{(\ell{-}1)}\}^{-1}(x)\big]}{\lambda\big[\{\mathcal{T}^\lowsup{(\ell{-}1)}\}^{-1}(x)\big]}  .\label{eq:density_tmp4}
\end{align}
Substituting \eqref{eq:density_tmp4} into the identity 
\[
\{\mathcal{T}^\lowsup{(L)}\}_\sharp\,\lambda(x) = \lambda\left[ \{\mathcal{T}^\lowsup{(L)}\}^{-1}(x) \right] \left| \nabla \{\mathcal{T}^\lowsup{(L)}\}^{-1}(x) \right|
\]
and applying \eqref{eq:density_tmp1}, the pushforward density of $\lambda$ under $\mathcal{T}^\lowsup{(L)}$ has the density
\begin{align}
\{\mathcal{T}^\lowsup{(L)}\}_\sharp\,\lambda(x) & = p^\lowsup{(1)}\big(x\big) \prod_{\ell=2}^L \frac{p^\lowsup{(\ell)}\big[\{\mathcal{T}^\lowsup{(\ell{-}1)}\}^{-1}(x)\big]}{\lambda\big[\{\mathcal{T}^\lowsup{(\ell{-}1)}\}^{-1}(x)\big]}  \\
& = \left\{\prod_{\ell=1}^L \zeta^\lowsup{(\ell)}\right\}^{-1} \hspace{-12pt}\left\{ \tilde{g}^\lowsup{(1)}(x)^2 + \tau^\lowsup{(1)} \lambda(x) \right\} \prod_{\ell=2}^L \bigg(\frac{\tilde{g}^\lowsup{(\ell)}\big[\{\mathcal{T}^\lowsup{(\ell{-}1)}\}^{-1}(x)\big]^2}{\lambda\big[\{\mathcal{T}^\lowsup{(\ell{-}1)}\}^{-1}(x)\big] } + \tau^\lowsup{(\ell)} \bigg). \nonumber
\end{align}
This concludes the derivation. \textqed

\section{Areas of annulus and disk}\label{sec:disk}
We consider a 2-dimensional toy example for estimating
{\it a priori} failure probabilities, where the prior distribution
that is uniform on the unit square, i.e., $\pi_0(x) = 1$ with $x\in
[0,1]^2$ and the failure function
\begin{equation}\label{eq:annulus}
f(x) = \indi_{\{R_i \le \|x-x_0\|_2\le R_{o}\}}(x),
\end{equation}
for given radii $0\le R_i<R_o$ and center $x_0=[0.4, 0.4]$. Thus, the event probability is the area of the annulus, 
\(
\zeta^*:=\mathrm{pr}_{\pi_0}(X \in \mathcal{A}) = \mathrm{Pi}\, (R_o^2 - R_i^2),
\)
where $\mathrm{Pi}$ is Archimedes' constant. 

The smoothed indicator function for Alg.~\ref{alg:dirt} is defined as a product of two sigmoids,
$$
f_{\gamma}(x) = \left[1+\exp\{\gamma(\|x-x_0\|_2^2-R_o^2)\}\right]^{-1} \left[1+\exp\{\gamma(R_i^2-\|x-x_0\|_2^2)\}\right]^{-1}.
$$
To approximate the smoothed optimal importance density with Alg.~\ref{alg:dirt},
we tune various control variables in the deep importance sampling
procedure such
that the Hellinger distance between the approximate density and the
optimal importance density $p^*(x)$ is about $0.3$ for all choices of
$R_i$ and $R_o$. This involves varying the final smoothing variable
$\gamma_L = \gamma^*$,  the univariate grid size $n$, the tensor rank $r$, and the initial
smoothing  variable $\gamma_1$. The intermediate densities
are defined throughout by $\gamma_{\ell+1}=\surd 10\,\gamma_\ell$. 
Once the approximation of the optimal importance density is computed, we
use $N=2^{16}$ samples to compute the deep importance sampling
estimator  $\hat{\zeta}_{\bar{p},N}$ in
\eqref{eq:dirt_est2}. %,

\begin{figure}[t]
\noindent
\includegraphics[width=0.45\linewidth]{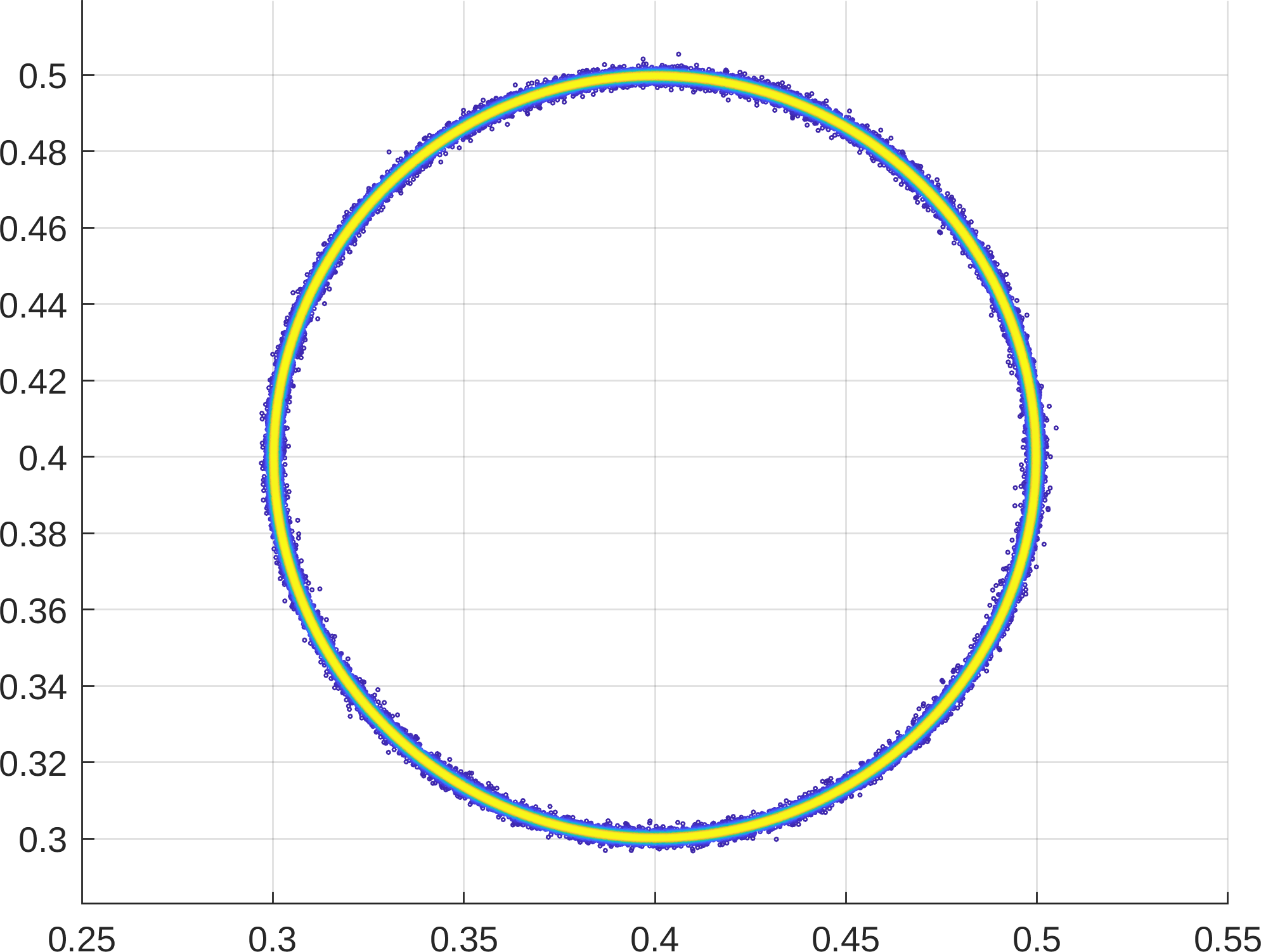}
\hfill
\includegraphics[width=0.45\linewidth]{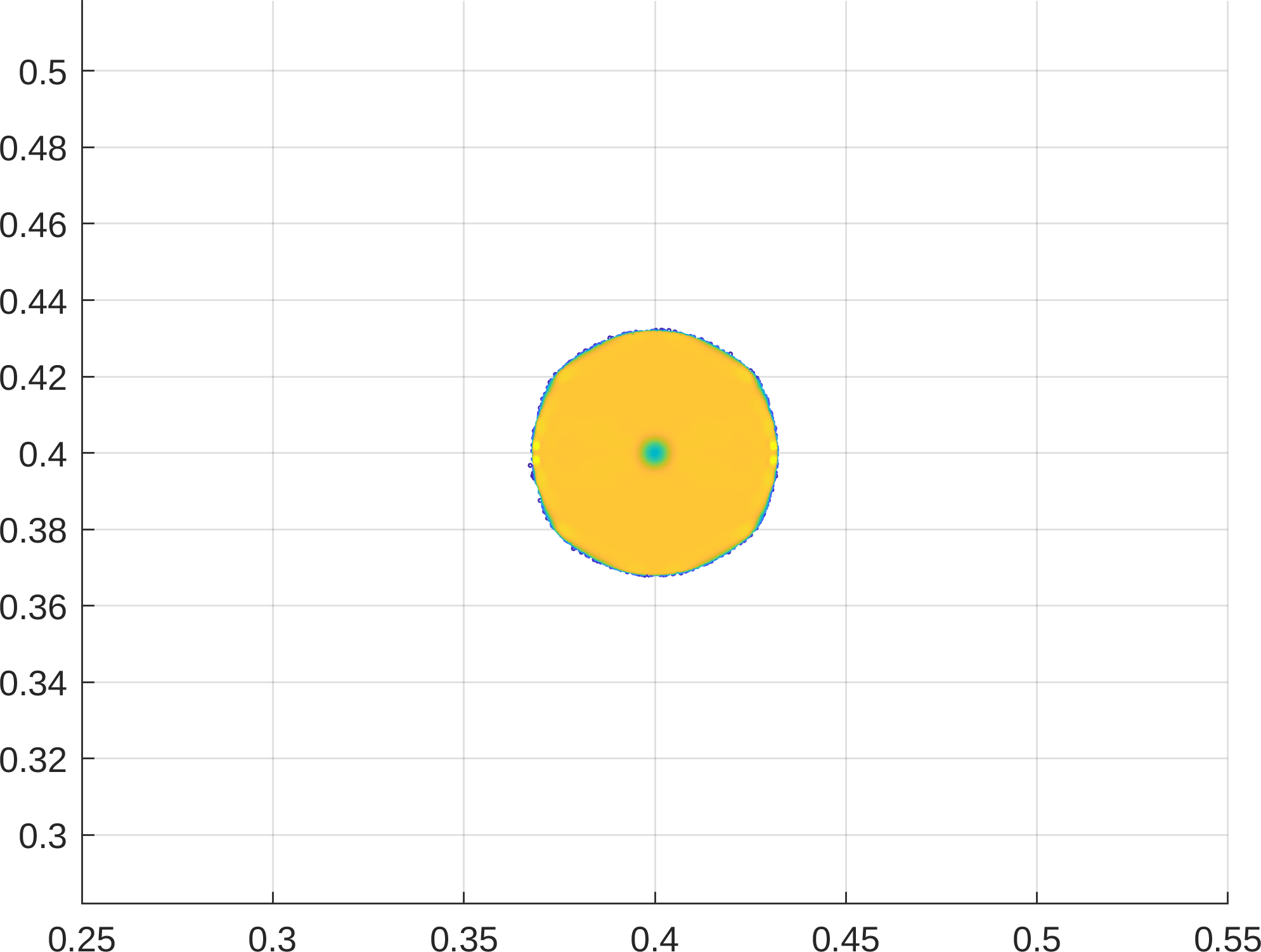}\vspace{6pt}
\caption{Samples drawn from the approximate importance densities, colored by their density values, for $R_o^2=10^{-2}$, $R_i^2=R_o^2-10^{-4}$ (left) and for
$R_o^2=10^{-3}$, $R_i^2=0$ (right).} \label{fig:ring}
\end{figure}
In the first experiment, we fix the outer radius $R_o=0.1$, and vary the inner radius
$R_i$, as shown in Fig.~\ref{fig:ring} (left), such that it approaches
$R_o$. The results are shown in Table~\ref{tab:an}. 
\begin{table}[t]
\centering
{\footnotesize
\caption{Annulus test with $R_o=0.1$ fixed. $N_{TT}$ is the total
number of function evaluations used in Alg.~\ref{alg:dirt} to
approximate the smoothed optimal importance density. The last column
gives the relative bias of the estimator in each case.}
\label{tab:an}
\noindent
\begin{tabular}{c|ccccc|cc}
$R_o^2-R_i^2$ & $\gamma^*$ & $n$ & $r$ & $\gamma_1$ & $N_{TT}$ & $D_H(p^*,\bar p)$ & $|\hat{\zeta}_{\bar{p},N} - \zeta^*|/\zeta^*$ \\ \hline
$10^{-3}$ & $10^4$ & 33 & 3  & $10^{-3}$  & 1386    & 0.308$\pm$0.0014  &  0.00244$\pm$0.00114 \\
$10^{-4}$ & $10^5$ & 65 & 3  & $10^{-4}$ & 3510    & 0.292$\pm$0.0033  &  0.00162$\pm$0.00158 \\
$10^{-5}$ & $10^6$ & 257 & 5  & $10^{-4}$ & 23130  & 0.292$\pm$0.0159  &  0.00293$\pm$0.00570 \\
$10^{-6}$ & $10^7$ & 513 & 10 & $10^{-5}$ & 112860 & 0.304$\pm$0.0111  &  0.00232$\pm$0.00180  \\
$10^{-7}$ & $10^8$ & 1025 & 20 & $10^{-6}$ & 533000& 0.379$\pm$0.0320  &  0.00616$\pm$0.00445 \\
\end{tabular}}
\end{table}
This setup requires finer discretizations, that is, larger values of
$n$, as the width of the annulus decreases. As a result, the number of
function evaluations to
approximate the optimal importance density, $N_{TT}$, grows rapidly.

In contrast, if the inner radius is fixed to $R_i=0$ and the outer
radius $R_o$  is varied, the optimal importance density function
$p^*(x) \propto f(x)\pi_0(x)$ is unimodal, representing just the
indicator function of the disk with radius $R_o$. As we can see in
Table~\ref{tab:disk}, in that case the approximation complexity, in
terms of function evaluations,
depends only logarithmically on the value of $\zeta^*$.
\begin{table}[t]
\centering
{\footnotesize
\caption{Disk test with $R_i=0$ fixed. $N_{TT}$ is the total number of
function evaluations used in Alg.~\ref{alg:dirt} for approximating
the smoothed optimal importance densities. The last column
gives the relative bias of the estimator in each case.}
\label{tab:disk}
\noindent
\begin{tabular}{c|ccccc|cc}
$R_o^2$ & $\gamma^*$ & $n$ & $r$ & $\gamma_1$ & $N_{TT}$ & $D_H(p^*,\bar p)$ & $|\hat{\zeta}_{\bar{p},N} - \zeta^*|/\zeta^*$ \\ \hline
$10^{-2}$ & $10^3$ & 17 & 2 & $10^{-2}$ & 340 & 0.224$\pm$0.0015 &  0.00136$\pm$0.00094 \\
$10^{-3}$ & $10^4$ & 17 & 2 & $10^{-2}$ & 340 & 0.221$\pm$0.0036 &  0.00111$\pm$0.00078 \\
$10^{-4}$ & $10^5$ & 17 & 2 & $10^{-3}$ & 476 & 0.218$\pm$0.0017 &  0.00105$\pm$0.00090 \\
$10^{-5}$ & $10^6$ & 17 & 2 & $10^{-4}$ & 612 & 0.218$\pm$0.0015 &  0.00144$\pm$0.00095 \\
$10^{-6}$ & $10^7$ & 17 & 2 & $10^{-5}$ & 748 & 0.218$\pm$0.0015 &  0.00193$\pm$0.00100 \\
$10^{-7}$ & $10^8$ & 17 & 2 & $10^{-5}$ & 748 & 0.222$\pm$0.0041 &  0.00105$\pm$0.00072 \\
\end{tabular}}
\end{table}

%{

%}


\begin{thebibliography}{10}

\bibitem{babuvska2007stochastic}
{\sc I.~Babu{\v{s}}ka, F.~Nobile, and R.~Tempone}, {\em A stochastic
  collocation method for elliptic partial differential equations with random
  input data}, SIAM Journal on Numerical Analysis, 45 (2007), pp.~1005--1034.

\bibitem{baptista2020adaptive}
{\sc R.~Baptista, Y.~Marzouk, and O.~Zahm}, {\em On the representation and
  learning of monotone triangular transport maps}, arXiv preprint
  arXiv:2009.10303,  (2020).

\bibitem{bigoni2016spectral}
{\sc D.~Bigoni, A.~P. Engsig-Karup, and Y.~M. Marzouk}, {\em Spectral
  tensor-train decomposition}, SIAM J. Sci. Comput., 38 (2016),
  pp.~A2405--A2439.

\bibitem{botev2008efficient}
{\sc Z.~I. Botev and D.~P. Kroese}, {\em An efficient algorithm for rare-event
  probability estimation, combinatorial optimization, and counting}, Methodol.
  Comput. Appl. Probab., 10 (2008), pp.~471--505.

\bibitem{bigoni2019greedy}
{\sc M.~Brennan, D.~Bigoni, O.~Zahm, A.~Spantini, and Y.~Marzouk}, {\em Greedy
  inference with structure-exploiting lazy maps}, Adv. Neural Inf. Process
  Syst., 33 (2020), pp.~8330--8342.

\bibitem{ROM:BWG_2008_2}
{\sc T.~Bui-Thanh, K.~E. Willcox, and O.~Ghattas}, {\em Model reduction for
  large-scale systems with high-dimensional parametric input space}, SIAM J.
  Sci. Comput., 30 (2008), pp.~3270--3288.

\bibitem{cappe2008adaptive}
{\sc O.~Capp{\'e}, R.~Douc, A.~Guillin, J.-M. Marin, and C.~P. Robert}, {\em
  Adaptive importance sampling in general mixture classes}, Stat. Comput., 18
  (2008), pp.~447--459.

\bibitem{IP:ChenSch_2015}
{\sc P.~Chen and C.~Schwab}, {\em Sparse-grid, reduced-basis {B}ayesian
  inversion}, Comput. Methods Appl. Mech. Eng.,  (2015), p.~in press.

\bibitem{cliffe2000}
{\sc K.~A. Cliffe, I.~G. Graham, R.~Scheichl, and L.~Stals}, {\em Parallel
  computation of flow in heterogeneous media modelled by mixed finite
  elements}, J. Comput. Phys., 164 (2000), pp.~258--282.

\bibitem{cohen2022nonlinear}
{\sc A.~Cohen, W.~Dahmen, O.~Mula, and J.~Nichols}, {\em Nonlinear reduced
  models for state and parameter estimation}, SIAM/ASA Journal on Uncertainty
  Quantification, 10 (2022), pp.~227--267.

\bibitem{cohen2010convergence}
{\sc A.~Cohen, R.~DeVore, and C.~Schwab}, {\em Convergence rates of best n-term
  galerkin approximations for a class of elliptic spdes}, Foundations of
  Computational Mathematics, 10 (2010), pp.~615--646.

\bibitem{DimRedu:CDW_2014}
{\sc P.~G. Constantine, E.~Dow, and Q.~Wang}, {\em Active subspace methods in
  theory and practice: Applications to kriging surfaces}, SIAM J. Sci. Comput.,
  36 (2014), pp.~A1500--A1524.

\bibitem{cui2021deep}
{\sc T.~Cui and S.~Dolgov}, {\em Deep composition of tensor-trains using
  squared inverse rosenblatt transports}, Found. Comput. Math., 22 (2022),
  pp.~1863--1922.

\bibitem{cui2023scalable}
{\sc T.~Cui, S.~Dolgov, and O.~Zahm}, {\em Scalable conditional deep inverse
  rosenblatt transports using tensor trains and gradient-based dimension
  reduction}, Journal of Computational Physics, 485 (2023), p.~112103.

\bibitem{cui2023self}
{\sc T.~Cui, S.~Dolgov, and O.~Zahm}, {\em Self-reinforced polynomial
  approximation methods for concentrated probability densities}, arXiv preprint
  arXiv:2303.02554,  (2023).

\bibitem{cui2014likelihood}
{\sc T.~Cui, J.~Martin, Y.~M. Marzouk, A.~Solonen, and A.~Spantini}, {\em
  Likelihood-informed dimension reduction for nonlinear inverse problems},
  Inverse Problems, 30 (2014), p.~114015.

\bibitem{ROM:CMW_2014}
{\sc T.~Cui, Y.~M. Marzouk, and K.~E. Willcox}, {\em Data-driven model
  reduction for the bayesian solution of inverse problems}, International
  Journal for Numerical Methods in Engineering, 102 (2015), pp.~966--990,
  \url{https://doi.org/10.1002/nme.4748}.

\bibitem{ROM:CMW_2016}
{\sc T.~Cui, Y.~M. Marzouk, and K.~E. Willcox}, {\em Scalable posterior
  approximations for large-scale bayesian inverse problems via
  likelihood-informed parameter and state reduction}, Journal of Computational
  Physic, 315 (2016), pp.~363--387.

\bibitem{cui2021unified}
{\sc T.~Cui and X.~T. Tong}, {\em A unified performance analysis of
  likelihood-informed subspace methods}, Bernoulli, 28 (2022), pp.~2788--2815.

\bibitem{del2006sequential}
{\sc P.~Del~Moral, A.~Doucet, and A.~Jasra}, {\em Sequential monte carlo
  samplers}, J. R. Stat. Soc. Series B, 68 (2006), pp.~411--436.

\bibitem{dodwell2021multilevel}
{\sc T.~J. Dodwell, S.~Kynaston, R.~Butler, R.~T. Haftka, N.~H. Kim, and
  R.~Scheichl}, {\em Multilevel monte carlo simulations of composite structures
  with uncertain manufacturing defects}, Probabilistic Eng. Mech., 63 (2021),
  p.~103116.

\bibitem{dafs-tt-bayes-2019}
{\sc S.~Dolgov, K.~Anaya-Izquierdo, C.~Fox, and R.~Scheichl}, {\em
  Approximation and sampling of multivariate probability distributions in the
  tensor train decomposition}, Stat. Comput., 30 (2020), pp.~603--625.

\bibitem{dolgov2014alternating}
{\sc S.~V. Dolgov and D.~V. Savostyanov}, {\em Alternating minimal energy
  methods for linear systems in higher dimensions}, SIAM J. Sci. Comput., 36
  (2014), pp.~A2248--A2271.

\bibitem{douc2007convergence}
{\sc R.~Douc, A.~Guillin, J.-M. Marin, and C.~P. Robert}, {\em Convergence of
  adaptive mixtures of importance sampling schemes}, Ann. Stat., 35 (2007),
  pp.~420--448.

\bibitem{DGKP-SEIR-2021}
{\sc R.~Dutta, S.~N. Gomes, D.~Kalise, and L.~Pacchiardi}, {\em Using mobility
  data in the design of optimal lockdown strategies for the {COVID-19}
  pandemic}, {PLoS} Comput. Biol., 17 (2021), pp.~1--25.

\bibitem{eigel2020low}
{\sc M.~Eigel, R.~Gruhlke, and M.~Marschall}, {\em Low-rank tensor
  reconstruction of concentrated densities with application to bayesian
  inversion}, Stat. Comput., 32 (2022), pp.~1--27.

\bibitem{eigel2018sampling}
{\sc M.~Eigel, M.~Marschall, and R.~Schneider}, {\em Sampling-free bayesian
  inversion with adaptive hierarchical tensor representations}, Inverse
  Problems, 34 (2018), p.~035010.

\bibitem{elfverson2016multilevel}
{\sc D.~Elfverson, F.~Hellman, and A.~M{\aa}lqvist}, {\em A multilevel monte
  carlo method for computing failure probabilities}, SIAM/ASA J. Uncertain.
  Quantif., 4 (2016), pp.~312--330.

\bibitem{evans1995methods}
{\sc M.~Evans and T.~Swartz}, {\em Methods for approximating integrals in
  statistics with special emphasis on bayesian integration problems},
  Statistical science,  (1995), pp.~254--272.

\bibitem{ROM:GFWG_2008}
{\sc D.~Galbally, K.~Fidkowski, K.~E. Willcox, and O.~Ghattas}, {\em Nonlinear
  model reduction for uncertainty quantification in large scale inverse
  problems}, International journal for numerical methods in engineering, 81
  (2008), pp.~1581--1608.

\bibitem{gelman1998simulating}
{\sc A.~Gelman and X.-L. Meng}, {\em Simulating normalizing constants: From
  importance sampling to bridge sampling to path sampling}, Statistical
  science,  (1998), pp.~163--185.

\bibitem{gorodetsky2019continuous}
{\sc A.~Gorodetsky, S.~Karaman, and Y.~M. Marzouk}, {\em A continuous analogue
  of the tensor-train decomposition}, Comput. Methods Appl. Mech. Eng., 347
  (2019), pp.~59--84.

\bibitem{griebel2021analysis}
{\sc M.~Griebel and H.~Harbrecht}, {\em Analysis of tensor approximation
  schemes for continuous functions}, Found. Comput. Math.,  (2021), pp.~1--22.

\bibitem{hackbusch2012tensor}
{\sc W.~Hackbusch}, {\em Tensor spaces and numerical tensor calculus}, vol.~42,
  Springer Science \& Business Media, 2012.

\bibitem{Johnson-1987}
{\sc M.~Johnson}, {\em Multivariate Statistical Simulation}, Wiley, New York,
  1987.

\bibitem{kong1992note}
{\sc A.~Kong}, {\em A note on importance sampling using standardized weights},
  University of Chicago, Dept. of Statistics, Tech. Rep, 348 (1992).

\bibitem{li2020fourier}
{\sc Z.~Li, N.~Kovachki, K.~Azizzadenesheli, B.~Liu, K.~Bhattacharya,
  A.~Stuart, and A.~Anandkumar}, {\em Fourier neural operator for parametric
  partial differential equations}, arXiv preprint arXiv:2010.08895,  (2020).

\bibitem{ROM:LWG_2010}
{\sc C.~Lieberman, K.~E. Willcox, and O.~Ghattas}, {\em Parameter and state
  model reduction for large-scale statistical inverse problems}, SIAM J. Sci.
  Comput., 32 (2010), pp.~2523--2542.

\bibitem{lu2021learning}
{\sc L.~Lu, P.~Jin, G.~Pang, Z.~Zhang, and G.~E. Karniadakis}, {\em Learning
  nonlinear operators via deeponet based on the universal approximation theorem
  of operators}, Nature machine intelligence, 3 (2021), pp.~218--229.

\bibitem{SuMo:MarNa_2009}
{\sc Y.~M. Marzouk and H.~N. Najm}, {\em Dimensionality reduction and
  polynomial chaos acceleration of {B}ayesian inference in inverse problems},
  J. Comput. Phys., 228 (2009), pp.~1862--1902.

\bibitem{el2012bayesian}
{\sc T.~Moselhy and Y.~Marzouk}, {\em Bayesian inference with optimal maps}, J.
  Comput. Phys., 231 (2012), pp.~7815--7850.

\bibitem{novikov2021ttde}
{\sc G.~S. Novikov, M.~E. Panov, and I.~V. Oseledets}, {\em Tensor-train
  density estimation}, in Proc. 37th Conf. on Uncertainty in Artificial
  Intelligence, vol.~161 of Proceedings of Machine Learning Research, 2021,
  pp.~1321--1331.

\bibitem{oseledets2010tt}
{\sc I.~Oseledets and E.~Tyrtyshnikov}, {\em {TT}-cross approximation for
  multidimensional arrays}, Linear Algebra and its Applications, 432 (2010),
  pp.~70--88.

\bibitem{oseledets2011tensor}
{\sc I.~V. Oseledets}, {\em Tensor-train decomposition}, SIAM J. Sci. Comput.,
  33 (2011), pp.~2295--2317.

\bibitem{papaioannou2016sequential}
{\sc I.~Papaioannou, C.~Papadimitriou, and D.~Straub}, {\em Sequential
  importance sampling for structural reliability analysis}, Structural safety,
  62 (2016), pp.~66--75.

\bibitem{parno2018transport}
{\sc M.~D. Parno and Y.~M. Marzouk}, {\em Transport map accelerated markov
  chain monte carlo}, SIAM/ASA J. Uncertain. Quantif., 6 (2018), pp.~645--682.

\bibitem{peherstorfer2016multifidelity}
{\sc B.~Peherstorfer, T.~Cui, Y.~Marzouk, and K.~Willcox}, {\em Multifidelity
  importance sampling}, Comput. Methods Appl. Mech. Eng., 300 (2016),
  pp.~490--509.

\bibitem{peherstorfer2018multifidelity}
{\sc B.~Peherstorfer, B.~Kramer, and K.~Willcox}, {\em Multifidelity
  preconditioning of the cross-entropy method for rare event simulation and
  failure probability estimation}, SIAM/ASA J. Uncertain. Quantif., 6 (2018),
  pp.~737--761.

\bibitem{rdgs-tt-gauss-2020}
{\sc P.~B. Rohrbach, S.~Dolgov, L.~Grasedyck, and R.~Scheichl}, {\em Rank
  bounds for approximating {Gaussian} densities in the {Tensor-Train} format},
  SIAM/ASA J. Uncertain. Quantif..,  (2022).
\newblock to appear.

\bibitem{rosenblatt1952remarks}
{\sc M.~Rosenblatt}, {\em Remarks on a multivariate transformation}, The Annals
  of Mathematical Statistics, 23 (1952), pp.~470--472.

\bibitem{scheichl-qmc-bayes-2017}
{\sc R.~Scheichl, A.~M. Stuart, and A.~L. Teckentrup}, {\em {Q}uasi-{M}onte
  {C}arlo and {M}ultilevel {M}onte {C}arlo methods for computing posterior
  expectations in elliptic inverse problems}, SIAM/ASA J. Uncertain. Quantif.,
  5 (2017), pp.~493--518.

\bibitem{schwab2012sparse}
{\sc C.~Schwab and A.~M. Stuart}, {\em Sparse deterministic approximation of
  bayesian inverse problems}, Inverse Problems, 28 (2012), p.~045003.

\bibitem{spantini2018inference}
{\sc A.~Spantini, D.~Bigoni, and Y.~Marzouk}, {\em Inference via
  low-dimensional couplings}, The Journal of Machine Learning Research, 19
  (2018), pp.~2639--2709.

\bibitem{tripathy2018deep}
{\sc R.~K. Tripathy and I.~Bilionis}, {\em Deep uq: Learning deep neural
  network surrogate models for high dimensional uncertainty quantification},
  Journal of computational physics, 375 (2018), pp.~565--588.

\bibitem{uribe2020cross}
{\sc F.~Uribe, I.~Papaioannou, Y.~M. Marzouk, and D.~Straub}, {\em
  Cross-entropy-based importance sampling with failure-informed dimension
  reduction for rare event simulation}, SIAM/ASA J. Uncertain. Quantif., 9
  (2021), pp.~818--847.

\bibitem{wagner2020multilevel}
{\sc F.~Wagner, J.~Latz, I.~Papaioannou, and E.~Ullmann}, {\em Multilevel
  sequential importance sampling for rare event estimation}, SIAM J. Sci.
  Comput., 42 (2020), pp.~A2062--A2087.

\bibitem{wagner2021error}
{\sc F.~Wagner, J.~Latz, I.~Papaioannou, and E.~Ullmann}, {\em Error analysis
  for probabilities of rare events with approximate models}, SIAM J. Numer.
  Anal., 59 (2021), pp.~1948--1975.

\bibitem{wan2020coupling}
{\sc X.~Wan and S.~Wei}, {\em Coupling the reduced-order model and the
  generative model for an importance sampling estimator}, Journal of
  Computational Physics, 408 (2020), p.~109281.

\bibitem{wang2022minimax}
{\sc S.~Wang and Y.~Marzouk}, {\em On minimax density estimation via measure
  transport}, arXiv preprint arXiv:2207.10231,  (2022).

\bibitem{SuMo:XiuKar_2002}
{\sc D.~Xiu and G.~E. Karniadakis}, {\em The {W}iener-{A}skey polynomial chaos
  for stochastic differential equations}, SIAM J. Sci. Comput., 24 (2002),
  pp.~619--644.

\bibitem{yan2019adaptive}
{\sc L.~Yan and T.~Zhou}, {\em Adaptive multi-fidelity polynomial chaos
  approach to bayesian inference in inverse problems}, Journal of Computational
  Physics, 381 (2019), pp.~110--128.

\bibitem{zhou2020adaptive}
{\sc L.~Yan and T.~Zhou}, {\em An adaptive surrogate modeling based on deep
  neural networks for large-scale bayesian inverse problems}, Communications in
  Computational Physics, 28 (2020), pp.~2180--2205.

\bibitem{zahm2018certified}
{\sc O.~Zahm, T.~Cui, K.~Law, A.~Spantini, and Y.~Marzouk}, {\em Certified
  dimension reduction in nonlinear bayesian inverse problems}, Mathematics of
  Computation, 91 (2022), pp.~1789--1835.

\bibitem{zhu2019physics}
{\sc Y.~Zhu, N.~Zabaras, P.-S. Koutsourelakis, and P.~Perdikaris}, {\em
  Physics-constrained deep learning for high-dimensional surrogate modeling and
  uncertainty quantification without labeled data}, Journal of Computational
  Physics, 394 (2019), pp.~56--81.

\end{thebibliography}
\end{document}